\documentclass[pdftex]{bmvc2k}

\usepackage{graphicx}
\usepackage{subfig}
\usepackage{booktabs}
\usepackage{amssymb}
\usepackage{url}

\title{Feature Space Saturation during Training}

\addauthor{Mats L. Richter}{matrichter@uni-osnabrueck.de}{1}
\addauthor{Justin Shenk}{shenkjustin@gmail.com}{2}
\addauthor{Wolf Byttner}{wolf@byttner.org}{3}
\addauthor{Anna Wiedenroth}{awiedenroth@uni-osnabrueck.de}{1}
\addauthor{Mikael Huss}{mikael.huss@gmail.com}{4}

\addinstitution{
 Institute of Cognitive Science\\
 University of Osnabrueck\\
 Osnabrueck, Germany
}

\addinstitution{
 VisioLab\\
 Berlin, Germany
}

\addinstitution{
 Rapid Health Ltd\\
 London, UK
}

\addinstitution{
 Peltarion\\
 Stockholm, Sweden 
}

\runninghead{Richter, Shenk, Byttner, Wiedenroth, Huss}{Feature Space Saturation}

\def\etal{\emph{et al}\bmvaOneDot}

\begin{document}

\maketitle

\begin{abstract}
	We propose layer saturation - a simple, online-computable method for analyzing the information processing in neural networks.
	First, we show that a layer's output can be restricted to an eigenspace of its covariance matrix without performance loss.
	We propose a computationally lightweight method that approximates the covariance matrix during training.
	From the dimension of its relevant eigenspace we derive \textit{layer saturation} - the ratio between the eigenspace dimension and layer width.
	We show evidence that saturation indicates which layers contribute to network performance.
	We demonstrate how to alter layer saturation in a neural network by changing network depth, filter sizes and input resolution.
	Finally we show that pathological patterns of saturation are indicative of parameter inefficiencies caused by a mismatch between input resolution and neural architecture.
\end{abstract}

\section{Introduction} %
\label{Introduction}

In recent years various techniques have been proposed for exploring the properties of neural network layers.
Understanding how neural networks process information and how this processing may be influenced is vital for designing more efficient and better performing neural architectures.        
The works of Zeiler \etal{} \cite{zeiler}, Szegedy \etal{} \cite{rethinkGeneralization} and Yosinski \etal{} \cite{transferrable} are examples of experimental work that show the boundaries and limits of generalization and transferability of features.
Recent works by Raghu \etal{} \cite{svcca} and Alain \etal{} \cite{alain2016} propose techniques that allow for a deeper analysis of networks on a layer wise level.

The common problem with these and other techniques for analyzing the properties of neural networks is their complexity and computational inefficiency, which makes them impractical to use in neural architecture development or in more quantitative studies \cite{alain2016, svcca, featureAttribution}.

This work shows, that a simple, on-line computable property like the covariance matrix of the layer outputs is able to give interesting insights into the dynamics of the inference process.
To enable practical application, we provide a technique to efficiently compute the covariance matrix.
We show how to use Principal Component Analysis (PCA) to project the output of all layers into low-dimensional spaces while not negatively affecting predictive performance.
We refer to these subspaces as \textit{relevant eigenspaces}.

Based on these findings, we derive \textit{saturation} as a metric for analyzing the dynamics of the inference process.
Similar to the work of Alain \etal{}~\cite{alain2016}, saturation can be thought of as a level indicator or thermometer, showing the complexity of the processing in the respective layers.
By analyzing the distribution of saturation values within the network, we identify the ``tail pattern'' as a pathological symptom of a parameter-inefficient inference process. 
Finally, we propose simple, saturation-based strategies for altering the neural architecture to resolve such parameter inefficiencies.

\section{Related work}
\label{related_work}
In this work, we are interested in analyzing convolutional neural network models layer by layer.
The most notable inspiration for this work is SVCCA by  Raghu \etal{} \cite{svcca} as well as the follow-up work by Morcos \etal{} \cite{svcca2}, who use singular value decomposition for comparing the learned features of different models and layers.
Another inspiration for this paper is Montavon \etal{} \cite{kernelPCA}, in which kernel PCA with radial basis functions alongside linear classifiers are used to perform their analysis.
Functionally similar to saturation are logistic regression probes proposed by Alain \etal{} \cite{alain2016}, which will be utilized in this work as well in order to relate saturation patterns to parameter-inefficiencies caused by unproductive layers.
Saturation was initially proposed by Shenk~\cite{justinthesis} and applied to model parameterization by Shenk \etal{} \citep{shenk}. 
The saturation metric is used by our follow-up work to study the role of the input resolution in neural network training \cite{sizematters}. 
Further follow-up publications are basing proposed design guidelines for neural architectures on insights gained by a saturation-based analysis \cite{goingdeeper, iranconf}.

\section{Layer eigenspaces}
\label{sec:layer-eigenspaces}
In this section, we will explore the properties of the variance eigenspaces of each layer's feature space in order to motivate the derivation of the metric saturation.
First, the methodology of computing the layer wise variance eigenspaces during training is described. 
This is followed by an experimental part, where we demonstrate that the eigendirections of the highest variance contain most of the information required to solve the trained classification task.
We will further show that relevant eigenspaces exist and that they contain fewer dimensions than the original feature spaces of the network.
Based on these findings, we will introduce saturation as a metric for studying the inference dynamics of neural network models.

\subsection{Computing variance eigenspaces and relevant eigenspaces}
Below is a brief discussion of the method we use to compute variance eigenspaces and relevant eigenspaces in our experiments.
We apply PCA on the layer output to determine the eigenspace of the layer's features.
Then we sum the largest eigenvalues that explain a percentage $\delta$ of the layer output variance.
The space spanned by these eigenvectors is the variance eigenspace. In this way, we find candidate eigenspaces for the layer. This process is described in section \ref{finding-layer-eigenspaces}.

We establish that the layer's output is contained in the eigenspace as follows. We project all validation set output vectors into the space and determine whether the network's validation performance changes.
To do so, we add special projection layers that only change the output at validation time.
This we call a \textit{projected network}.
We then apply Student's paired \textit{t}-test to determine if the network validation performance difference between the projected and the normal network is statistically significant.
We pick the smallest $\delta$ such that there is not a statistically significant difference (p<.01). The details are in section \ref{projecting_output}.
We consider the eigenspaces of projected networks with no statistically significant changes in performance as an approximation of the \textit{relevant eigenspaces}.

\subsection{Finding layer eigenspaces}
\label{finding-layer-eigenspaces}
A common problem of many analysis tools for neural networks is, that they are resource intensive to compute.
For example, logistic regression probes by Alain and Bengio~\cite{alain2016} and SVCCA by Raghu \etal{}~\cite{svcca} can require significantly more computation time and RAM than the training of the model, especially for large datasets and models.
For practical application, this is a significant drawback.
Ideally, the analysis can be conducted during training with little computational overhead to minimize the cycle time of experiments.
For this reason, we propose an on-line algorithm for computing the covariance matrix during the regular forward pass.
In this section, we are interested in finding a subspace of the layer output space to which we can restrict layer output vectors $z_{l,i}$ without changing the network's validation performance. We use PCA on the layer output matrix $A_l := (z_{l,1},...,z_{l,n})$ of $n$ samples at training time, thus $A_l \in \mathbb{R}^{n \times w}$ where $w$ is the layer width.
To do this efficiently, we compute the covariance matrix $Q(Z_l,Z_l)$, where $Z_l := \sum^{n}_{i=1}(z_{l,i})/n$, using the covariance approximation algorithm between two random variables $X$ and $Y$ with $n$ samples:
\begin{equation}
	Q(X, Y) = \frac{\sum^{n}_{i=1} x_i y_i}{n} - \frac{(\sum^{n}_{i=1} x_i)  (\sum^{n}_{i=1} y_i)}{n^2}
\end{equation}
We make this computation more efficient by exploiting the shape of the layer output matrix $A_l$:
We compute $\sum^{n}_{i=1} x_i y_i$ for all feature combinations in layer $l$ by calculating the running squares $\sum^{B}_{b=0}A_{l,b}^T A_{l,b}$ of the batch output matrices $A_{l,b}$ where $b \in \{0,...,B-1\}$ for $B$ batches. We replace $\frac{(\sum^{n}_{i=1} x_i)  (\sum^{n}_{i=1} y_i)}{n^2}$ by the outer product $\bar{A}_l \bigotimes \bar{A}_l$ of the sample mean $\bar{A}_l$.
This is the running sum of all outputs $z_{l,k}$, where $k \in \{0,...,n\}$ at training time, divided by the total number of training samples $n$.
The final formula for covariance approximation is then:
\begin{equation}
	Q(Z_l, Z_l) = \frac{\sum^{B}_{b=0}A_{l,b}^T A_{l,b}}{n} -(\bar{A}_l \bigotimes \bar{A}_l)
\end{equation}
Since we only store the sum of squares, the running mean and the number of observed samples, we require constant memory and computation is done batch-wise.
The algorithm requires roughly the same number of computations as the processing of a forward pass of the respective layer does; thus we compute saturation after every epoch. The variables are reset at the beginning of each epoch to minimize the bias induced by weight updates during training.
Our algorithm uses a thread-save common value store on a single compute device or node, which furthermore allows to update the covariance matrix asynchronous when the network is trained in a distributed manner.

In convolutional layers, we treat every kernel position as an individual observation.\footnote{This turns an output-tensor of shape (samples $\times$ height $\times$ width $\times$ filters) into a data matrix of shape (samples $\cdot$ height $\cdot$ width $\times$ filters).}
The advantage of this strategy is that no information is lost, while keeping $Q$ at a manageable size.
This strategy was proposed by Raghu \etal{} \cite{svcca} after their initial publication and Garg \etal{}  \cite{sat_pruning1}, who use it in their PCA-based pruning strategy for CNNs.

To determine the eigenspace $E_l^{k}$ such that the projection of output vectors $z_{l,i}$ to $E_l^{k}$ is as lossless as possible, we find $k := arg \, max(\frac{\sum \lambda_k}{N}) \leq \delta$ where $\sum \lambda_k$ is the sum of the largest $k$ eigenvalues.\footnote{In order to achieve more accurate results on small networks, we treat the threshold as soft. If $max(\sum \lambda_k) > \delta \leq max(\sum \lambda_k) + 0.02$ an additional dimension is added. If $dim \, E_l^k = 0$, because a single dimension exceeds the $\delta$-threshold, we set $E := \{v_1\}$.}
This technique is similar to how Raghu \etal{} \cite{svcca} apply singular value decomposition in SVCCA - that study fixes $\delta$ to 99\% of the variance.
Pruning strategies by Garg \etal{} \cite{sat_pruning1} and Chakraborty \etal{} \cite{sat_pruning2} settle on $\delta$ of 99.9\% and 99\% respectively.

\subsection{Exploring the properties of projected networks}
\label{projecting_output}
Our approximation of the \textit{relevant eigenspace} is by the nature of PCA a linear approximation.
Since neural networks are non-linear models, it is not guaranteed that a linear subspace can capture the information relevant for the inference process accurately. %
Therefore, we demonstrate experimentally in this section the usefulness of PCA for this application.
First, we study the effects of different values for $\delta$ and show that variance eigenspaces contain more information necessary for the inference process than randomly chosen orthonormal subspaces of equal dimensionality.
We will then demonstrate, on VGG13 and VGG19, that $\delta$ can be chosen such that the variance eigenspaces in all layers of the network are relevant eigenspaces.
To study the effect of different values for $\delta$ we introduce PCA-Layers, inserted after any non-output (fully connected and convolutional) layer $l$. At training time PCA-Layers are pass-through layers. At testing time they project the output of the  preceding layer $A_{l}$ into  the variance eigenspace $E_l^k$.
This is done by multiplying $A_l$ with the projection matrix, $P_{E^k_l} = E^k_l (E^k_l)^T$. %
For convolutional layers, we compute the $(1 \times 1)$ convolution $A_l \ast vec(P_{E_l^k})$. The net effect is to turn a dataset problem into a network parameter one; we study the properties of samples by changing the projection parameters.

First, we study how well the eigenspaces are able to maintain the predictive performance of the model compared to random orthonormal subspaces of equal dimensionality.
We train 20 different variations of VGG-style networks on CIFAR10 \cite{CIFAR-10}.\footnote{In order to include networks of different width (filter sizes) and depth (number of layer) we trained VGG[11,13,16,19] as well as variations of all those architectures with filter sizes reduced by a factor of [2,4,8,16]. All models were trained on a batch size of 128 using the Adam optimizer and a learning rate of 0.001 for 30 epochs.}
The \textit{relative performance} of a network for a value of $\delta$ is the ratio between the test accuracy with enabled and disabled PCA-Layers.
We indirectly control the dimensionality of $E^k_l$ with $\delta$, which is set globally for the entire network.

\begin{figure}[h!]
	\centering
	\includegraphics[width=0.4\columnwidth]{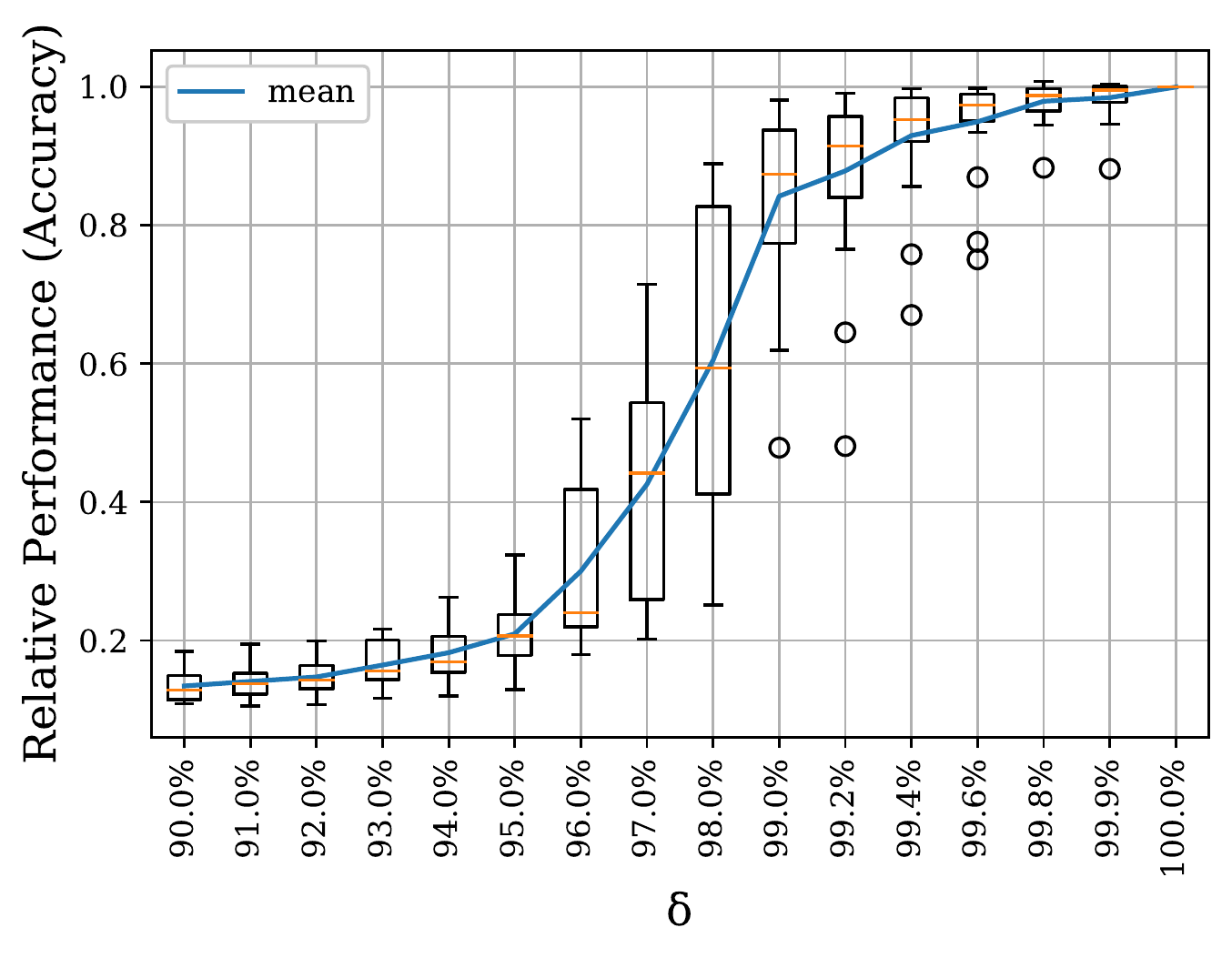}
	\includegraphics[width=0.4\columnwidth]{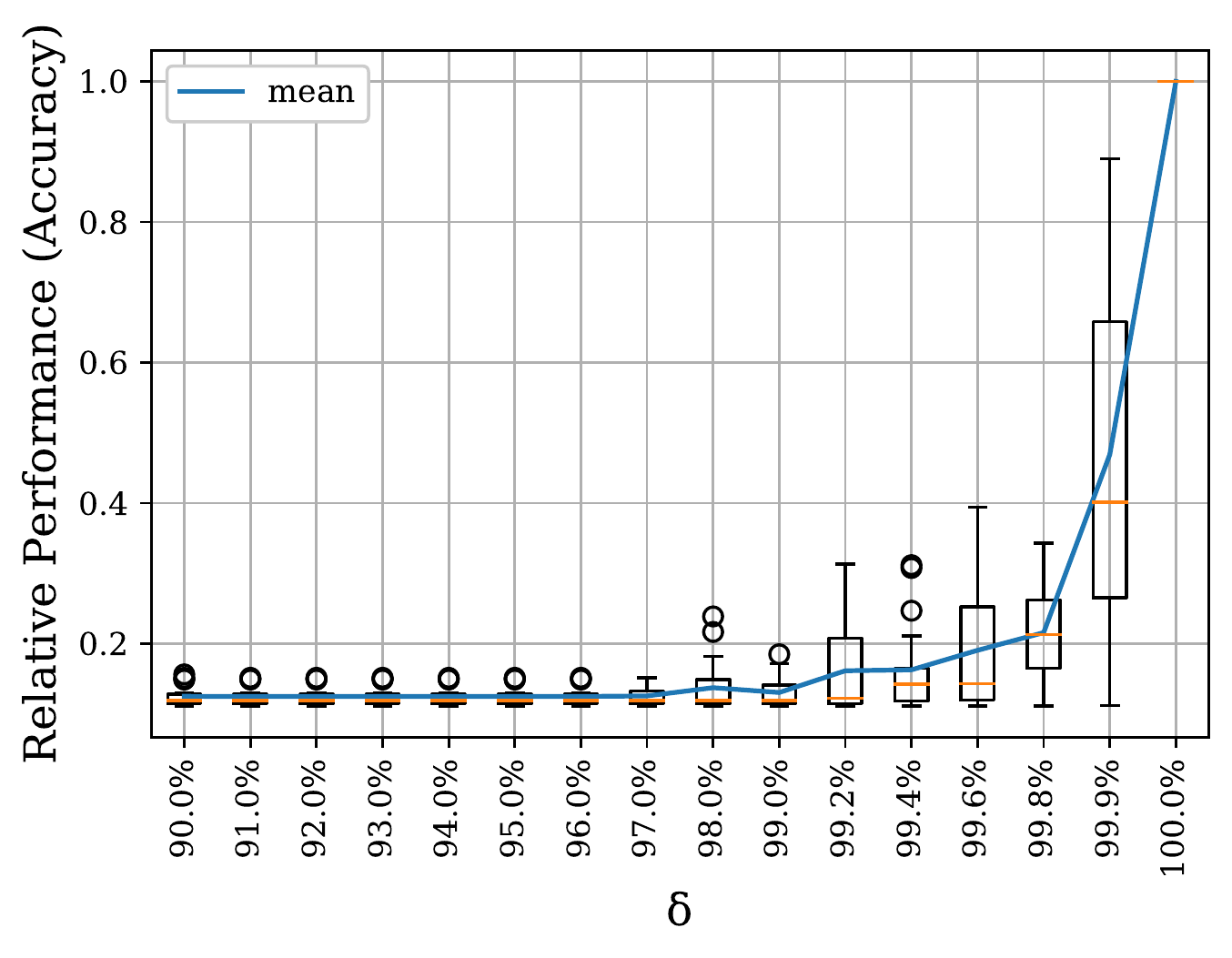}
	\captionsetup{justification=centering}
	\caption{Projecting the output of all hidden layers into an eigenspace $E^k_l$ that explains $ \delta$ \% of the variance (left) maintains predictive performance better than projecting the layers into a random orthonormal subspace of the same size (right).}
	\label{fig:projections_both}
	\vspace{-0.5cm}
\end{figure}

The results in Fig.~\ref{fig:projections_both} show that the eigenspaces of the layer outputs are able to maintain the information better than equally sized random orthonormal projections.
Relative performance given lower values of $\delta$ also degrades slower than random projections.

Next, we explore the upper bounds of this projection by finding a $\delta < 1$ for a trained model such that the difference in predictive performance to the model with disabled PCA-Layers is insignificant.
For each network we compare a network's projected and unprojected CIFAR10 \textit{validation set} performance.
We use the eigenspace $E_l^k$ computed during training. We test whether a network's performance changes relative to the unprojected performance.
\begin{table}[!htb]
	\begin{center}

		\begin{tabular*}{0.7\columnwidth}{c @{\extracolsep{\fill}} c c c c c }
			\toprule
			$\delta$ & $\mu_{p-\bar{p}}$  & $\sigma$ & t & p
			\\
			\midrule
			0.9999 & -0.0004  & 0.0008 & -2.42 & 0.023 \\
			\textit{0.9998} & -0.0005  & 0.0009 & -2.81 & 0.010 \\
			\textit{0.999} & \textbf{-0.0017}  & 0.0016 & -5.50 & 0.000 \\
			\textit{0.998} & -0.0017  & 0.0022 & -3.92 & 0.001 \\
			\textbf{0.996} & -0.0005  & 0.0030 & -0.91 & 0.371 \\
			\textit{0.994} & 0.0037  & 0.0043 &  4.45 & 0.000 \\
			\textit{0.99} & 0.0178  & 0.0136 &  6.68 & 0.000 \\
			\bottomrule
		\end{tabular*}
			\vspace{0.3cm}
			\caption{Based on $n=26$ trained VGG13 models, we find that $\delta=0.996$ allows us to approximate the relevant eigenspace in all layers, such that the performance $p$ becomes indistinguishable from the unprojected model. 
			This is based on a \textit{t}-statistic (student's two-tailed t-test), $\mu \neq 0$ (p<.01), selected $\delta$ at $\alpha=0.1$. $\mu \neq 0$ in \textit{italics}. $\mu_{p-\bar{p}}$ is the average pairwise distance between the test accuracies of unprojected and projected networks.}
		\label{tab:VGG13_ttest_short}
		\vspace{-0.5cm}

	\end{center}
\end{table}
Applying Student's two-tailed \textit{t}-test to VGG13 (Table \ref{tab:VGG13_ttest_short}) shows that at $\delta = 0.996$ we cannot distinguish the projected network from the base network at significance p$<.1$. However, at $\delta = 0.999$, the projected network \textbf{outperforms} the base network by 0.17\% at significance p$<.01$. This means we can \textbf{increase} validation performance by \textbf{restricting} layer output to a subspace known at training time.

VGG13's validation performance improves in the range $\delta \in [0.998,0.9998]$, with a maximum improvement $0.17\%$.\footnote{The full results are available in the supplementary material.} This shows that there are noisy or not generalizing feature dimensions in the layer output; we use this to develop the concept of \textit{layer saturation} and improve network performance further.
\begin{table}[ht]
	\begin{center}
	
	\begin{tabular*}{0.7\columnwidth}{c @{\extracolsep{\fill}} c c c c c c }
		\toprule
		$\delta$ & t & p & $\mu_{Sat}$ & $\sigma_{Sat}$ & $\sum_l dim E^k_l$ \\
		\midrule
		0.9997 & -2.82 & 0.008 & 51.2 & 0.7 & $2071 \pm 93$ \\
		0.9996 & -1.28 & 0.208 & 48.8 & 0.6 & $1938 \pm 88$ \\
		0.9995 & \textbf{-0.352} & 0.727 & 47.1 & 0.7 & $1841 \pm 86$ \\
		0.9994 &  2.18 & 0.035 & 45.6 & 0.7 & $1766 \pm 84$ \\
		0.9993 &  2.62 & 0.012 & 44.5 & 0.7 & $1705 \pm 83$ \\ 
		\bottomrule
	\end{tabular*}
	\vspace{0.3cm}
	\caption{The table depicts VGG19 \textit{t}-statistic, as well as the mean and standard deviation of saturation and the sum of intrinsic dimensions ($\sum_l dim E^k_l$). It is also possible to find a relevant eigenspace for VGG19. Based on $n=40$ trained models, the relevant eigenspace of the projected network can be computed by using the depicted values of $\delta$ using the same $alpha$ as in Table \ref{tab:VGG13_ttest_short}.
	Interestingly, from the 5860 dimension in all feature spaces combined, only up to $2071 \pm 93$ are used to construct all relevant eigenspaces in the network.
}
	\label{tab:VGG19_ttest_short}
	\vspace{-0.5cm}

	\end{center}
\end{table}

We also study saturation behavior in VGG19 (Table \ref{tab:VGG19_ttest_short}). At $\delta=0.9995$, the $t$-value is insignificant. We need $1841 \pm  86$ of $5860$ dimensions to describe the data. Thus, we can \textbf{remove} two out of three dimensions without changing network performance.

\subsection{Saturation}
\label{saturation}
Based on previous results, we are interested in analyzing the sequence of eigenspaces that the data traverses during the forward pass.
For this purpose, we propose layer saturation:
\begin{equation}
	s_l = \frac{dim \, E^k_l}{dim \, Z_l}
\end{equation}
Intuitively this \textit{layer saturation} ratio represents the proportion of spatial dimensions occupied by the information in a layer $l$.
Therefore we can think of saturation as a level indicator which shows the fraction of useful dimensions in the output space.
We can analyze and compare the inference dynamics of multiple networks by plotting the saturation level of each network layer.

\section{Exploring the Properties of Saturation}
\label{saturatiion_exp}

\subsection{Saturation patterns are stable over different model runs}
\label{sec:stability}
The first question we want to answer is whether $s_l$ is stable enough to observe meaningful patterns.
To investigate, we train VGG16 and ResNet18 100 times using the same setup as in section 3.3. 

\begin{figure}[htb!]
	\centering
	\includegraphics[width=0.32\columnwidth]{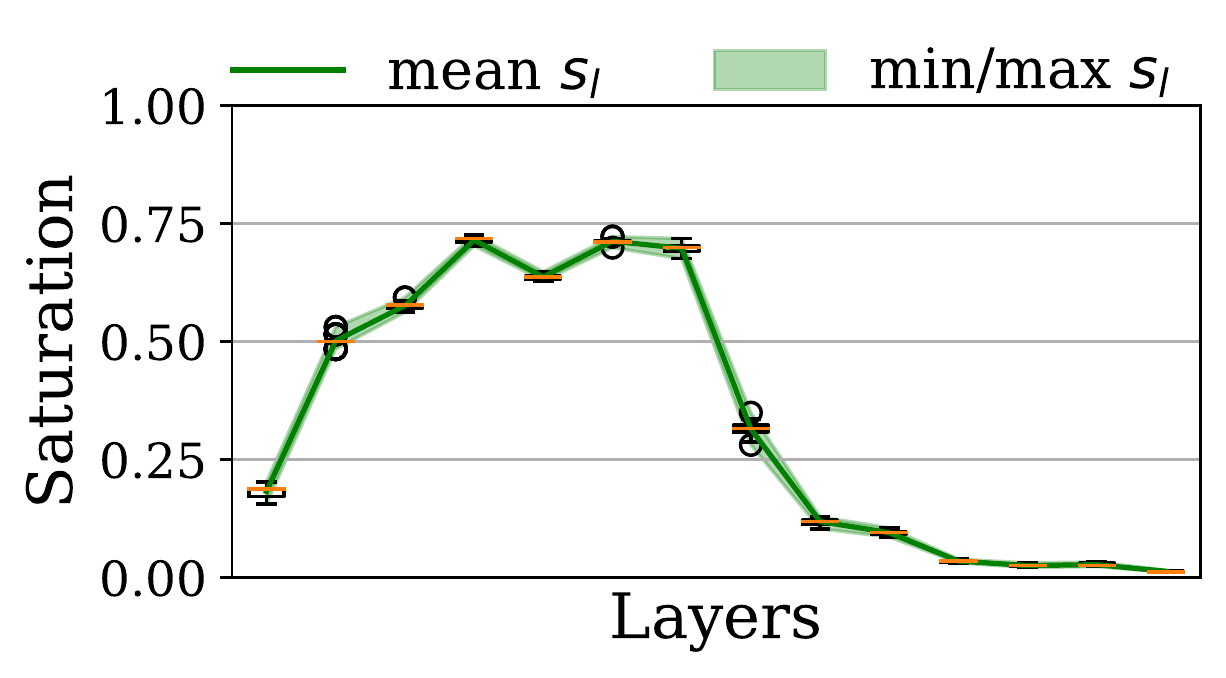}
	\includegraphics[width=0.32\columnwidth]{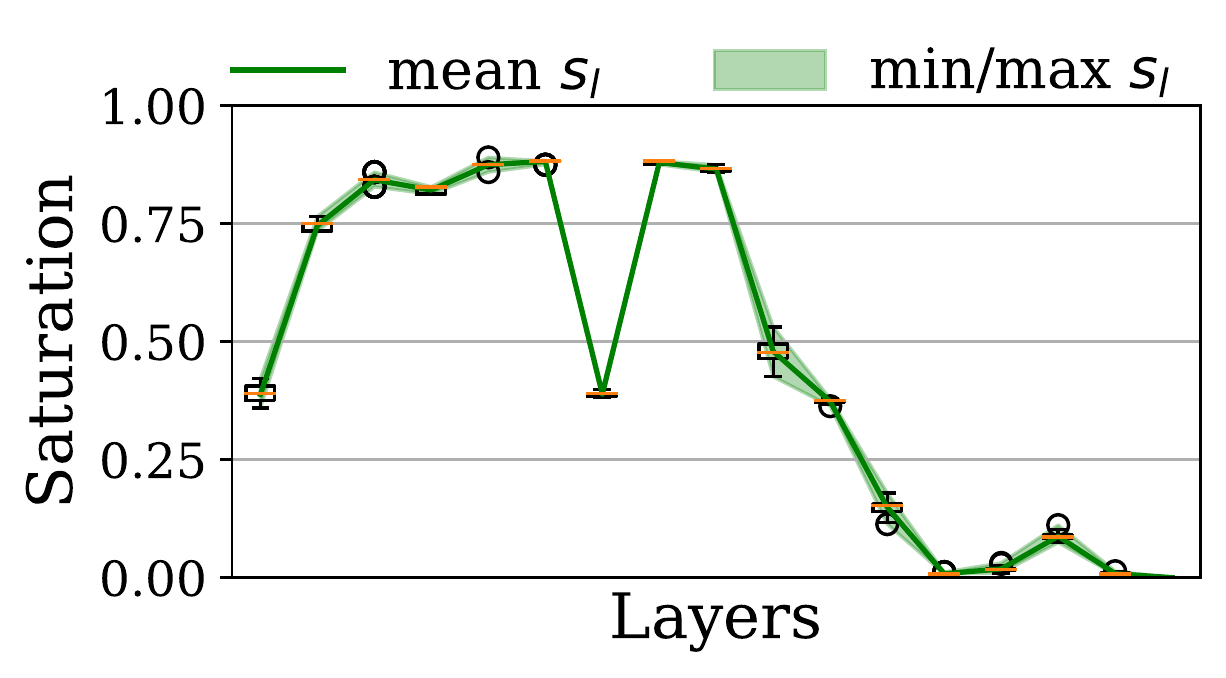}
	\vspace{0.3cm}
	\captionsetup{justification=centering}
	\caption{Saturation only deviates slightly for VGG16 (left) and ResNet18 (right) between training using the same setup for 100 training, which allows us to analyze saturation-patterns with little bias caused by random fluctuations.}
	\label{fig:stability}
	\vspace{-0.3cm}

\end{figure}

From the result in Fig.~\ref{fig:stability} we observe that for ResNet18 as well as VGG16 the emerging saturation patterns are stable.
In fact, the standard deviation $\sigma_s$ of VGG16`s saturation is 0.281 while the standard deviation of the accuracy-performance $\sigma_{acc}$ from the same model is significantly higher at 0.511, while both values are bound in [0, 1].
The same can be said for ResNet18, where $\sigma_s = 0.353$ and $\sigma_{acc} = 0.523$.
Based on these observations, we can conclude that saturation is sufficiently stable to allow for the analysis of convolutional neural networks.

\subsection{Saturation and Network Width}
\label{sec:widt}
We find \cite{iranconf} that models with lower average saturation $s_{\mu}$ tend to perform better than architecturally similar models with higher $s_{\mu}$ (see later in Fig. \ref{fig:woof}).
There are two main factors influencing $s_{\mu}$ \cite{iranconf}:
Problem difficulty, increasing $s_{\mu}$, and the width of the network, decreasing $s_{\mu}$.
The width of a network is the number of filters or units in each layer.
Effectively, more difficult problems require more capacity in each layer and thus more computational resources to be processed effectively.
Therefore, finding a sweet spot for $s_{\mu}$ for a given architecture and dataset by scaling its width optimizes the efficiency of the model for the given setup.

    \begin{figure}[htb!]
    \centering
    \includegraphics[scale=0.4]{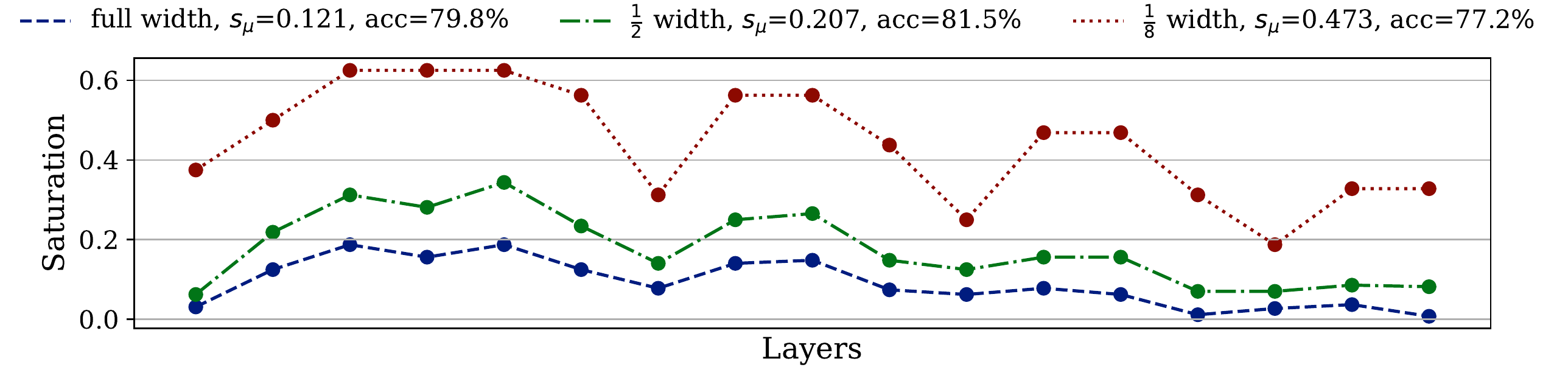}
    \vspace{0.1cm}
    \caption{ResNet18 can be considered overparameterized for the ImageNette-dataset, which is apparent from the low saturation level (blue line). Halving the number of filters also halves the computational and memory footprint while slightly improving performance (green line). Reducing the width of the network too much results in a poor performing, underparameterized model (red line). The saturation of all layers (dots) are depicted in the order of the forward pass from input to output.}
    \label{fig:width_exp1}
	\vspace{-0.3cm}
  \end{figure}

To demonstrate this, we train ResNet18 on ImageNette, a 10-class subset of ImageNet, using an $224 \times 224$ input resolution. 
To demonstrate over-, under- and well- parameterized variants of the model, we scale its width by a factor of 1, $\frac{1}{8}$ and $\frac{1}{2}$ respectively.
From the results in Fig. \ref{fig:width_exp1}, we can see that ResNet18 with $\frac{1}{2}$ width provides the best performance while requiring half the memory and FLOPS compared to ResNet18 with full width. We attribute the slightly poorer predictive performance of the full-width-model to overfitting.

From additional experiments on VGG and ResNet-style models, we find that an average saturation $s_{\mu}$ of roughly 20\% to 30\% delivers the best performance in the tested scenarios, assuming all networks have roughly the same relative distribution in saturation across layers and follow the conventional pyramidal structure of modern classifiers \cite{vgg, resnet, efficentNet}.
Models with $s_{\mu}$ below this interval provide approximately similar performance at reduced efficiency, while models with  higher $s_{\mu}$ will degrade in predictive performance, as figure \ref{fig:width_exp1} and the results of our follow-up work \cite{iranconf} demonstrate.

By adjusting the scaling of the network's width based on the current $s_{\mu}$, the network can be optimized in an informed manner.
While it is possible to manipulate individual layers similarly, we advise against it for multiple reason.
First, the saturation of individual layers is also subject to noise induced by some components like ($1 \times 1$) convolutions, downsampling and skip connections, which makes clear rules of action hard to quantify.
Second, this practice can theoretically mask true inefficiencies like tail-patterns (see sections \ref{sec:resolution} and \ref{sec:OptimizeNetworks}). 
For example, strongly overparameterizing the productive part of the model brings the saturation of these layers down until tail-patterns in the (less overparameterized) unproductive layers become hard to locate.

\subsection{Saturation patterns provide insight into the distribution of the inference process across the network}
\label{sec:resolution}
We move on to investigate the emerging saturation patterns when viewing the saturation levels of the individual layers in order of the forward pass.
To accomplish this, we use logistic regression probes by Alain \etal{} \cite{alain2016}. Logistic regression probes measure the intermediate quality of the solution in a layer by training a logistic regression classifier on its output solving the same task as the model, allowing us to measure solution progress.
The size and complexity of extracted features increases with every consecutive convolutional layer due to the expansion of the receptive field. We investigate whether this influences saturation and whether this can be connected to some patterns observed in logistic regression probes.

We train ResNet18 on Cifar10 (deliberately choosing a low-resolution dataset to avoid side effects caused by the addition of information by increasing the resolution) using three distinct input resolutions: $32 \times 32$ (Cifar10 native), $224 \times 224$ (ResNet18`s design resolution) and $1024 \times 1024$ (intentionally over-sized).
We compute the test accuracy of logistic regression probes and saturation on every convolutional layer.

\begin{figure}[htb!]
	\centering
	\subfloat[$32 \times 32$ (Cifar10 native)]{\includegraphics[width=0.32\columnwidth]{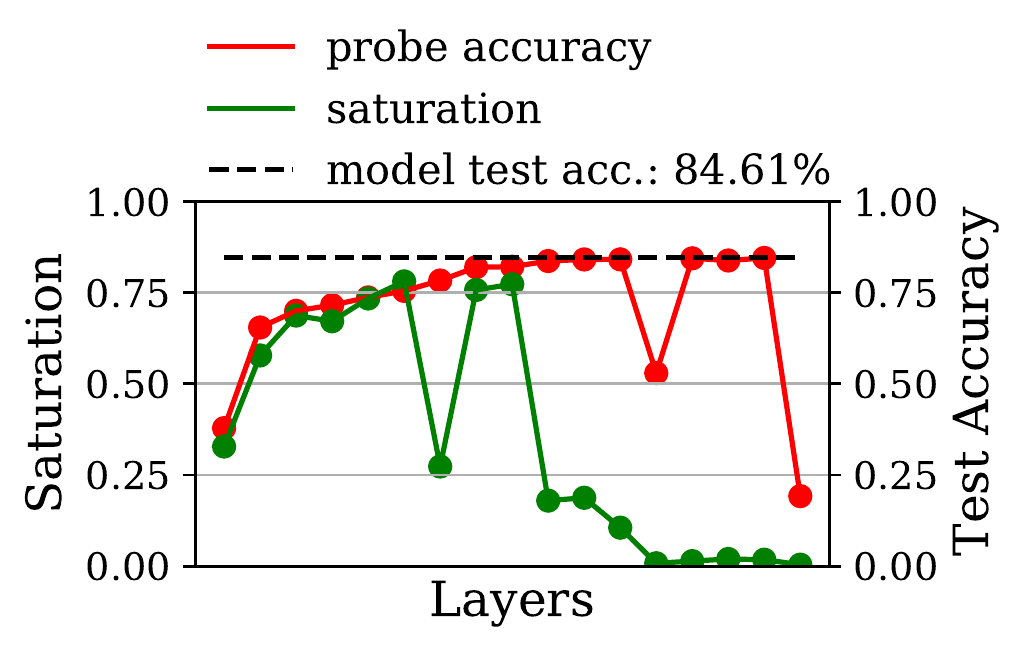}}%
	\subfloat[$224 \times 224$ (model standard)]{\includegraphics[width=0.32\columnwidth]{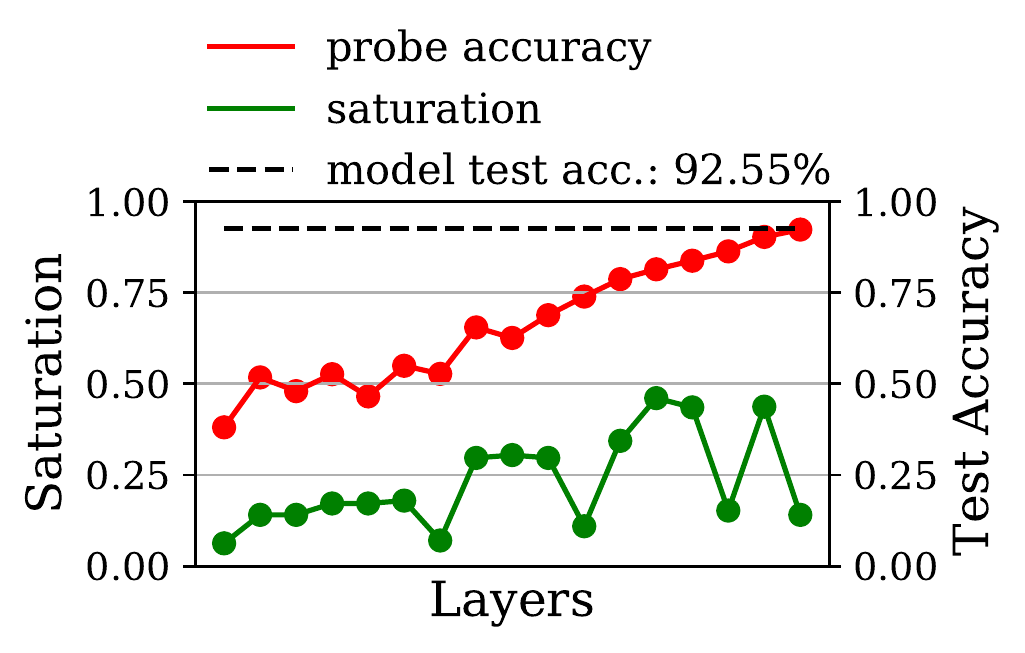}}%
	\subfloat[$1024 \times 1024$]{\includegraphics[width=0.32\columnwidth]{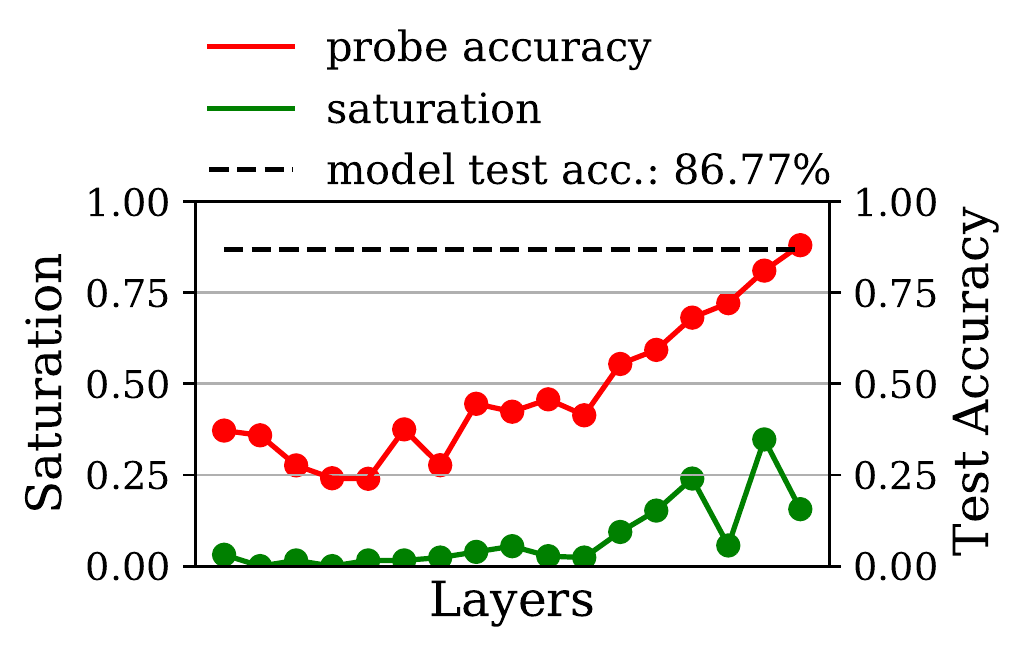}}
	\vspace{0.2cm}
	\captionsetup{justification=centering}
	\caption{The input resolution changes how the inference is distributed in the model, observable from probe accuracy and saturation patterns. Even distribution and best accuracy is achieved at the design resolution of the model ($224 \times 224$ pixels).}
	\label{fig:res_exp}
\end{figure}
In Fig.~\ref{fig:res_exp} we observe three distinct patterns in saturation and probe performances.
At Cifar10's native resolution of $32 \times 32$ pixels, we observe a high saturated sequence of layers followed by a low saturated sequence.
Only the high saturated sequence is contributing qualitatively to the inference process according to the logistic regression probes.
This pattern is inverted, when drastically over-sizing the resolution to $1024 \times 1024$ pixels.
The best predictive performance and most even distribution of the inference process is achieved when training ResNet18 on its original design resolution of $224 \times 224$ pixels.
According to the logistic regression probes, we can see that low saturated subsequences of layers are indicative of unproductive layers in the network. We refer to this pattern indicating a parameter inefficiency as \textit{tail pattern}. 
From additional experiments on a variety of models and datasets (see supplementary material for detailed results) we deduct a definition of a tail pattern: 

\textit{A tail is a subsequence of at least 3 consecutive layers in a feed-forward neural architecture with an average saturation at least 50\% lower relative to the average saturation of the rest of the network.}

This definition is imperfect and does not fit all patterns that we would visually classify as similar to a tail pattern.
However, to test the implications of the presence of such patterns, it is necessary to have a more rigorous definition than purely visual observation.

The experiment of this section demonstrates that we can use the saturation values of the individual layers to gain insights on how the inference process is distributed among the networks layers.
The insights shown in this work are expanded on upon in our follow-up works \cite{sizematters, goingdeeper}, where we explain these inefficiencies experimentally with the expansion of the receptive field.

\section{Optimizing Convolutional Neural Networks with Saturation}
\label{sec:OptimizeNetworks}
In this section, we focus on the practical application of saturation for optimizing neural architectures.
While, as previously demonstrated, it is possible to remove a tail pattern and therefore parameter inefficiency by altering the input resolution of the model (see Fig. \ref{fig:res_exp}), an increase in resolution scales quadratic with the memory footprint and the FLOPs required per forward pass, making it an expensive solution.
We therefore present two simple strategies for altering the neural architecture on a few selected examples as an alternative way of removing this inefficiency.
The first strategy is to bring the receptive field of the feature extractor closer to the input resolution.
To optimize ResNet18 for a $32 \times 32$ pixel resolution, we replace the first two layers, which are sometimes referred to as ``stem'' by a $3 \times 3$ convolution, with stride size 1. By removing the initial downsampling layers, we reduce the receptive field size of ResNet18 from $435 \times 435$ to $109 \times 109$ pixels.
\vspace{-0.4cm}
\begin{figure}[htb!]
	\centering
	\subfloat[ResNet18]{\includegraphics[width=0.32\columnwidth]{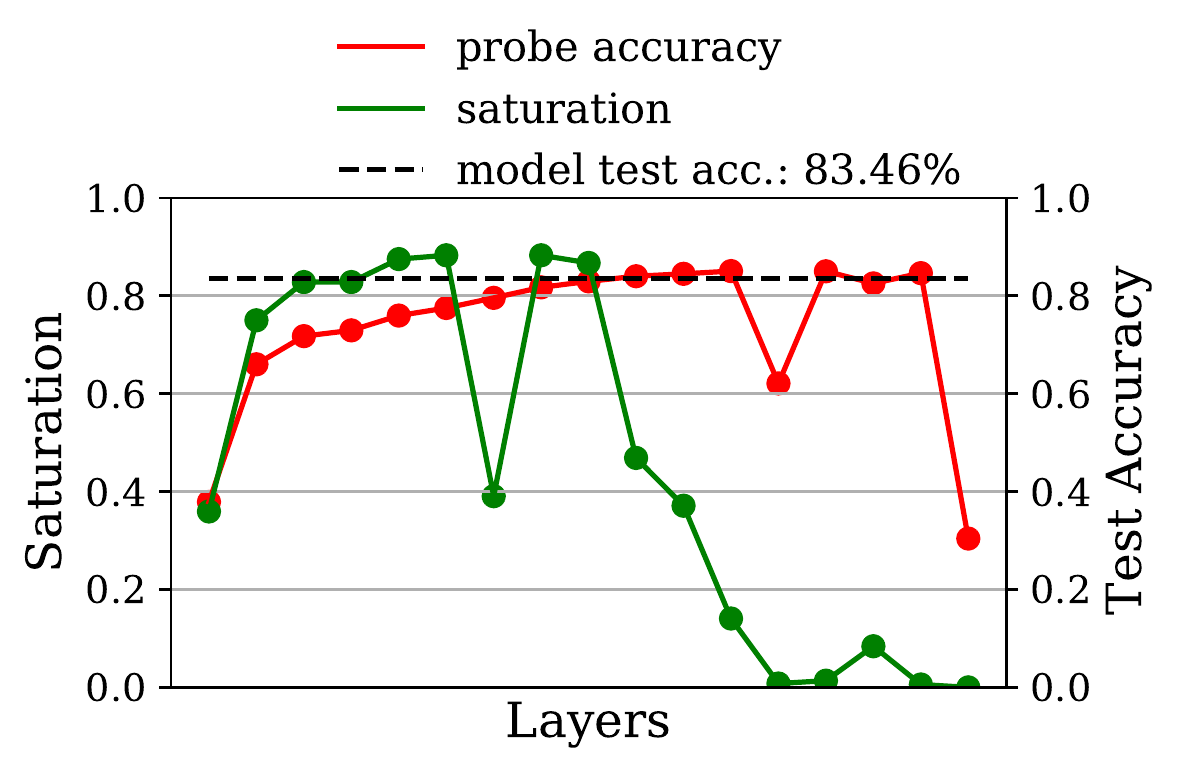}}\quad
	\subfloat[ResNet18 (no Stem)]{\includegraphics[width=0.32\columnwidth]{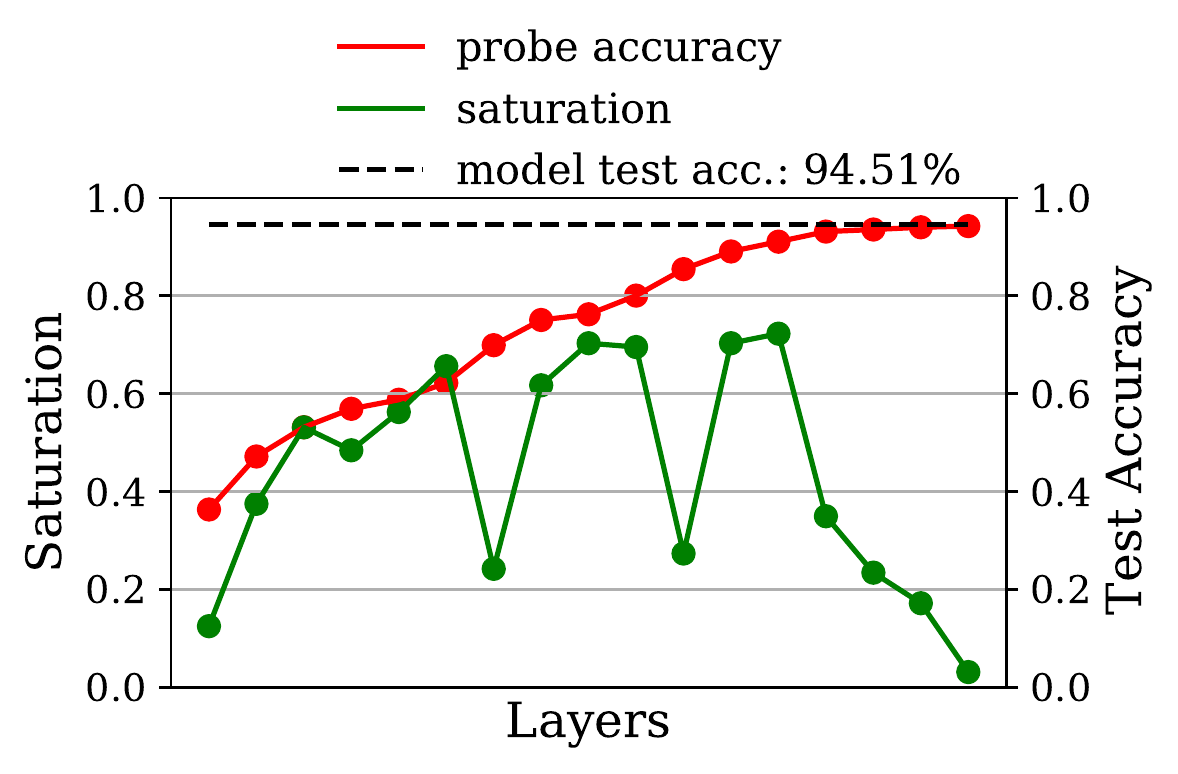}}
	\vspace{0.2cm}
	\captionsetup{justification=centering}
	\caption{Reducing the size of the receptive field (here by removing the two initial downsampling layers) removes the tail pattern and improves the test accuracy on Cifar10 at native resolution.}
	\label{fig:betterResnet}

\end{figure}

In Fig.~\ref{fig:betterResnet} we can see an improved version of ResNet18 that was trained on Cifar10 at $32 \times 32$ pixel resolution. The performance of the model drastically increases to over $94\%$ accuracy and the tail pattern is removed.
The computations required for a forward pass of a single image increase from 0.04 GFLOPS to 0.56 GFLOPS. For comparison, increasing the input resolution to the original resolution of ResNet18, which is $224 \times 224$ pixels, yields the same result but requires 1.83 GFLOPs per forward pass per image.
\begin{figure}[h!]
	\centering
	\subfloat[VGG19]{\includegraphics[width=0.32\columnwidth]{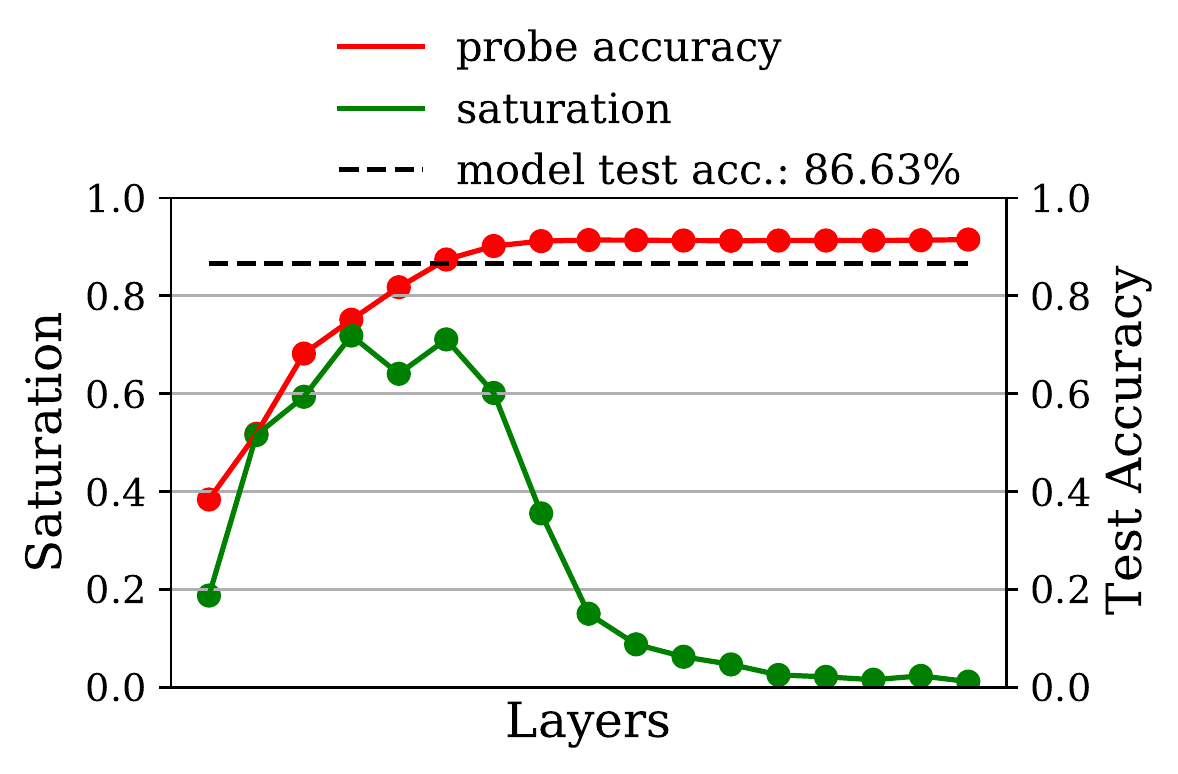}}\quad
	\subfloat[VGG19 (truncated)]{\includegraphics[width=0.32\columnwidth]{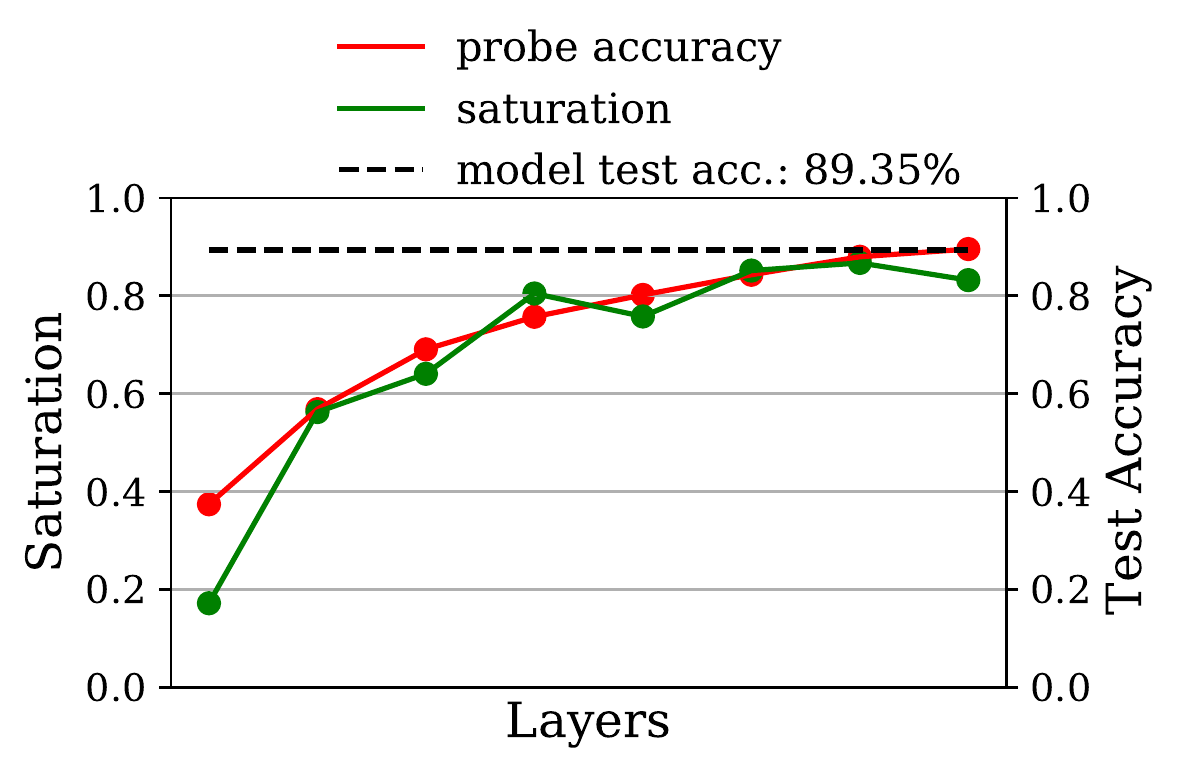}}
	\vspace{0.3cm}
	\captionsetup{justification=centering}
	\caption{Removing the tail layers of VGG19 and retraining the truncated model reduced the computational and memory footprint, meanwhile improving the performance slightly. Both models are trained on Cifar10 at native resolution.}
	\label{fig:truncatedvgg}
	\vspace{-0.5cm}
\end{figure}

\begin{figure}[htb!]
	\centering
	\subfloat[VGG16 full width]{\includegraphics[width=0.32\columnwidth]{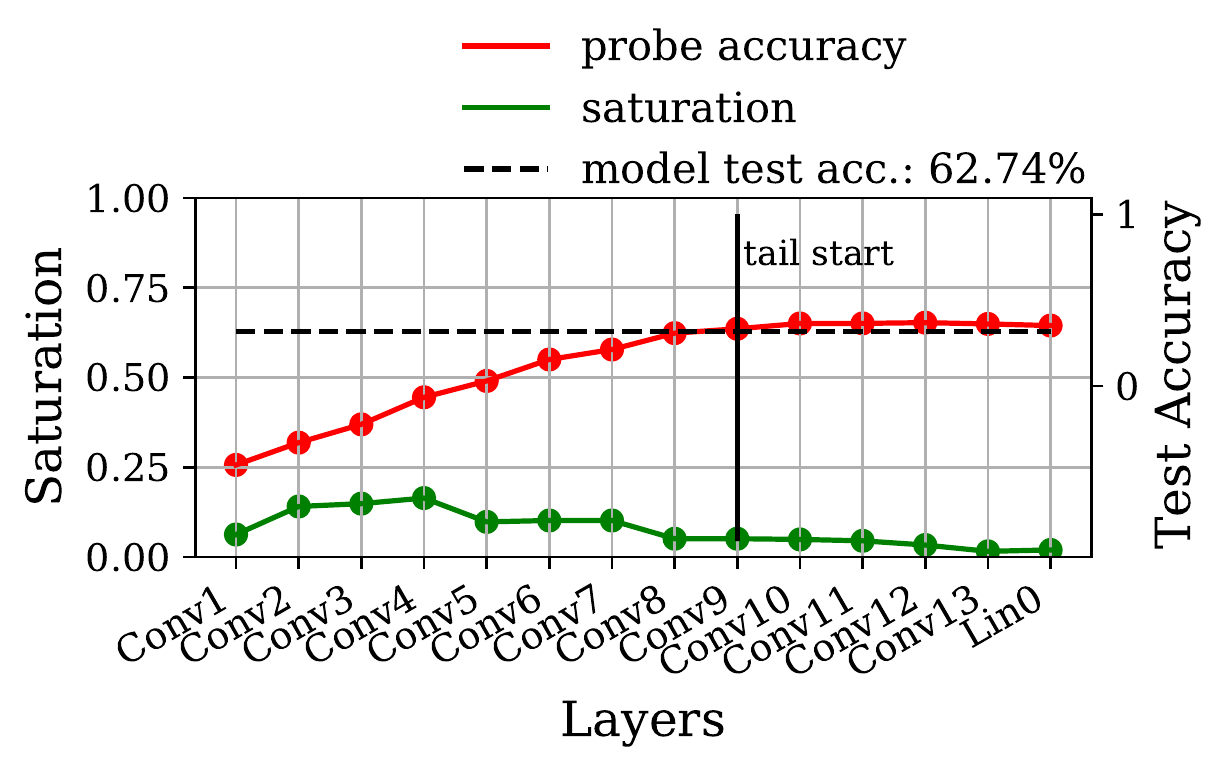}}\quad
	\subfloat[VGG16 $\frac{1}{8}$ width]{\includegraphics[width=0.32\columnwidth]{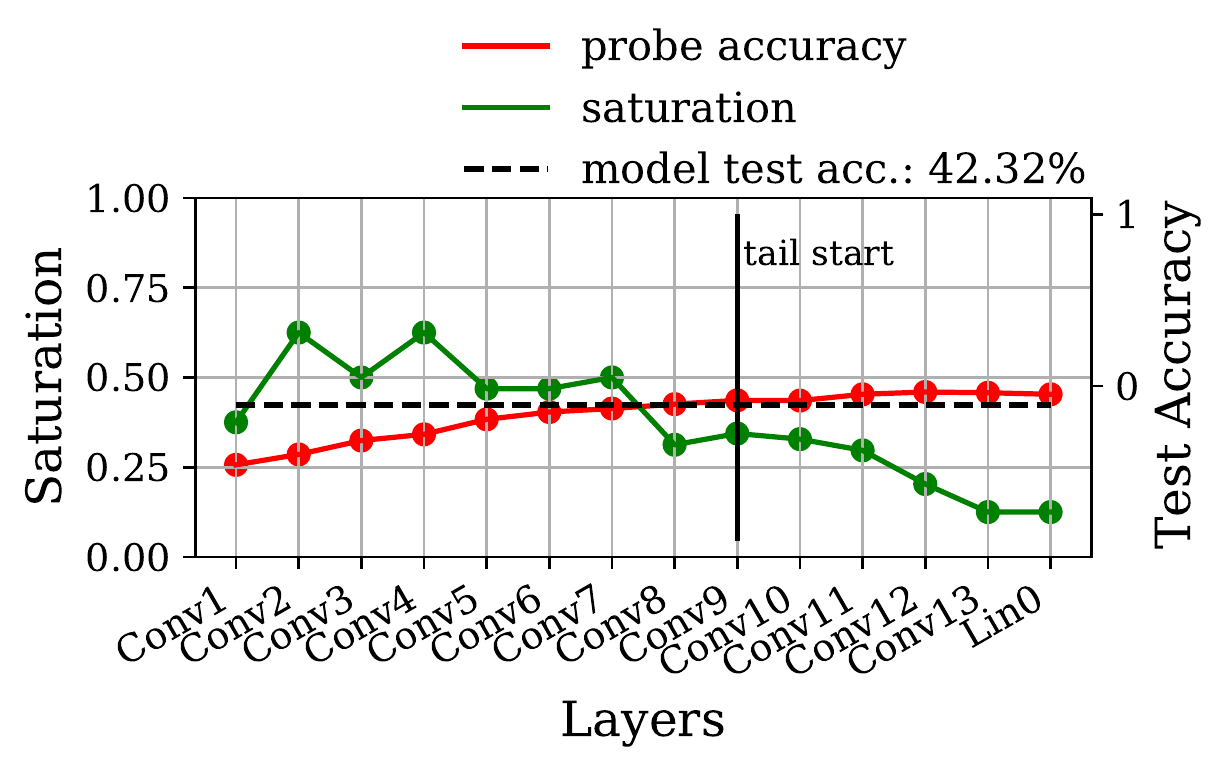}}
	\vspace{0.3cm}
	\captionsetup{justification=centering}
	\caption{The width of the VGG16 trained on the ImageWoof dataset at $32 \times 32$ pixel input resolution does not change the size of the tail, meaning that inefficiencies caused by low-saturated tails and under-parameterized layers are independent.}
	\label{fig:woof}
	\vspace{-0.5cm}

\end{figure}

Another, more simple possibility to remove the tail pattern is to remove the layers of the tail-pattern and retrain the model.
A quick test on VGG19 (see Fig. \ref{fig:truncatedvgg}) demonstrates that retraining the truncated model reduced the number of parameters from 20,170,826 to 2,363,338, the FLOPS needed per forward pass from 399 MFLOPS to 229 MFLOPS. The accuracy meanwhile improved from 86.63\% to 89.35\%.

Generally, the inefficiencies of \textit{width} (too high / low $s_{\mu}$) and \textit{depth} (tail pattern) can be considered independent, see Fig. \ref{fig:woof}.
Both models are trained on the ImageWoof-dataset, another 10-class subset of ImageNet, and the images were downsampled to $32 \times 32$ pixels. 
The resulting difference in saturation and predictive performance does not affect the number of inactive layers caused by the tail pattern.
Based on these observations, we conclude that width and depth can be treated as mostly independent factors when optimizing an architecture for a fixed input resolution, as long as the saturation is distributed similarly. 
For this reason, optimizing the depth should be done first to guarantee an even distribution of saturation.
Then the width can be scaled to put $s_{\mu}$ into the sweet-spot of $s_{\mu} \in (20\%, 30\%)$ (see Section \ref{sec:widt}). 

\section{Conclusion}
\label{conclusions}
In this work, we propose a novel, on-line computable metric ``saturation'' for analyzing neural networks layer by layer.
We base the development of saturation on the results of experiments that demonstrate the relevance of low-dimensional eigenspaces of the hidden layer output for the quality of the inference process.
We show that saturation yields reliable and reproducible results.
Over a series of experiments, we show that saturation can be used to detect two independent kinds of parameter-inefficiency in neural network architectures: Tail-patterns indicating unproductive layers, and over- or underparamterization that can be detected by the average saturation of the network.
Based on these findings, we propose simple strategies and guidelines to remove those inefficiencies from the neural architecture, gaining predictive performance and efficiency.

\bibliography{egbib}

\clearpage
\appendix

\title{Appendix}

\section{Experiments on the ablation of probe
performance due to downsampling strategies}
\label{apx:ablation}
As a concession to practicality, it is necessary to reduce the dimensionality of the feature maps to train logistic regression probes on the output of convolutional layers.
This problem is addressed by the original authors \cite{alain2016} with two fairly crude solutions. 
The first being a global average pooling on the feature map and the second one a random selection of positions on the feature map.
We reject the random strategy, since we want to avoid adding a random component to our measurements that may introduce noise or instability.
The global pooling strategy looks more promising to us, since global pooling is also performed on most recent architectures as an interface between convolutional and dense sections of the network~\cite{resnet,inceptionv3,efficentNet,widenets,wideresnet,vgg}.
However, global average pooling inside neural architectures is generally performed on the last layer's output, which can be expected to be fairly low dimensional on the height and width axis compared to earlier layers. Furthermore, due to the smaller receptive fields of earlier layers, the encoded information will be more local and thus more heterogeneous based on the position of the entry.
Since these circumstances are likely to negatively affect the performance and / or introduce artifacts into the probe performance measurements of early layers, we decide to look for less invasive downsampling strategies.

We test downsampling of feature maps using the nearest interpolation algorithm as well as adaptive average pooling to a smaller feature map size.
For testing, we use two models as test benches. First, the modified version of ResNet18 described in the last experiment of section \ref{sec:resolution}, which is trained on a $32 \times 32$ pixel resolution. 
Second, the original ResNet18 implementation trained on $224 \times 224$ pixel input resolution.
Both models are trained for 90 epochs using stochastic gradient descent with an initial learning rate of 0.1 and a momentum of 0.9. The learning rate is multiplied with 0.1 every 30 epochs.
These training setups both feature no tail pattern, which is important for testing the effect of the downsampling strategies, since we expect more aggressive downsampling to have a negative effect on probe performance.
Negative effects would be harder to interpret on late layers with a tail pattern, since a tail effectively means that the problem is already solved and the layers perform basically on the same performance level as the output.
We choose two input resolutions to observe the effects of downsampling on two different scales.

We train probes on feature maps reduced to a maximum of 1, 2, 3, 4, 5, 6 and 7 pixels in height and width using both average pooling and downsampling.

\begin{figure}[htb!]
	\centering
	\subfloat[Probes trained on downsampled feature maps. The model was trained on $32 \times 32$ input resolution.]{\includegraphics[width=0.6\columnwidth]{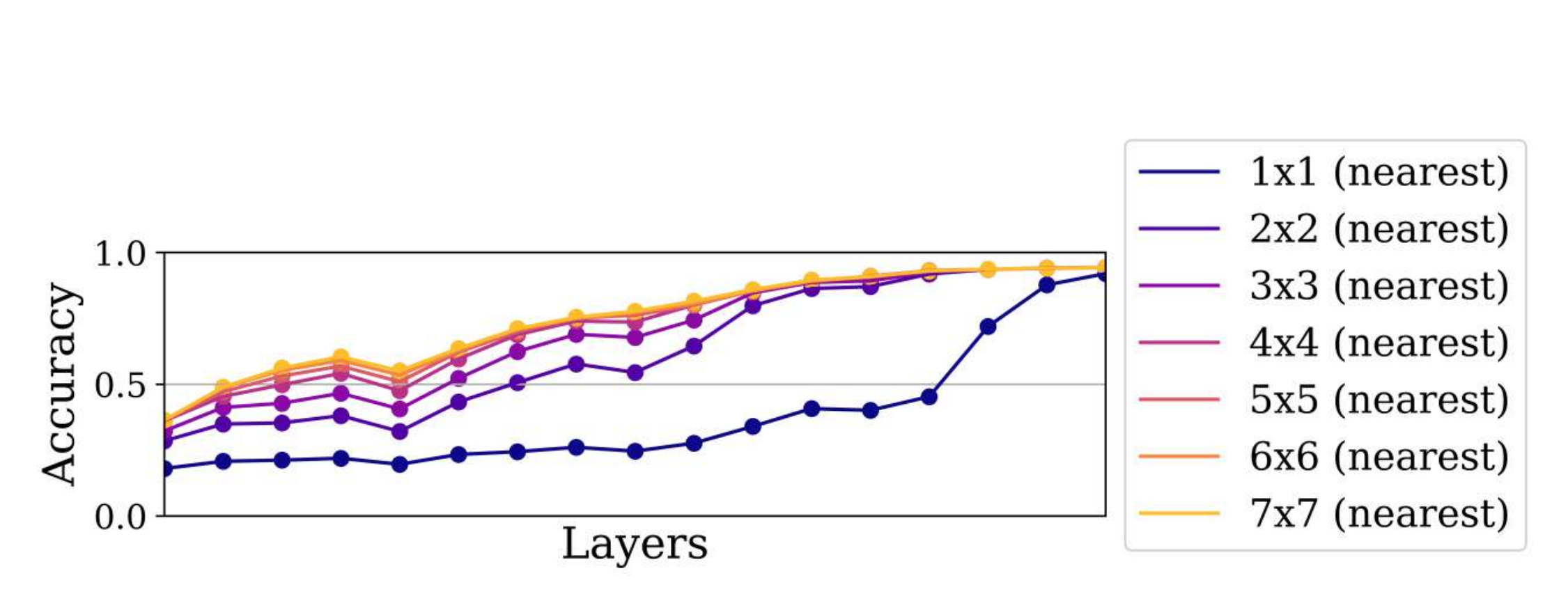}}\quad
	\subfloat[Probes trained on downsampled feature maps. The model was trained on $224 \times 224$ input resolution.]{\includegraphics[width=0.6\columnwidth]{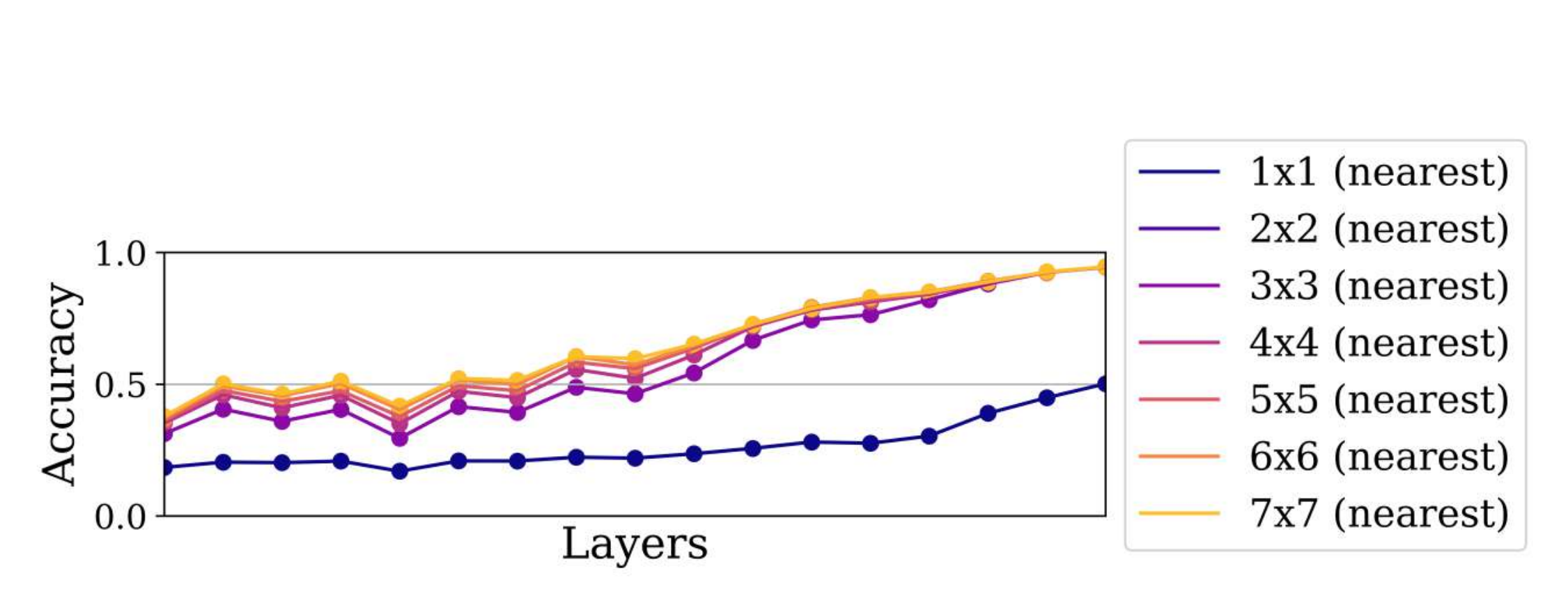}}\quad
	\subfloat[Probes trained on adaptive average pooled feature maps. The model was trained on $32 \times 32$ input resolution.]{\includegraphics[width=0.6\columnwidth]{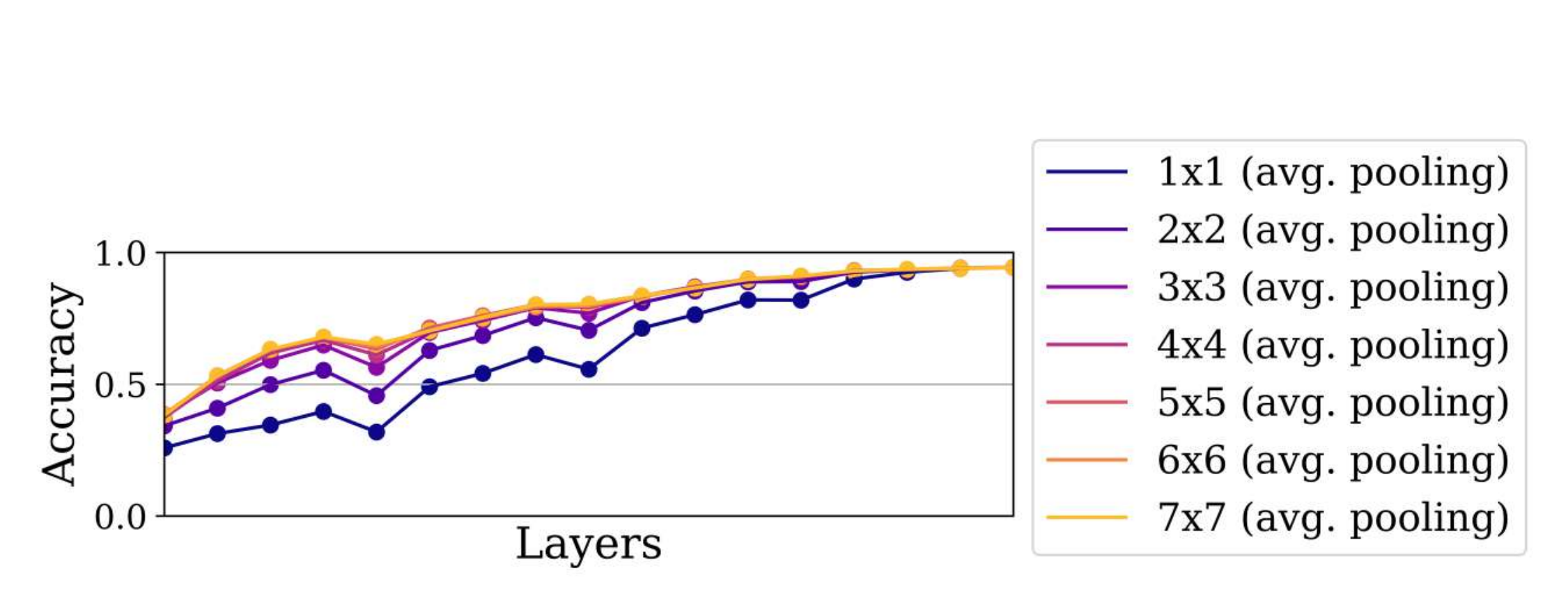}}\quad
	\subfloat[Probes trained on adaptive average pooled feature maps. The model was trained on $224 \times 224$ input resolution.]{\includegraphics[width=0.6\columnwidth]{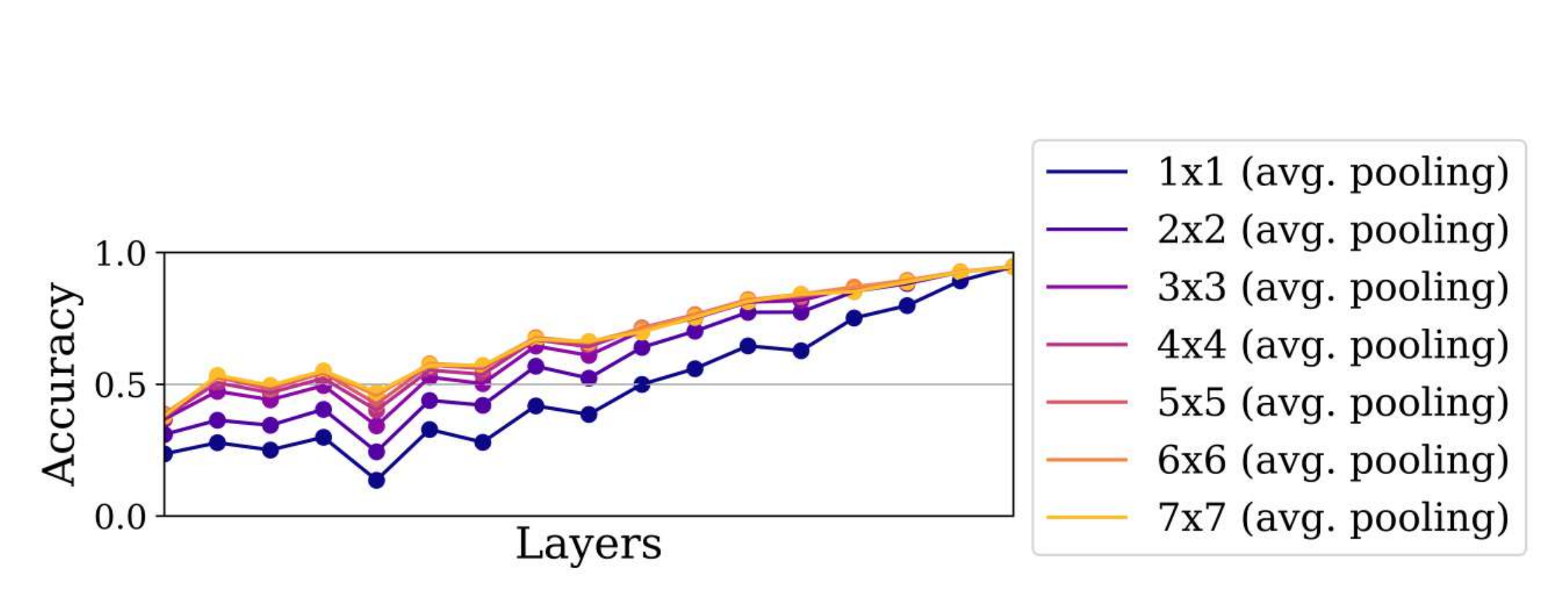}}\quad
	\vspace{0.3cm}
	\caption{As expected, more aggressive reduction in feature maps sizes negatively impact the probe performance. Average pooling seems to be able to maintain the structure of probe performances better. Downsampling to a single pixel induces heavy artifacts, destroying the otherwise prevalent pattern of probe performances.}
	\label{fig:ablation}
\end{figure}

The results can be seen in Fig.~\ref{fig:ablation}. On large feature map resolutions, we observe that both strategies produce very similar patterns. However, the structural integrity of those patterns is better maintained on average pooled feature maps. 
When reduced to a single depth-vector, the nearest-downsampling destroys the pattern otherwise visible on any other downsampling resolution.
Other than that, increasing the size of the reduced feature maps has a globally negative impact on probe performance. Earlier layers being affected worse from the decrease than later layers, which is expected for the aforementioned reasons.
Based on the results of these experiments, we decided to compute probe performances in this paper using feature maps adaptive average pooled to a resolution of $4\times 4$, which is computable with reasonable computational resources while maintaining the relative structure of the probe performances.

\clearpage

\section{Details on experimental setups}

\subsection{Details on experimental setups from experiments in section 3}

\subsubsection{Dataset and data augmentation}
The experiments are conducted on CIFAR10.
The images are channel-wise normalized with $\mu = (0.4914, 0.4822, 0.4465)$ and $\sigma = (0.2023, 0.1994, 0.2010)$.
At training time, the images are first cropped randomly with a 4 pixel zero-padding on all edges. The size of the crop is $32 \times 32$ pixels.
Then the crops are horizontally flipped randomly with a probability of 50\%.
The images of the training set are reshuffled after each epoch.

\subsubsection{Models}
The experiments use VGG11, 13, 16 and 19 as well as four additional variations of the aforementioned architectures.
The variations have all filter sizes reduced by a factor of 2, 4, 8 and 16.
Furthermore, the architectures are slightly modified by adding batch normalization layers after each convolutional layer. In addition, the flattening layer serving as the connection between convolution feature extractor and densely connected classifier is replaced by a global pooling layer.
The variations are included in order to not only include models of various depth, but also of varying width in our results. These are two common degrees of freedom in neural architecture design.
The modification to the architectures are made to include common architectural features that can be considered standard in most modern architectures.
Furthermore, the global pooling layer makes the models agnostic towards the input resolution.
This enables us to alter the input resolution without changing the number of parameters inside the model.
PCA-Layers for projecting the network are added after each convolutional and linear layer.

\subsubsection{Training setup and parameters}
We use the same training setup for all models we test in this chapter.
Since we are interested in in-development scenarios, we do not apply hyperparameter optimization.
Instead, we use default-values of PyTorch wherever possible and otherwise settings that are generally in the common range of hyperparameters used for similar classification tasks.
The exact hyperparameter settings are depicted in Table \ref{tab:hyper3}.
We find that 30 epochs is enough time for all models to converge to a stable solution on CIFAR10.

\begin{table}[htb!]
	\centering

	\begin{tabular}{ll}
		\toprule
		\textbf{Parameter}  & \textbf{Values} \\
		\midrule
		Epochs              & 30              \\
		Batch size          & 128             \\
		Optimizer           & ADAM            \\
		ADAM: beta1         & 0.9             \\
		ADAM: beta2         & 0.999           \\
		ADAM: epsilon       & 1e-8            \\
		ADAM: learning rate & 0.001           \\
		\bottomrule
	\end{tabular}
	\vspace{0.3cm}
	\caption{Hyperparameters common to each of the experiments in Section \ref{sec:layer-eigenspaces}}
	\label{tab:hyper3}
\end{table}

\subsubsection{Number of experiments conducted}
In the experiment in Fig.~\ref{fig:projections_both} a total of 60 models are trained.
Each neural architecture is trained 3 times using the same setup.
The experiments in Tables \ref{tab:VGG13_ttest_short} and \ref{tab:VGG19_ttest_short}  are repeated 26 and 40 times respectively on the same model using the same setup.
We also ran 15 additional experiments on ResNet18 and VGG11, similar to the aforementioned experiments.
The results of these are depicted in Table \ref{tab:vgg11ttest} and \ref{tab:resnet18ttest}.

\subsubsection{Details on PCA-Layer Projections}
The eigenvalues and eigenvectors of the output of all PCA-Layers are computed when the switch from training to testing occurs at the end of an epoch. At this point, the projection matrix is computed and the aggregation variables (running sum, running squares and number of seen samples) are reset in each PCA-Layer.
The PCA-Layers keep the last computed covariance matrix in memory as an internal variable.
This allows us to recompute the projection matrix $P_{E^k_l}$.

\subsection{Details on experimental setups from experiments in section \ref{saturatiion_exp} and \ref{sec:OptimizeNetworks}}

\subsubsection{Dataset}
The experiments conducted in Section \ref{saturatiion_exp} use models trained on CIFAR10, ImageWoof and ImageNette. 
We additionally reproduce some results on MNIST and TinyImageNet and compute the saturation levels of ResNet18 on ImageNet.
These results are not depicted in Section \ref{saturatiion_exp}; however, these results are included in Appendix \ref{add_results}.
We choose these datasets to test our hypothesis on different levels of complexity, regarding the number of classes as well as the natural resolution of the images.

The preprocessing and data augmentation is the same as in Section \ref{sec:layer-eigenspaces}.
The input resolution differs depending on the dataset and the running experiment. If not mentioned otherwise, the images are processed in their native resolution. MNIST data is additionally transformed into RGB to avoid changes in the neural architecture.
In any case, the resizing is performed after the augmentation pipeline is applied on a batch of images.

\subsubsection{Models}
The experiment uses the same models as in Section \ref{sec:layer-eigenspaces}. Additionally, we train ResNet18 and ResNet34 with filter sizes reduced by a factor of 1, 2, 4 and 8.
We include the ResNet architectures to include another architecture, with a feature (skip connections), that may affect how the information is flowing though the network.
We also remove the skip-connections on ResNet18 and 34 for two experiments to observe the effects of disabled skip connections.
Different from previously described experiments, none of these architectures have PCA-Layers.

\subsubsection{Training setup and parameters}
We choose a static training setup of all models and datasets, with the same reasoning as in the experiments conducted in Section \ref{sec:layer-eigenspaces}.
Compared to the setup described of the experiments in Section \ref{sec:layer-eigenspaces}, the batch size is changed to 32 (16 in the cases of ResNet34, VGG16 and 19) due to memory limitations.
However, we find through brief exploration that slight changes in the hyperparameter optimization described here as well as additional epochs of training do not influence the results described in Section \ref{saturatiion_exp} in any meaningful way.

\begin{table}[htb!]
	\centering 
	\begin{tabular}{ll}
		\toprule
		\textbf{Parameter}  & \textbf{Values}                       \\
		\midrule
		Epochs              & 30                                    \\
		Batch size          & 32 (16 for ResNet34, VGG16 and VGG19) \\
		Optimizer           & ADAM                                  \\
		ADAM: beta1         & 0.9                                   \\
		ADAM: beta2         & 0.999                                 \\
		ADAM: epsilon       & 1e-8                                  \\
		ADAM: learning rate & 0.001                                 \\
		\bottomrule
	\end{tabular}
	\label{tab:hyper4}
	\vspace{0.3cm}
	\caption{Hyperparameters common to each of the experiments in Section \ref{saturatiion_exp}}
\end{table}

\subsubsection{Probe setup}
The data used for training the probes is extracted after the final epoch of training.
In case of fully connected layers, the data is simply aggregated and saved as a single data matrix of shape $(samples \times \#neurons)$.
In case of convolutional layers, this is neither practical nor possible due to limitations in hardware.
Alain and Bengio \cite{alain2016} propose Global Pooling or a random selection of features to bypass this issue.
However, we are afraid that global pooling the entire feature map can potentially bias the data and thus the probe performance as a result.
To mitigate this potential bias, we only adaptive average pool the feature map to $(4 \times 4 \times \#filters)$.
The reduced feature maps are then flattened into a vector and stored as a data matrix of shape $(samples \times 4^2 \cdot \#filters)$.
We are aware that this method still is not free of ablation. 
We study the effects of different downsampling techniques in appendix \ref{apx:ablation}. The result show that this approach has a tolerable impact for our purposes, since the visual difference in the structure probe performances is very slim compared to less aggressive downsampling strategies.
The training test split remains unchanged from the original data.
The probes are logistic regression classifiers minimizing cross entropy using the SAGA solver implementation of scikit-learn.
The logistic regression is fitted for 100 epochs.

\section{Additional Results}
\label{add_results}
In this section, we present additional results and insights from the experiments presented in section \ref{saturatiion_exp} and \ref{sec:OptimizeNetworks}.

\subsection{Feature map downsampling}
The resolution of the feature map has a multiplicative effect on the number of computations required for updating the covariance matrix.
Another problem is that the early convolutional layers yield more data points than later layers for computing the covariance matrix, since their feature map is larger.
To address both issues, we experiment with downsampling the feature maps using the nearest interpolation.
We find that downsampling feature maps such that the resolution of the feature map never exceed $32 \times 32$ a pixel did not visibly change the saturation pattern.
We did not apply this method in any of the experiments, since we did not explore the biases induced by this method enough.
We include this section only to mention that this a potential path for making the computations more efficient.

\subsection{On-line covariance computation and floating point precision}
Another issue when on-line computing a covariance matrix is the precision of floating-point values.
Neural networks are generally processed in full precision.
However, for large amounts of data, the compounding round-off errors induced by the 32-bit precision of the variables may induce errors.
For this reason, all computations concerning saturation are performed in double precision.
This is also true for the PCA-Layers.
Before the update of the covariance matrix is performed, the data is cast in double precision.
The running sum, running squares are double precision float arrays as well.

\subsection{Effect of eigenspace projections on the reconstruction of a convolutional autoencoder}
\label{additional-figures}
To visualize the effect of projection into the eigenspace, we train an autoencoder on the Food101 dataset.

\subsubsection{Convolutional autoencoder architecture}

\begin{table}[htb!]
	\centering
	\begin{tabular}{ll}
		\toprule
		\textbf{Encoder}                                   & \textbf{Decoder}                                   \\
		\midrule
		512 $\times$ 512 $\times$ 3 Input                  & $(3\times 3)$ conv, 8 ReLU           \\
		$(3\times 3)$ conv, 16 filters, ReLU & upsampling, nearest, scale-factor 2                \\
		$(2\times 2)$ max pooling, strides 2               & $(3\times 3)$ conv, 8 filters, ReLU  \\
		$(3\times 3)$ conv, 8 filters, ReLU  & upsampling, nearest, scale-factor 2                \\
		$(2\times 2)$ max pooling, strides 2               & $(3\times 3)$ conv, 16 filters, ReLU \\
		$(3\times 3)$ conv, 8 filters, ReLU  & upsampling, nearest, scale-factor 2                \\
		$(2\times 2)$ max pooling, strides 2               & $(3\times 3)$ conv, 3 filters, ReLU  \\
		\bottomrule
	\end{tabular}
	\vspace{0.3cm}
	\caption{Convolutional Autoencoder Architecture. All convolutional Layers use same-padding}
	
	\label{tab:cae_arc}
\end{table}
\clearpage
\subsubsection{Hyperparameters}
\begin{table}[htb!]
	\centering
	\begin{tabular}{ll}
		\toprule
		\textbf{Parameter}  & \textbf{Values}    \\
		\midrule
		Input Resolution    & $(224 \times 224)$ \\
		Epoch               & 50                 \\
		Batch size          & 128                \\
		Optimizer           & ADAM               \\
		ADAM: beta1         & 0.9                \\
		ADAM: beta2         & 0.999              \\
		ADAM: epsilon       & 1e-8               \\
		ADAM: learning rate & 0.0001             \\
		\bottomrule
	\end{tabular}
	\vspace{0.3cm}
	\caption{Hyperparameters for the convolutional autoencoder.}
	\label{tab:my_label}
\end{table}

\clearpage

\subsubsection{Reconstruction examples for different values of {$\delta$}{}}\label{appendix:cae}
To test the effects of convolutional and linear projections.

\begin{table}[htb!]
	\centering
	\label{tab:test1}
	\begin{tabular}
		{p{0.8in}p{0.8in}p{0.4in}p{0.8in}} 
		\hline
		$\delta$   & loss  & Reconstruction \\
		\hline \textbf{Original} & -     & \parbox[c]{1em}{
		\includegraphics[width=0.91in]{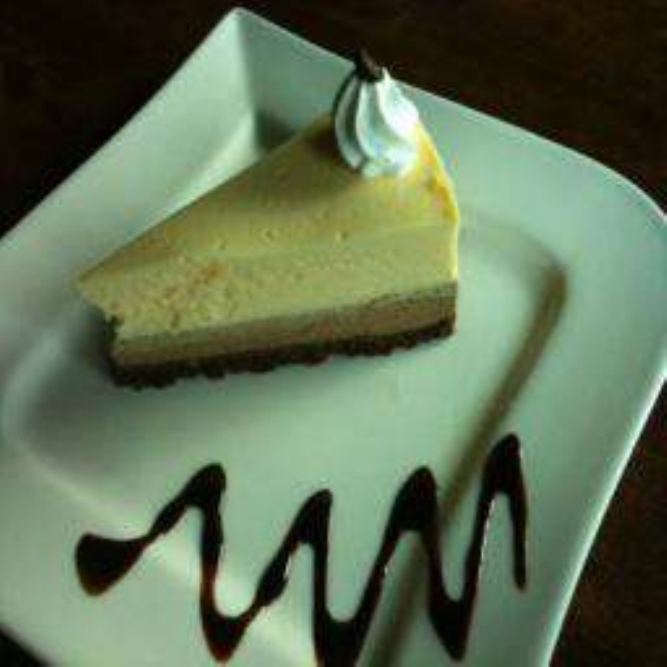}} \\
		\hline \textbf{100\%}    & 0.033 & \parbox[c]{1em}{
		\includegraphics[width=0.91in]{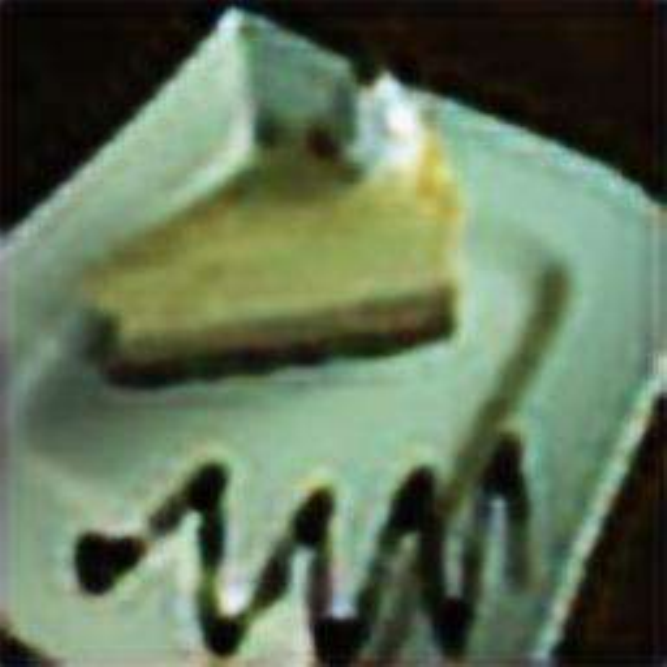}} \\
		\hline \textbf{99.9\%}   & 0.065 & \parbox[c]{1em}{
		\includegraphics[width=0.91in]{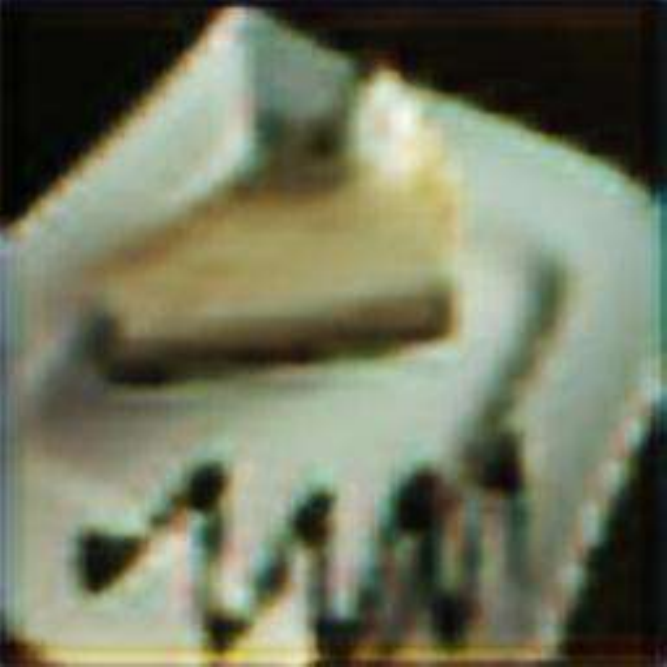}}  \\
		\hline \textbf{99.5\%}   & 0.089 & \parbox[c]{1em}{
		\includegraphics[width=0.91in]{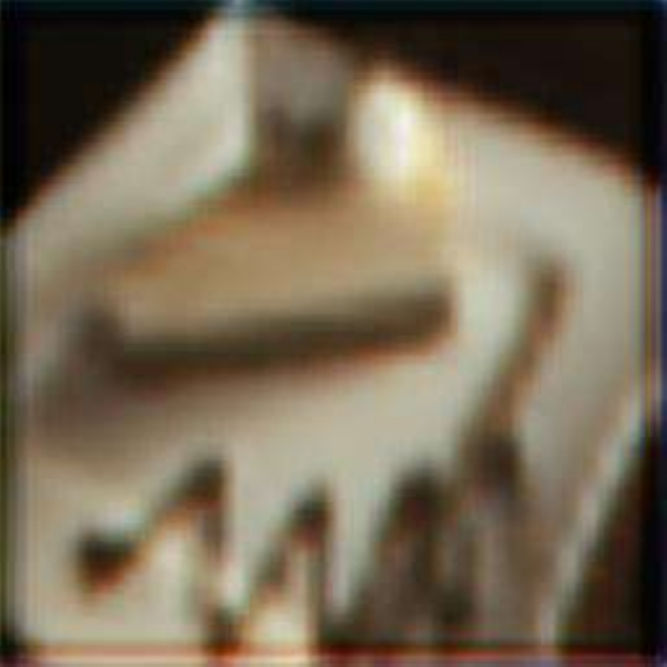}}  \\
		\hline \textbf{99\%}     & 0.120 & \parbox[c]{1em}{
		\includegraphics[width=0.91in]{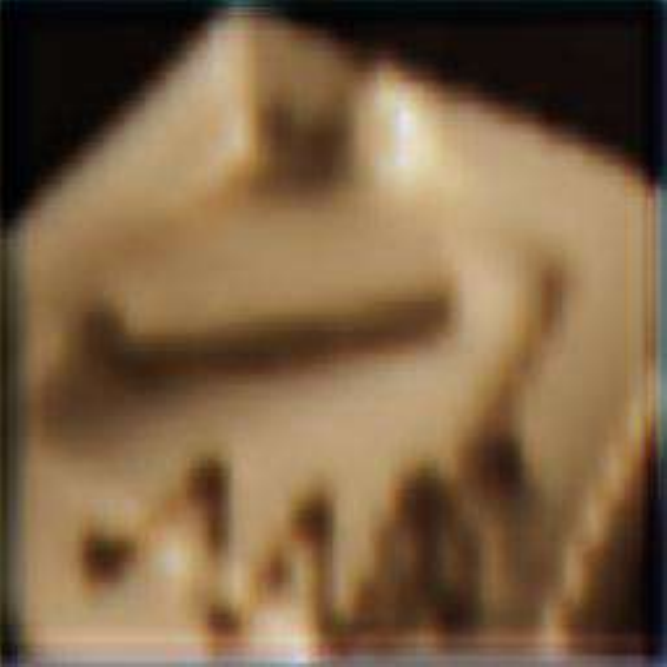}}  \\
		\hline \textbf{95\%}     & 0.216 & \parbox[c]{1em}{
		\includegraphics[width=0.91in]{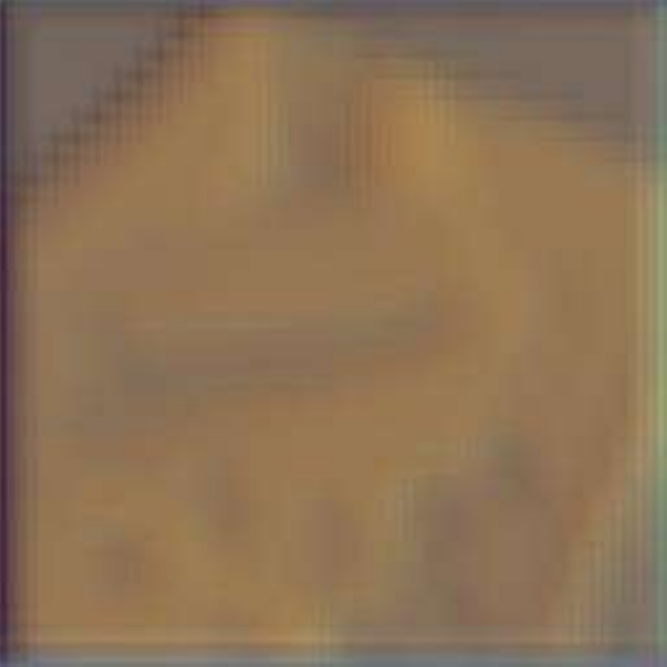}}  \\
		\hline \textbf{90\%}     & 0.234 & \parbox[c]{1em}{
		\includegraphics[width=0.91in]{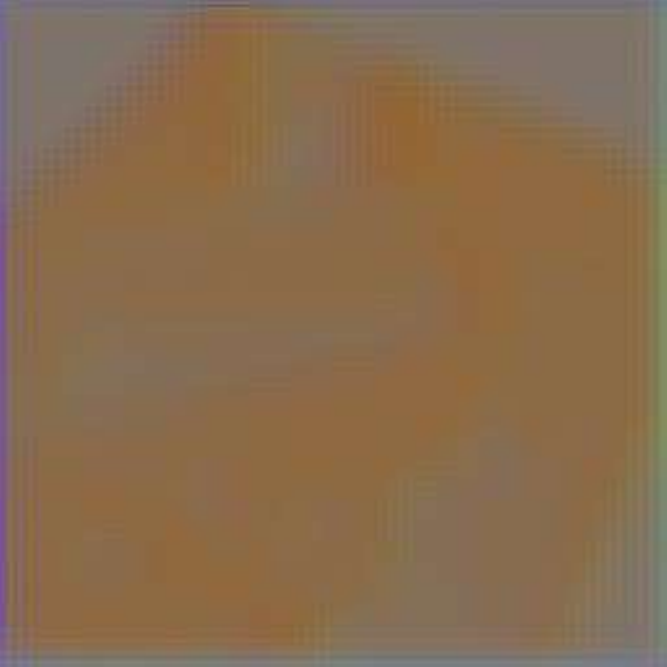}}  \\
	\end{tabular}
	\vspace{0.3cm}
	\label{tab:cae_conv_reconstruction_bottleneck}
	\caption{Reconstruction examples for different values of $\delta$. PCA is applied on all convolutional layers. }
\end{table}

\begin{table}[htb!]
	\centering
	\begin{tabular}
		{p{0.8in}p{0.8in}p{0.8in}p{0.8in}p{0.8in}}
		\hline $\delta$  & $dim \, E^k_{encoding}$ & loss  & Reconstruction               \\
		\hline \textbf{Original} & -                       & -     & \parbox[c]{1em}{
		\includegraphics[width=1in]{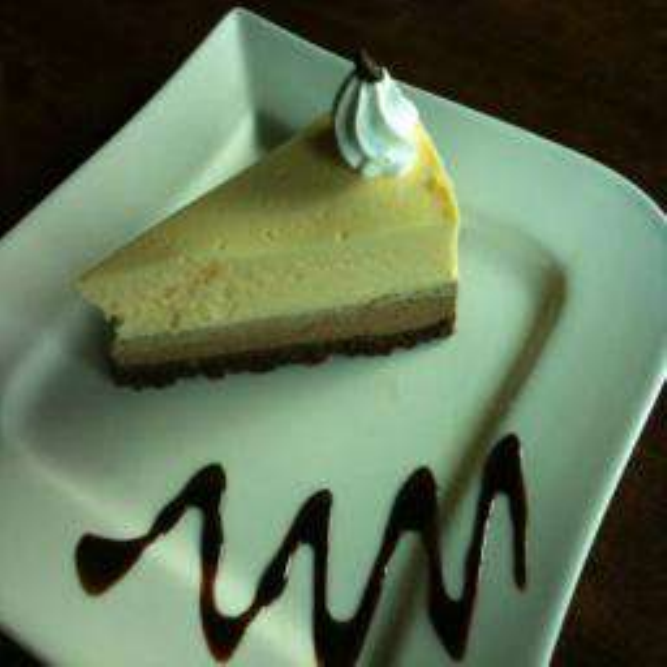}}     \\
		\hline \textbf{100\%}    & 8192                    & 0.033 & \parbox[c]{1em}{
		\includegraphics[width=1in]{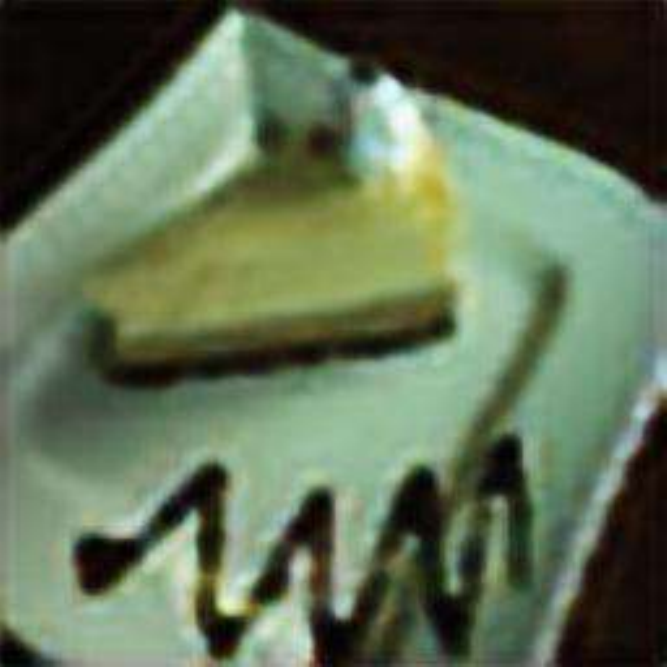}} \\
		\hline \textbf{99.9\%}   & 4374                    & 0.035 & \parbox[c]{1em}{
		\includegraphics[width=1in]{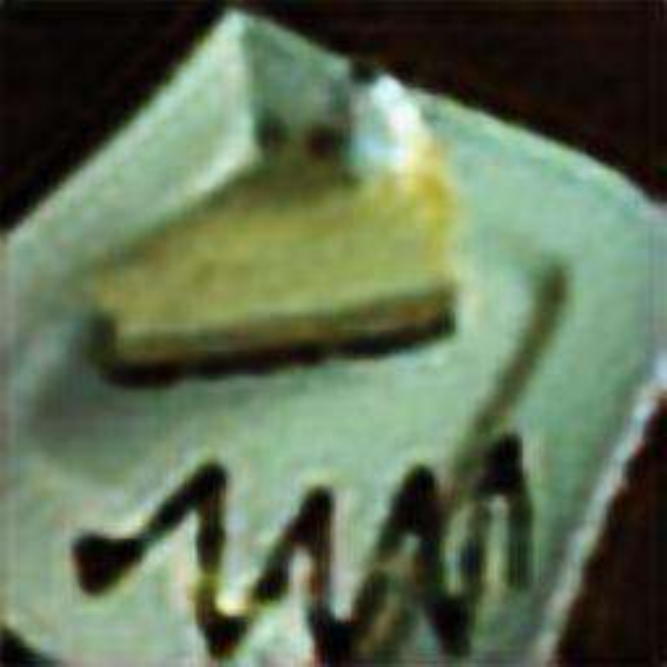}}  \\
		\hline \textbf{99.5\%}   & 1332                    & 0.049 & \parbox[c]{1em}{
		\includegraphics[width=1in]{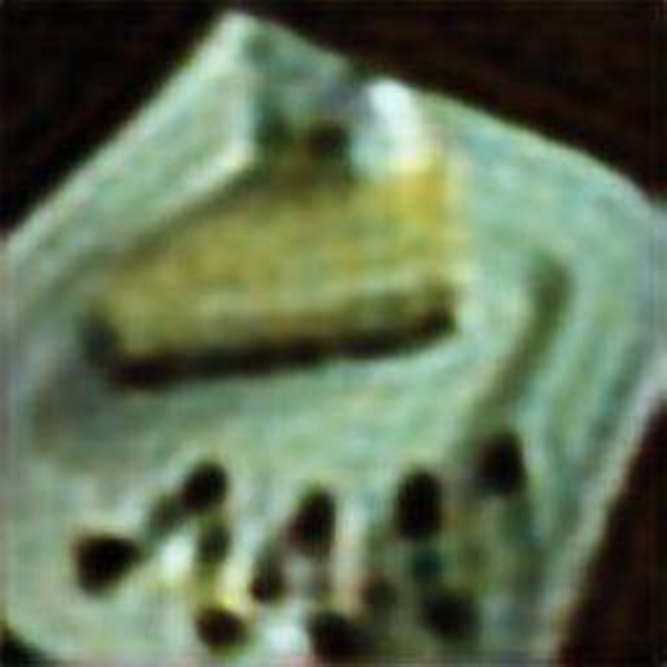}}  \\
		\hline \textbf{99\%}     & 597                     & 0.062 & \parbox[c]{1em}{
		\includegraphics[width=1in]{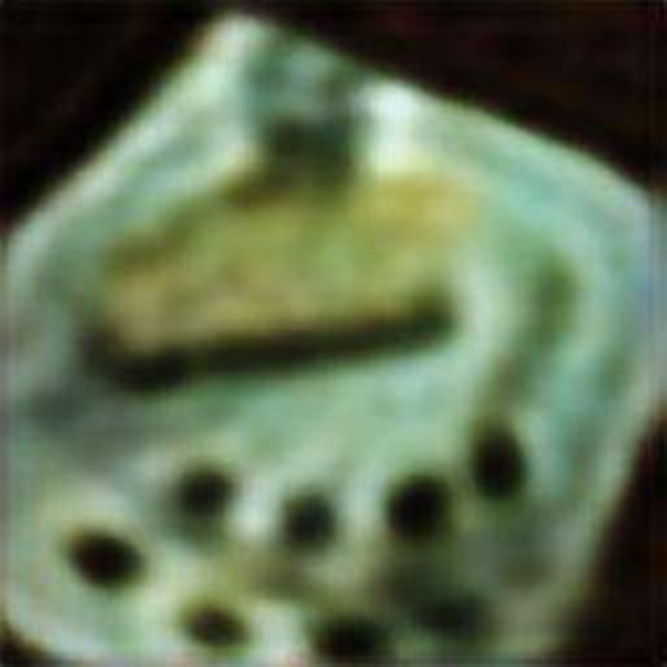}}    \\
		\hline \textbf{95\%}     & 17                      & 0.148 & \parbox[c]{1em}{
		\includegraphics[width=1in]{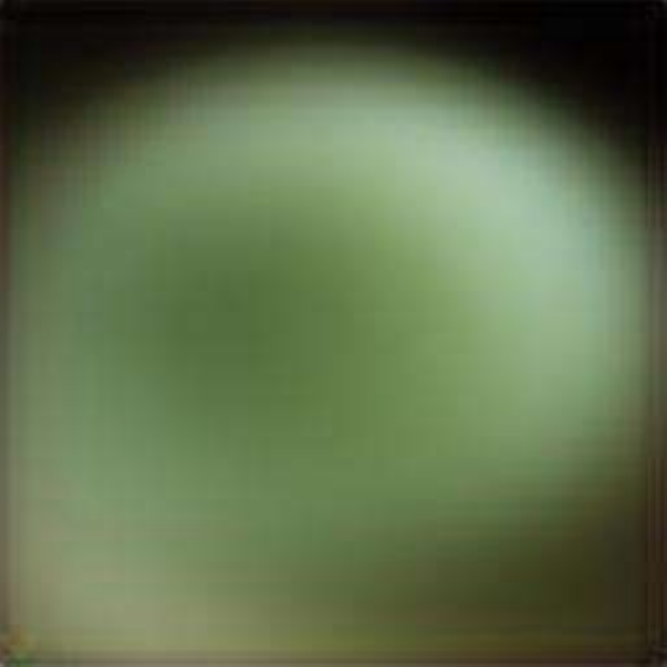}}     \\
		\hline \textbf{90\%}     & 1                       & 0.222 & \parbox[c]{1em}{
		\includegraphics[width=1in]{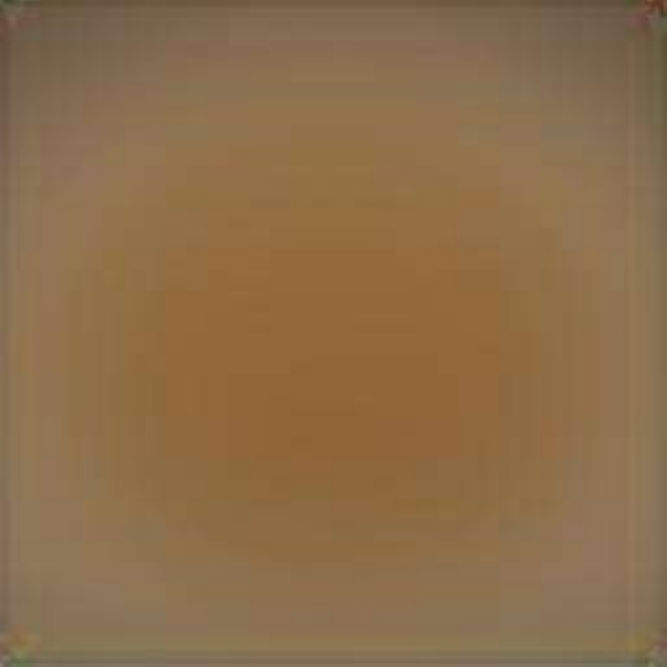}}      \\
	\end{tabular}
	\vspace{0.3cm}
	\caption{Reconstruction examples for different values of $\delta$. PCA  is applied on the fully connected encoding layer.}
	\label{tab:test2}
\end{table}

\clearpage

\onecolumn
\subsection{Probe performances and saturation patterns of ResNet18 and 34 with disabled skip-connections trained on CIFAR10}
\label{apx:noskip_tails}

\begin{figure*}[htb!]
	\centering
	\includegraphics[width=\textwidth]{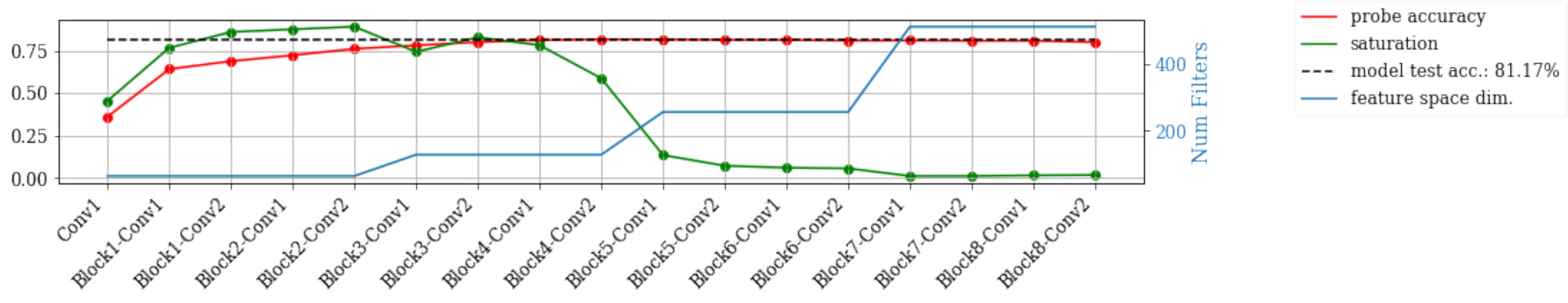}
	\vspace{0.3cm}
	\caption{ResNet18 without Skip connections trained on CIFAR10 with $32 \times 32$ pixel input resolution}
	\label{fig:18noskip}
\end{figure*}

\begin{figure*}[htb!]
	\centering
	\includegraphics[width=\textwidth]{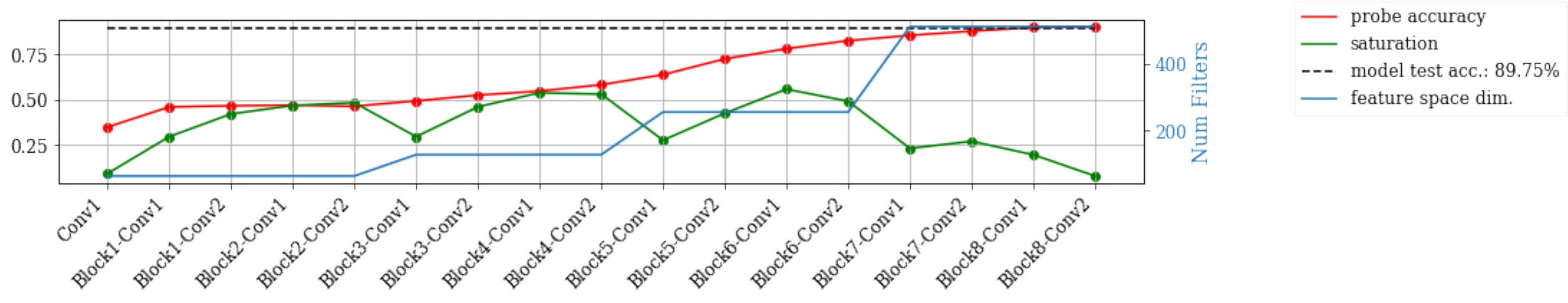}
		\vspace{0.3cm}

	\caption{ResNet18 without Skip connections trained on CIFAR10 with $224 \times 224$ pixel input resolution}
	\label{fig:18noskipBigCifar}
\end{figure*}

\begin{figure*}[htb!]
	\centering
	\includegraphics[width=\textwidth]{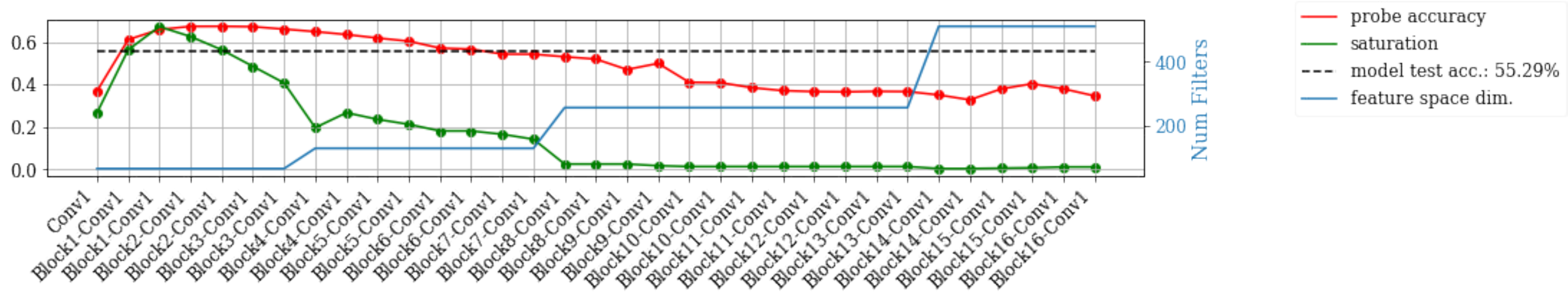}
	\vspace{0.3cm}
	\caption{ResNet34 without Skip connections trained on CIFAR10 with $32 \times 32$ pixel input resolution}
	\label{fig:34noskip}
\end{figure*}

\begin{figure*}[htb!]
	\centering
	\includegraphics[width=\textwidth]{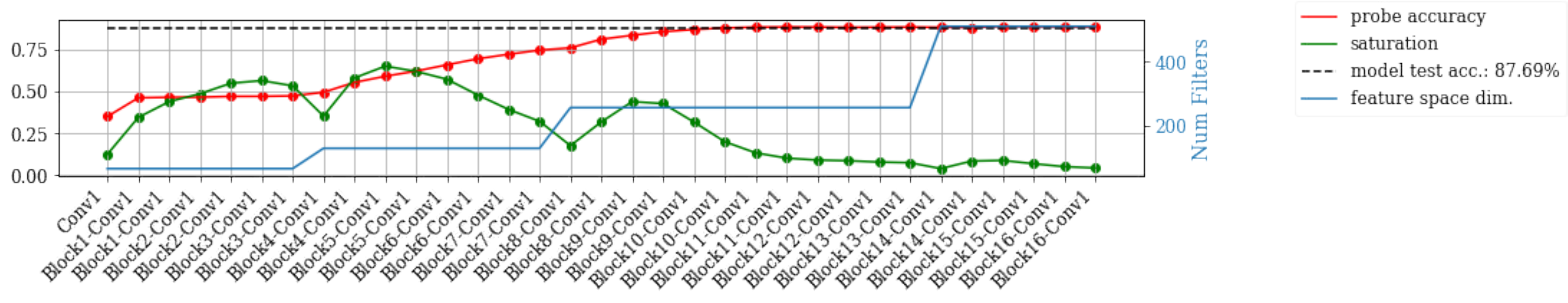}
	\vspace{0.3cm}
	\caption{ResNet34 without Skip connections trained on CIFAR10 with $224 \times 224$ pixel input resolution}
	\label{fig:34noskipBigCifar}
\end{figure*}

\clearpage

\subsection{Probe performances and saturation patterns for TinyImageNet}
\label{apx:tinyimgnet_tails}

\begin{figure*}[htb!]
	\centering
	\includegraphics[width=\textwidth]{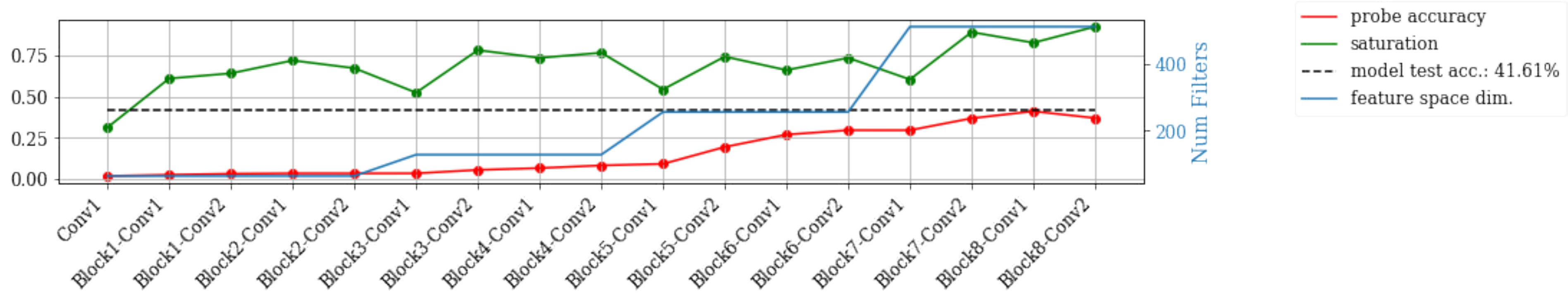}
	\vspace{0.3cm}
	\caption{ResNet18 trained on TinyImageNet with $64 \times 64$ pixel input resolution}
	\label{fig:18Tiny}
\end{figure*}

\begin{figure*}[htb!]
	\centering
	\includegraphics[width=\textwidth]{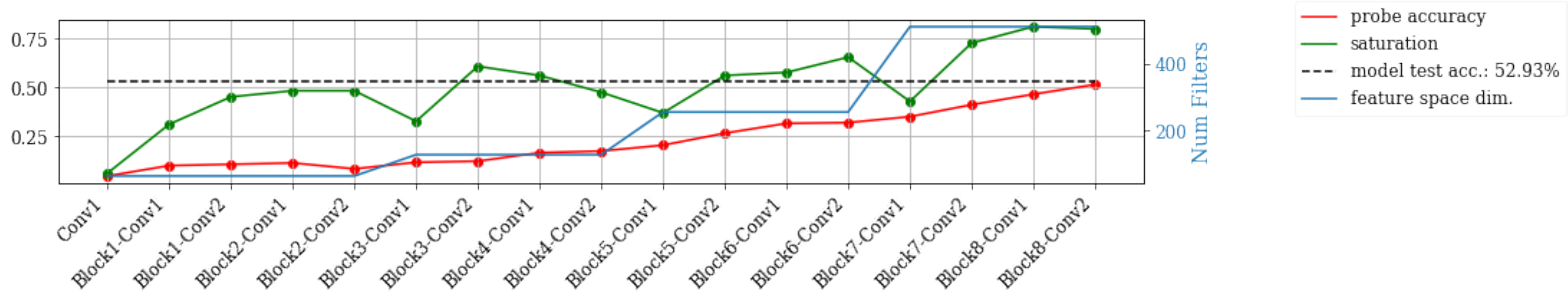}
	\vspace{0.4cm}
	\caption{ResNet18 trained on TinyImageNet with $224 \times 224$ pixel input resolution}
	\label{fig:18BigTiny}
\end{figure*}

\begin{figure*}[htb!]
	\centering
	\includegraphics[width=\textwidth]{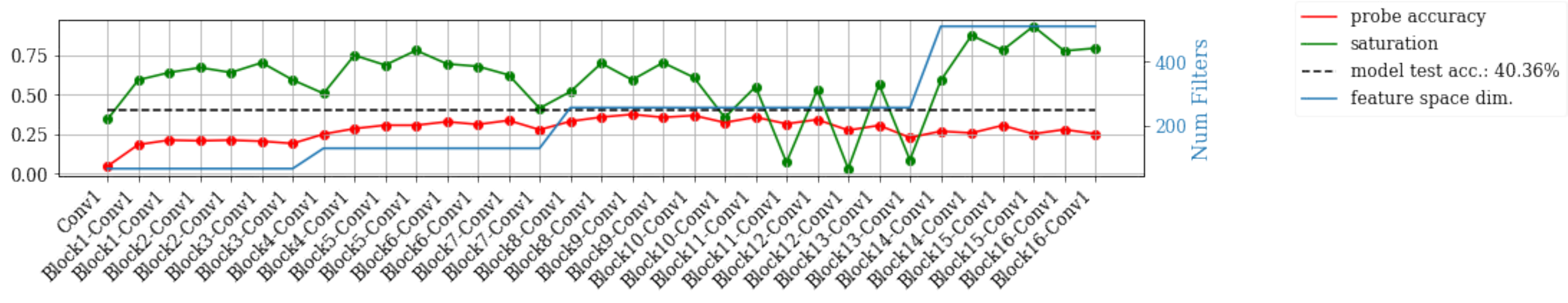}
	\vspace{0.3cm}
	\caption{ResNet34 trained on TinyImageNet with $64 \times 64$ pixel input resolution}
	\label{fig:34_Tiny}
\end{figure*}

\begin{figure*}[htb!]
	\centering
	\includegraphics[width=\textwidth]{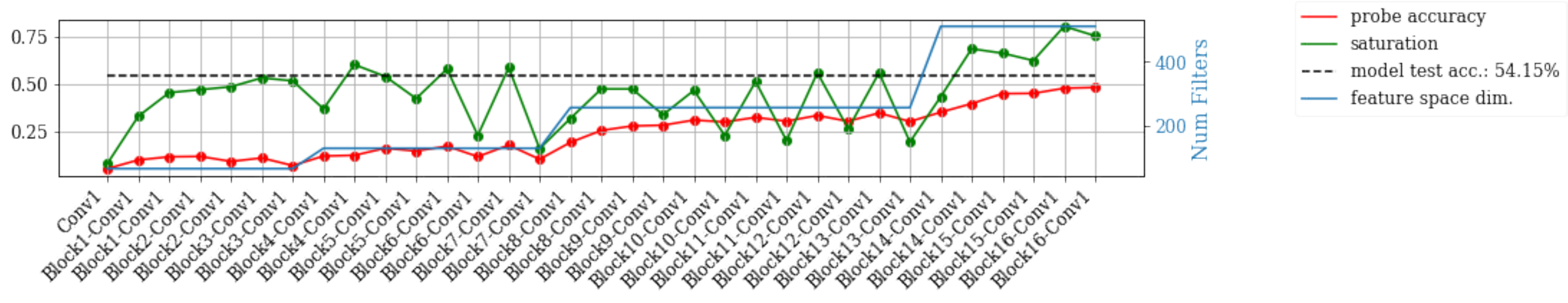}
	\vspace{0.3cm}
	\caption{ResNet34 trained on TinyImageNet with $224 \times 224$ pixel input resolution}
	\label{fig:34_BigTiny}
\end{figure*}

\clearpage

\subsection{Probe performances and saturation patterns for MNIST}
\label{apx:mnist_tails}

\begin{figure*}[htb!]
	\centering
	\includegraphics[width=\textwidth]{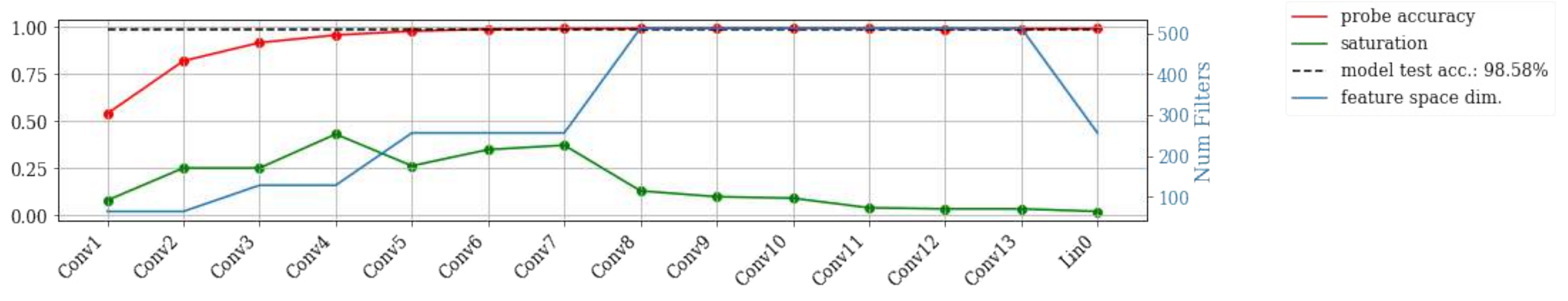}
	\vspace{0.3cm}
	\caption{VGG16 trained on MNIST width $32 \times 32$ pixel input resolution}
	\label{fig:16_MNIST}
\end{figure*}

\begin{figure*}[htb!]
	\centering
	\includegraphics[width=\textwidth]{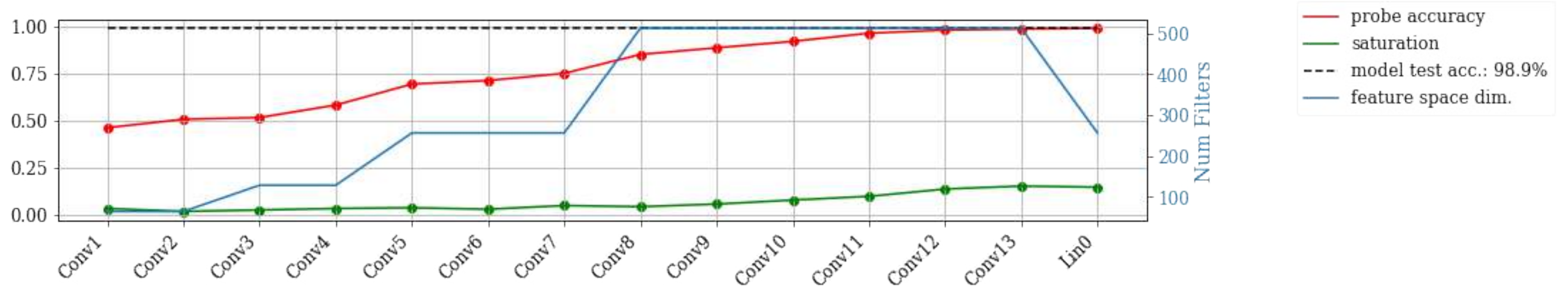}
	\vspace{0.3cm}
	\caption{VGG16 trained on MNIST width $224 \times 224$ pixel input resolution}
	\label{fig:16_big}
\end{figure*}

\begin{figure*}[htb!]
	\centering
	
	\includegraphics[width=\textwidth]{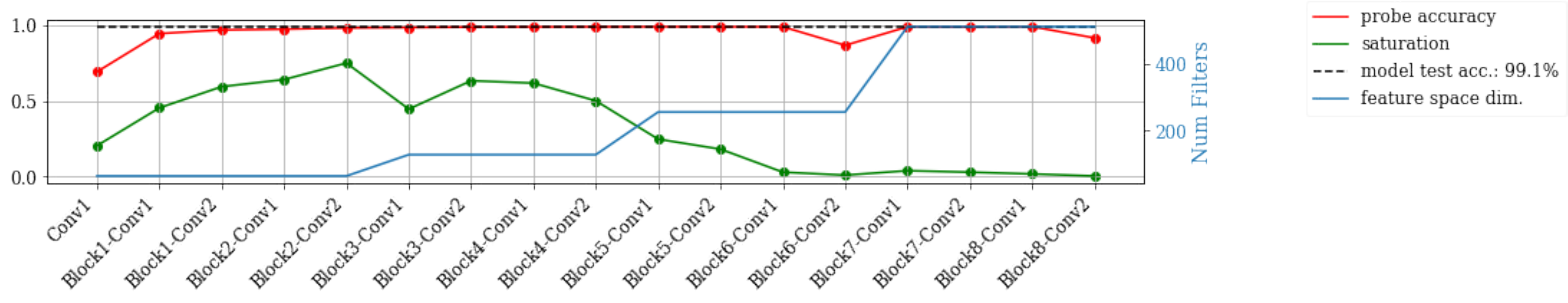}
	\vspace{0.3cm}
	\caption{ResNet18 trained on MNIST width $32 \times 32$ pixel input resolution}
	\label{fig:18_MNIST}
\end{figure*}

\begin{figure*}[htb!]
	\centering
	\includegraphics[width=\textwidth]{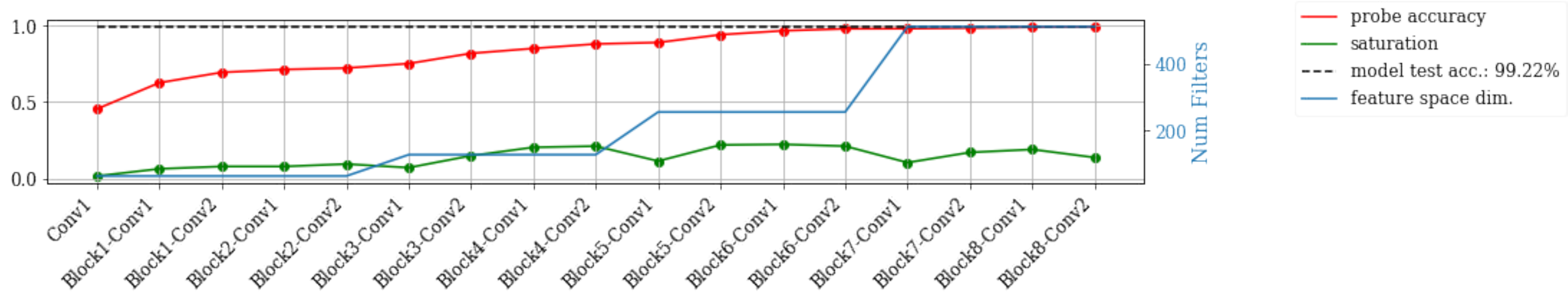}
	\vspace{0.3cm}
	\caption{ResNet18 trained on MNIST width $224 \times 224$ pixel input resolution}
	\label{fig:18_BigMNIST}
\end{figure*}

\begin{figure*}[htb!]
	\centering
	\includegraphics[width=\textwidth]{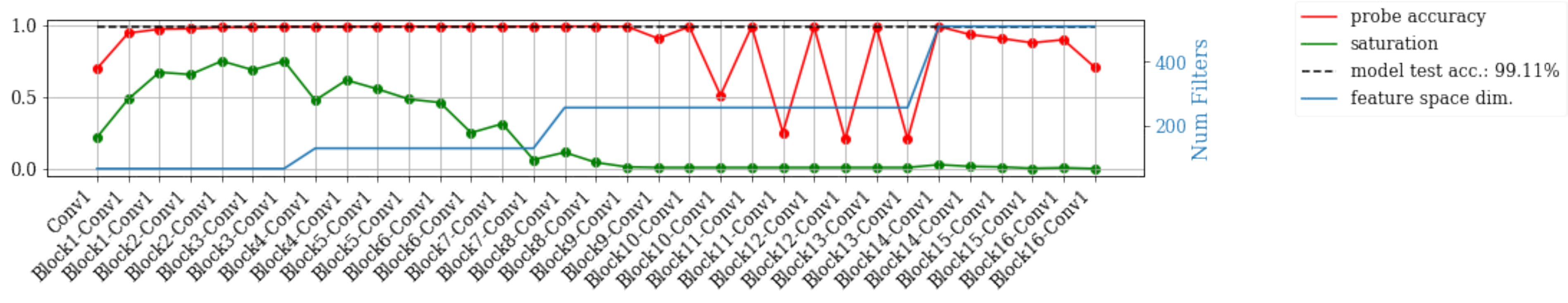}
	\vspace{0.3cm}
	\caption{ResNet34 trained on MNIST width $32 \times 32$ pixel input resolution}
	\label{fig:34_MNIST}
\end{figure*}

\begin{figure*}[h!]
	\centering
	\includegraphics[width=\textwidth]{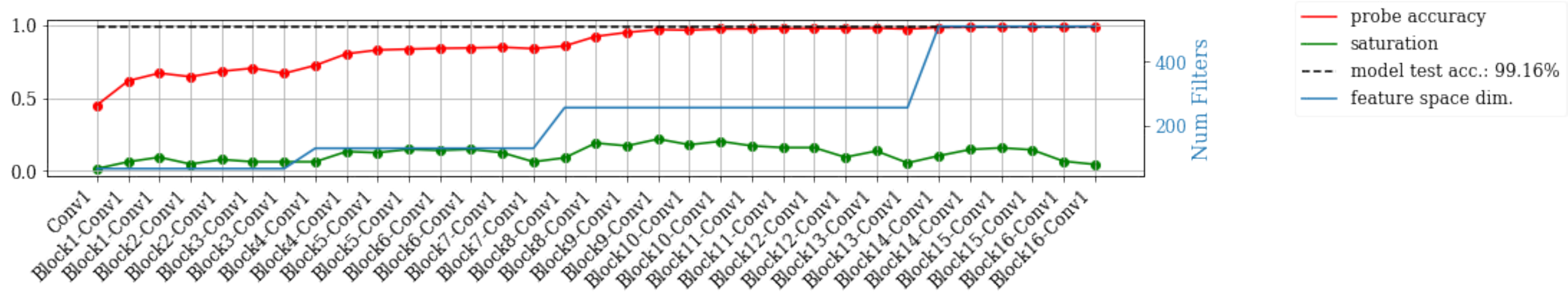}
	\vspace{0.3cm}
	\caption{ResNet34 trained on MNIST width $224 \times 224$ pixel input resolution}
	\label{fig:34_BigMNIST}
\end{figure*}

\clearpage

\subsection{Collages of VGG and ResNet-style models}
\label{apx:collages_tails}

\begin{figure*}[htb!]
	\centering
	\includegraphics[width=1.0\textwidth]{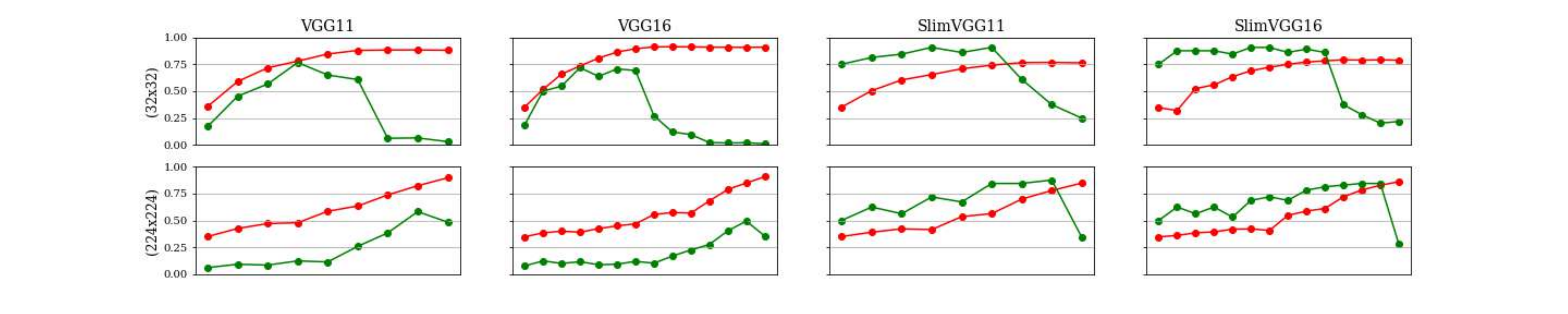}
	\vspace{0.3cm}
	\caption{Similar VGG-style models trained on CIFAR10. The model are altered in depth, filter size and input size. Their basic architecture however stays the same. The slim version of of the models has all filter sizes reduced by a factor of 8.}
	\label{fig:colVgg}
\end{figure*}

\begin{figure*}[htb!]
	\centering
	\includegraphics[width=1.0\textwidth]{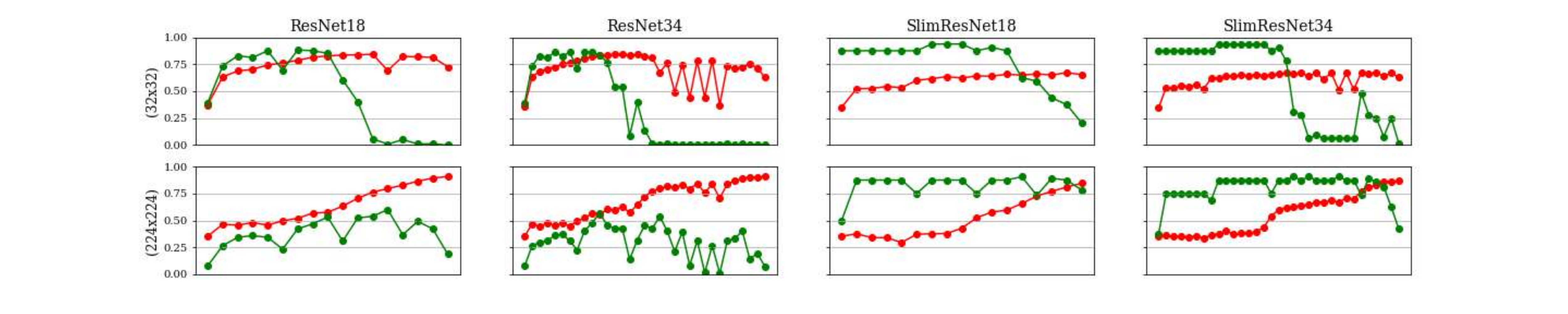}
	\vspace{0.3cm}
	\caption{Similar ResNet-style models trained on CIFAR10. The models are altered in depth, filter size and input size. Their basic architecture however stays the same. The slim version of the models has all filter sizes reduced by a factor of 8.}
	\label{fig:colResNet}
\end{figure*}

\begin{figure*}[htb!]
	\centering
	\includegraphics[scale=0.3]{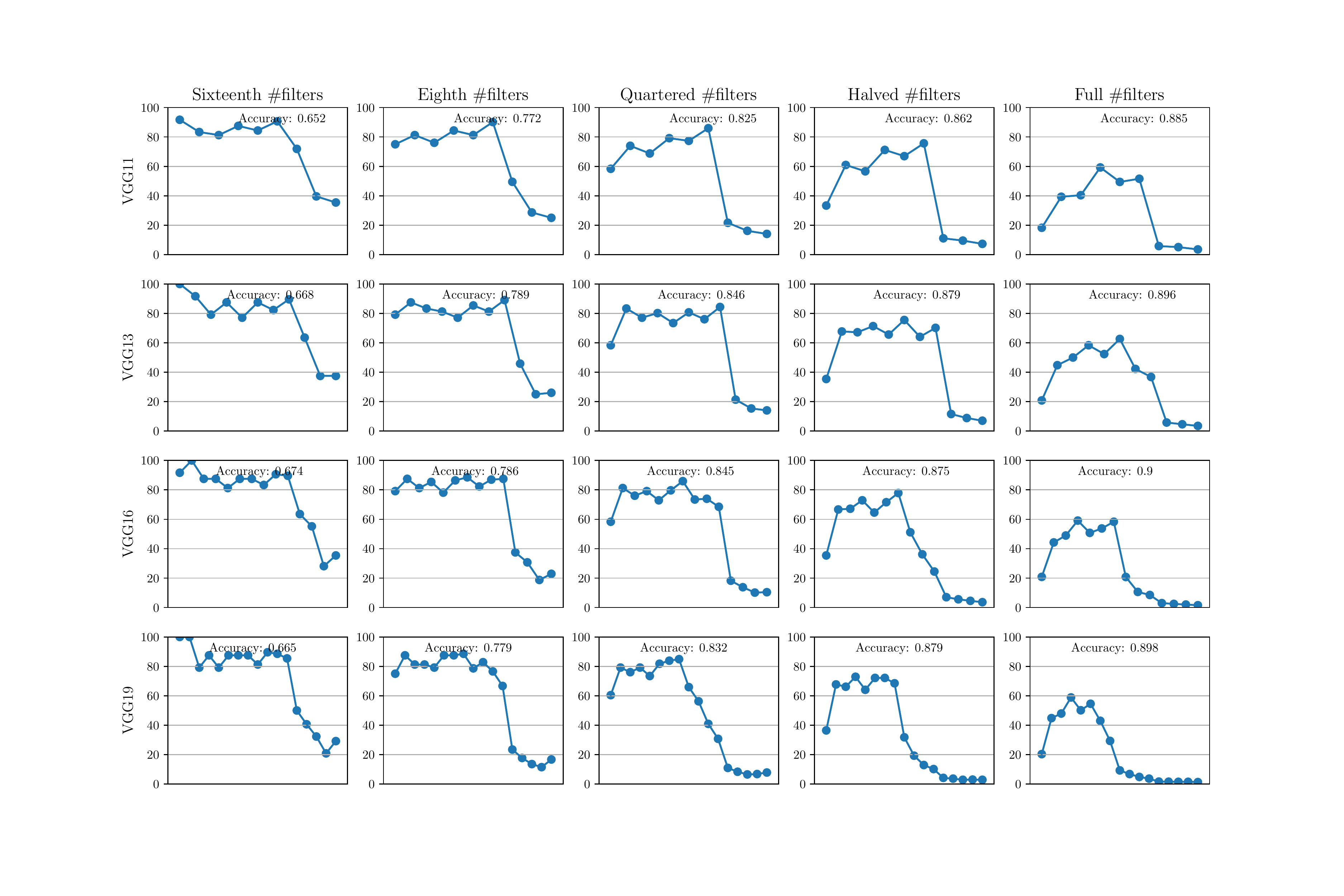}
	\vspace{0.3cm}
 	\caption{Layer-wise saturation of all tested CNN architectures trained on CIFAR10 for 30 epochs in the configuration for Section \ref{projecting_output}. The input resolution is $(32 \times 32)$ pixels for all models. The layers are represented on the x-axis in the same sequence as the information is propagated through the network at inference time. The y-axis describes the saturation value. Note the gradual change in the distribution while the number of filters and depth increase.}
	\label{fig:collage_cd}
\end{figure*}

\clearpage

\subsection{Full \textit{t}-Test tables of VGG11, VGG13, VGG16 and ResNet18}
\label{t-test-appendix}

\begin{table*}[htb!]
	\begin{center}
		\begin{tabular*}{\textwidth}{c @{\extracolsep{\fill}} c c c c c }
			\toprule
			Explained $\sigma$ & Mean difference & sample $\sigma$ & t-stat & p-value \\
			\midrule
			0.9999 & -0.0001  & 0.0007 & -0.714 & 0.487 \\
			0.9998 & -0.0005  & 0.0008 & -2.44 & 0.029 \\
			0.9997 & -0.0007  & 0.0012 & -2.07 & 0.058 \\
			0.9996 & -0.0005  & 0.0018 & -1.1 & 0.292 \\
			0.9995 & -0.0010  & 0.0018 & -2.16 & 0.049 \\
			0.9994 & -0.0008  & 0.0016 & -1.8 & 0.094 \\
			0.9993 & -0.0010  & 0.0015 & -2.72 & 0.017 \\
			\textbf{0.9992} & -0.0013  & 0.0014 & -3.65 & 0.003 \\
			0.9991 & -0.0011  & 0.0018 & -2.51 & 0.025 \\
			0.999 & -0.0015  & 0.0021 & -2.74 & 0.016 \\
			0.998 & -0.0019  & 0.0029 & -2.49 & 0.026 \\
			0.997 & -0.0013  & 0.0034 & -1.46 & 0.166 \\
			0.996 & -0.0009  & 0.0039 & -0.888 & 0.390 \\
			0.995 & 0.0005  & 0.0037 &  0.537 & 0.600 \\
			0.994 & 0.0022  & 0.0038 &  2.2 & 0.045 \\
			\textbf{0.993} & 0.0056  & 0.0071 &  3.03 & 0.009 \\
			\textbf{0.992} & 0.0114  & 0.0110 &  4.03 & 0.001 \\
			\textbf{0.99} & 0.0260  & 0.0179 &  5.61 & 0.000 \\
			\textbf{0.98} & 0.1073  & 0.0253 &  16.4 & 0.000 \\
			\textbf{0.97} & 0.2824  & 0.0954 &  11.5 & 0.000 \\
			\textbf{0.96} & 0.4356  & 0.0767 &  22 & 0.000 \\
			\textbf{0.95} & 0.5118  & 0.0616 &  32.2 & 0.000 \\
			\textbf{0.94} & 0.5658  & 0.0568 &  38.6 & 0.000 \\
			\textbf{0.93} & 0.6385  & 0.0476 &  52 & 0.000 \\
			\textbf{0.92} & 0.7070  & 0.0510 &  53.7 & 0.000 \\
			\textbf{0.91} & 0.7574  & 0.0240 &  122 & 0.000 \\
			\textbf{0.9} & 0.7727  & 0.0090 &  333 & 0.000 \\

			\bottomrule
		\end{tabular*}
		\vspace{0.3cm}
		\caption{Sum of projections in VGG11 (n=15). $\mu \neq 0$ (p<.01) in \textbf{bold}.}
		\label{tab:vgg11ttest}
	\end{center}
\end{table*}

\begin{table*}[!htb]
	\begin{center}

		\begin{tabular*}{\textwidth}{c @{\extracolsep{\fill}} c c c c c c }
			\toprule
			Explained $\sigma$ & Mean difference & sample $\sigma$ & t-stat & p-value
			\\
			\midrule
			0.9999 & -0.0004  & 0.0008 & -2.42 & 0.023  \\
			\textbf{0.9998} & -0.0005  & 0.0009 & -2.81 & 0.010  \\
			\textbf{0.9997} & -0.0010  & 0.0010 & -5.26 & 0.000  \\
			\textbf{0.9996} & -0.0009  & 0.0010 & -4.92 & 0.000  \\
			\textbf{0.9995} & -0.0011  & 0.0010 & -5.46 & 0.000  \\
			\textbf{0.9994} & -0.0012  & 0.0012 & -4.91 & 0.000  \\
			\textbf{0.9993} & -0.0012  & 0.0012 & -4.83 & 0.000  \\
			\textbf{0.9992} & -0.0013  & 0.0013 & -5.17 & 0.000  \\
			\textbf{0.9991} & -0.0016  & 0.0015 & -5.48 & 0.000  \\
			\textbf{0.999} & -0.0017  & 0.0016 & -5.50 & 0.000  \\
			\textbf{0.998} & -0.0017  & 0.0022 & -3.92 & 0.001  \\
			0.996 & -0.0005  & 0.0030 & -0.910 & 0.371  \\
			\textbf{0.994} & 0.0037  & 0.0043 &  4.45 & 0.000  \\
			\textbf{0.992} & 0.0096  & 0.0062 &  7.91 & 0.000  \\
			\textbf{0.99} & 0.0178  & 0.0136 &  6.68 & 0.000  \\
			\textbf{0.98} & 0.1123  & 0.0377 &  15.2 & 0.000  \\
			\textbf{0.97} & 0.2254  & 0.0578 &  19.9 & 0.000  \\
			\textbf{0.96} & 0.4803  & 0.1022 &  24.0 & 0.000  \\
			\textbf{0.95} & 0.7026  & 0.0368 &  97.3 & 0.000  \\
			\textbf{0.94} & 0.7536  & 0.0227 &  169 & 0.000  \\
			\textbf{0.93} & 0.7654  & 0.0202 &  193 & 0.000  \\
			\textbf{0.92} & 0.7785  & 0.0164 &  242 & 0.000  \\
			\textbf{0.91} & 0.7867  & 0.0143 &  280 & 0.000  \\
			\textbf{0.9} & 0.7929  & 0.0117 &  345 & 0.000  \\
			\bottomrule

		\end{tabular*}
		\vspace{0.3cm}
		\caption{Sum of projections in VGG13 (n=26). $\mu \neq 0$ ($\alpha=0.01$) in \textbf{bold}.}
	\end{center}
\end{table*}

\begin{table*}[htb!]
	\begin{center}
		\begin{tabular*}{\textwidth}{c @{\extracolsep{\fill}} c c c c c c c c}
			\toprule
			Explained $\sigma$ & $\mu_{diff}$ & $\sigma_{sample}$ & t-stat & p-value & $\mu_{Sat}$ & $\sigma_{Sat}$ & $\mu(\sum dim E^k_l)$ \\
			\midrule
			0.9999& -0.0003  & 0.0008 & -2.65 & 0.011 & 60.0 & 0.6 & $2613 \pm 102$ \\
			\textbf{0.9998}& -0.0006  & 0.0011 & -3.31 & 0.002 & 54.5 & 0.6 & $2268 \pm 97$ \\
			\textbf{0.9997}& -0.0006  & 0.0014 & -2.82 & 0.008 & 51.2 & 0.7 & $2071 \pm 93$ \\
			0.9996& -0.0003  & 0.0016 & -1.28 & 0.208 & 48.8 & 0.6 & $1938 \pm 88$ \\
			0.9995& -0.0001  & 0.0017 & -0.352 & 0.727 & 47.1 & 0.7 & $1841 \pm 86$ \\
			0.9994& 0.0007  & 0.0019 &  2.18 & 0.035 & 45.6 & 0.7 & $1766 \pm 84$ \\
			0.9993& 0.0009  & 0.0022 &  2.62 & 0.012 & 44.5 & 0.7 & $1705 \pm 83$ \\
			0.9992& 0.0012  & 0.0031 &  2.42 & 0.020 & 43.4 & 0.7 & $1653 \pm 80$ \\
			\textbf{0.9991}& 0.0016  & 0.0032 &  3.14 & 0.003 & 42.5 & 0.7 & $1608 \pm 79$ \\
			\textbf{0.998}& 0.0107  & 0.0148 &  4.57 & 0.000 & 36.0 & 0.7 & $1318 \pm 73$ \\
			\textbf{0.996}& 0.0771  & 0.0585 &  8.33 & 0.000 & 30.0 & 0.7 & $1074 \pm 67$ \\
			\textbf{0.994}& 0.1873  & 0.0812 &  14.6 & 0.000 & 26.3 & 0.7 & $934 \pm 62$ \\
			\textbf{0.992}& 0.2754  & 0.0822 &  21.2 & 0.000 & 23.7 & 0.6 & $837 \pm 58$ \\
			\textbf{0.99}& 0.3643  & 0.0900 &  25.6 & 0.000 & 21.8 & 0.6 & $765 \pm 54$ \\
			\textbf{0.98}& 0.6176  & 0.0413 &  94.6 & 0.000 & 16.1 & 0.5 & $556 \pm 41$ \\
			\textbf{0.97}& 0.6559  & 0.0386 &  107 & 0.000 & 13.1 & 0.4 & $451 \pm 32$ \\
			\textbf{0.96}& 0.7008  & 0.0384 &  115 & 0.000 & 11.2 & 0.3 & $385 \pm 27$ \\
			\textbf{0.95}& 0.7351  & 0.0337 &  138 & 0.000 & 9.8 & 0.3 & $339 \pm 24$ \\
			\textbf{0.94}& 0.7550  & 0.0265 &  180 & 0.000 & 8.8 & 0.2 & $303 \pm 21$ \\
			\textbf{0.93}& 0.7639  & 0.0231 &  209 & 0.000 & 7.9 & 0.2 & $275 \pm 19$ \\
			\textbf{0.92}& 0.7727  & 0.0167 &  293 & 0.000 & 7.2 & 0.2 & $252 \pm 17$ \\
			\textbf{0.91}& 0.7775  & 0.0143 &  344 & 0.000 & 6.6 & 0.2 & $233 \pm 16$ \\
			\textbf{0.9}& 0.7796  & 0.0127 &  387 & 0.000 & 6.1 & 0.2 & $215 \pm 15$ \\

			\bottomrule

		\end{tabular*}
		\vspace{0.3cm}
		\caption{Sum of projections in VGG19 (n=40). $\mu \neq 0$ (p<.01) in \textbf{bold}.}

	\end{center}
\end{table*}

\begin{table*}[htb!]
	\begin{center}
		\begin{tabular*}{\textwidth}{c @{\extracolsep{\fill}} c c c c c c c c}
			\toprule
			Explained $\sigma$ & $\mu_{diff}$ & $\sigma_{sample}$ & t-stat & p-value & $\mu_{Sat}$ & $\sigma_{Sat}$ & $\mu(\sum dim E^k_l)$ \\
			\midrule
			1.0 & 0.0000  & 0.0000 &  nan & nan & 100.0 & 0.0 & $3904 \pm 0$ \\
			0.9999 & -0.0002  & 0.0012 & -0.52 & 0.611 & 78.5 & 0.5 & $2338 \pm 87$ \\
			0.9998 & 0.0000  & 0.0013 &  0.0796 & 0.938 & 75.6 & 0.4 & $2153 \pm 74$ \\
			0.9997 & -0.0002  & 0.0016 & -0.521 & 0.610 & 73.9 & 0.4 & $2043 \pm 67$ \\
			0.9996 & -0.0009  & 0.0020 & -1.66 & 0.119 & 72.5 & 0.4 & $1963 \pm 63$ \\
			0.9995 & -0.0005  & 0.0022 & -0.813 & 0.430 & 71.3 & 0.4 & $1900 \pm 61$ \\
			0.9994 & -0.0006  & 0.0019 & -1.18 & 0.256 & 70.4 & 0.3 & $1847 \pm 57$ \\
			0.9993 & -0.0007  & 0.0019 & -1.48 & 0.162 & 69.4 & 0.4 & $1802 \pm 55$ \\
			0.9992 & -0.0007  & 0.0022 & -1.29 & 0.217 & 68.7 & 0.4 & $1763 \pm 54$ \\
			0.9991 & -0.0006  & 0.0022 & -1.13 & 0.279 & 67.9 & 0.4 & $1728 \pm 52$ \\
			0.998 & 0.0031  & 0.0046 &  2.63 & 0.020 & 62.7 & 0.3 & $1493 \pm 40$ \\
			0.996 & 0.0213  & 0.0285 &  2.9 & 0.012 & 57.4 & 0.4 & $1294 \pm 32$ \\
			\textbf{0.994} & 0.0389  & 0.0454 &  3.32 & 0.005 & 54.0 & 0.5 & $1181 \pm 28$ \\
			\textbf{0.992} & 0.0579  & 0.0596 &  3.76 & 0.002 & 51.3 & 0.5 & $1100 \pm 27$ \\
			\textbf{0.99} & 0.0812  & 0.0782 &  4.02 & 0.001 & 49.2 & 0.6 & $1038 \pm 26$ \\
			\textbf{0.98} & 0.1899  & 0.1042 &  7.06 & 0.000 & 41.7 & 0.7 & $841 \pm 26$ \\
			\textbf{0.97} & 0.2918  & 0.1057 &  10.7 & 0.000 & 37.0 & 0.7 & $731 \pm 26$ \\
			\textbf{0.96} & 0.3649  & 0.0834 &  16.9 & 0.000 & 33.6 & 0.6 & $654 \pm 25$ \\
			\textbf{0.95} & 0.4333  & 0.0757 &  22.2 & 0.000 & 30.9 & 0.6 & $595 \pm 24$ \\
			\textbf{0.94} & 0.4544  & 0.0667 &  26.4 & 0.000 & 28.6 & 0.6 & $548 \pm 24$ \\
			\textbf{0.93} & 0.4787  & 0.0668 &  27.7 & 0.000 & 26.7 & 0.6 & $508 \pm 24$ \\
			\textbf{0.92} & 0.4896  & 0.0638 &  29.7 & 0.000 & 25.1 & 0.6 & $475 \pm 23$ \\
			\textbf{0.91} & 0.5119  & 0.0582 &  34 & 0.000 & 23.6 & 0.6 & $446 \pm 22$ \\
			\textbf{0.9} & 0.5296  & 0.0574 &  35.8 & 0.000 & 22.4 & 0.6 & $421 \pm 21$ \\

			\bottomrule
		\end{tabular*}
		\vspace{0.3cm}
		\caption{Sum of projections in ResNet18 (n=15). $\mu \neq 0$ (p<.01) in \textbf{bold}.}
		\label{tab:resnet18ttest}
	\end{center}
\end{table*}

\clearpage

\section{Tail Patterns on various Architectures and Datasets}
In this section, we will provide additional tail patterns that were observed during experiments.
The black vertical bar in some of these plots marks the first layer with the receptive field size of the input greater than the input resolution.
We find that this property predicts unproductive sequences of layers well for sequential architecture like the VGG-network family but not when more than one pathway is present (for example skip or dense connections).
The experiments use the same experimental setup described in appendix section \ref{sec:resolution}.

\subsection{Different Types of Tail Patterns - A brief explanation}
\begin{figure}[h!]
	\centering
	\subfloat[VGG16 tail layers maintain the quality of the intermediate solution]{\includegraphics[width=0.6\columnwidth]{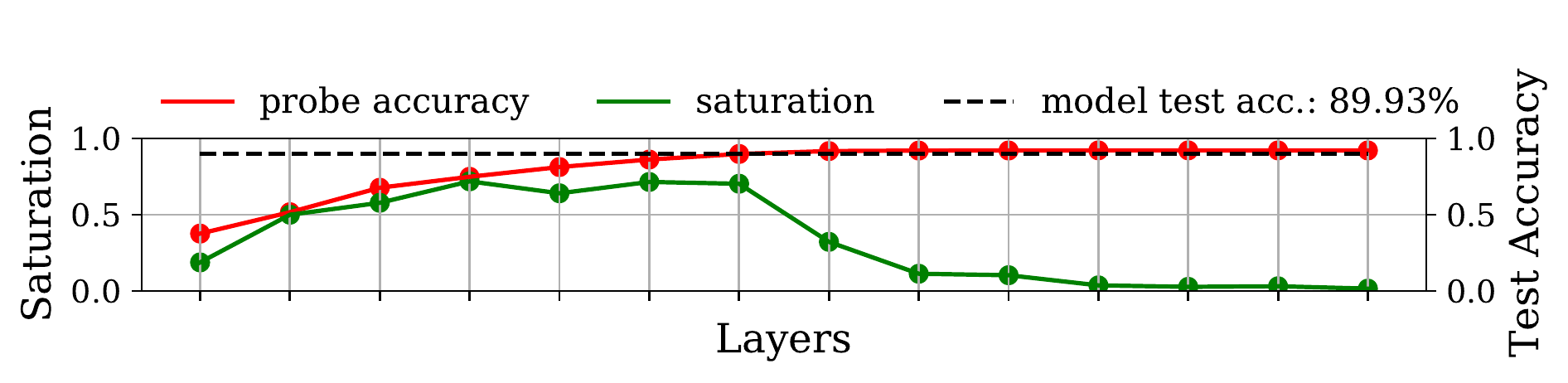}}\quad
	\subfloat[The tail of DenseNet18 shows a decay in probe performance, indicating that the last DenseBlock is skipped entirely \cite{alain2016}.]{\includegraphics[width=0.6\columnwidth]{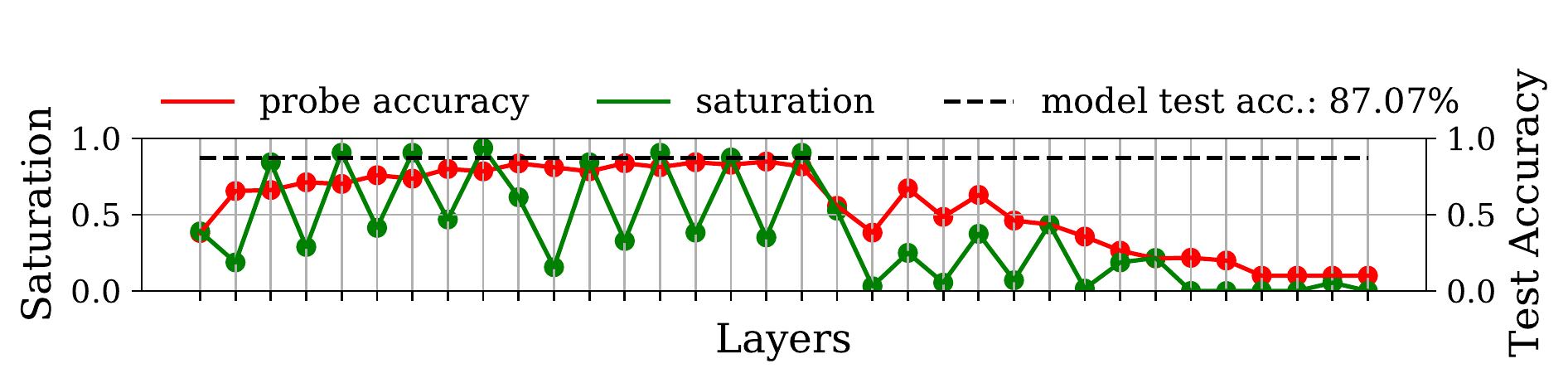}}\quad
	\subfloat[ResNet34 skips most residual blocks in the tail, which is apparent by the zig-zag pattern in probe performances caused by the starts and end of skip-connections \cite{alain2016}.]{\includegraphics[width=0.6\columnwidth]{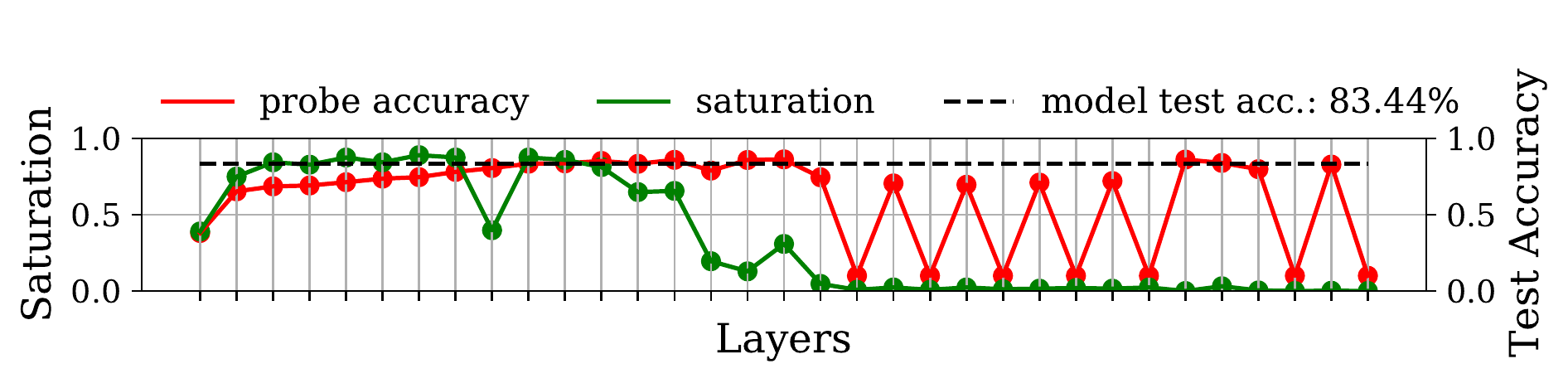}}
	\vspace{0.3cm}
	\caption{Depending on the neural architecture, tail patterns may deviate in their appearance in probe performance.
	In sequential architectures (a) the layers maintain the quality of the intermediate solution.
	If shortcut connections exist in the architecture, layers may be \textit{skipped}. 
	Skipped layers are apparent by their decaying probe performance \cite{alain2016}. 
	This is apparent on DenseNet18 (b) and ResNet34 (b) where a single DenseBlock and multiple ResiduaBlocks are skipped respectively. All models are trained on Cifar10 at native resolution.}
	\label{fig:tail_examples}
\end{figure}

We find that saturation is subject to noise induced by certain features of the neural architecture like the increase or decrease in filters from layer to layer, the use of $1 \times 1$ convolutions and downsampling layers are common culprits for zig-zag-like behavior or sudden dips and spikes in saturation, an example for the latter is DenseNet18 in figure \ref{fig:tail_examples} (b).
It has to be stressed that these factors are not random or create non-reproducible perturbations. Instead, they usually result in anomalous patterns that a very stable over multiple runs (which is exemplified in Section \ref{sec:stability}).

Logistic regression probes are considerably more robust against the aforementioned properties. However, they are influenced by the path the information takes during the forward pass, revealing different \textit{types} of tail patterns that can be differentiated based on the processing in the tail-layers.
The three examples found commonly are exemplified in figure \ref{fig:tail_examples}. 
These example also give insights into how neural network process information differently, which is the main reason why we dedicate an additional section to these findings in the appendix.
All the networks are trained on Cifar10 using a $32 \times 32$ pixel input resolution.
In figure \ref{fig:tail_examples} (a) we find a pass-through tail, where the layers process the information but do not advance the quality of the intermediate solution.
We find this type of tail pattern is typical for sequential neural networks (which you can see from other results in appendix \ref{apx:noskip_tails}, \ref{apx:mnist_tails}) and \ref{apx:tinyimgnet_tails}).
The second type of tail, depicted in figure \ref{fig:tail_examples} (b), is caused by the multiple pathways inside the DenseBlock of DenseNet.
Information can pass from any previous layer to the current layer within the DenseBlock, effectively allowing the information to skip layers.
When layers are skipped, the intermediate solution quality degrades and instantaneously recovers after the skipped section is over.
The latter is apparent in the depicted example by the high model performance relative to the probe performance of the last DenseBlock layers.
This phenomenon was initially observed on a simple MLP-example by Alain and Bengio \cite{alain2016}.
If necessary, the signal may jump more than a single building block in the architecture.
An example of which can be seen in figure \ref{fig:tail_examples} (c) on a ResNet34 architecture.
This jumping is indicated by the zig-zag-pattern in the probe performance, where the higher performing layer resembles the first and lower performing layer the second layer of a residual block.

This shows that architecture decisions, influencing the potential pathway's information can take from input to output, can have a significant influence on the way the model processes (or chooses not to process) information.
In any case, the semantic of the tail-pattern remains unchanged, since a skipped layer and an unproductive layer can both be considered a parameter and computational inefficiency.

\clearpage

\subsection{VGG11, 13, 16, 19 - MNIST}

\begin{figure}[htb!]
	\centering
	\includegraphics[width=0.9\textwidth]{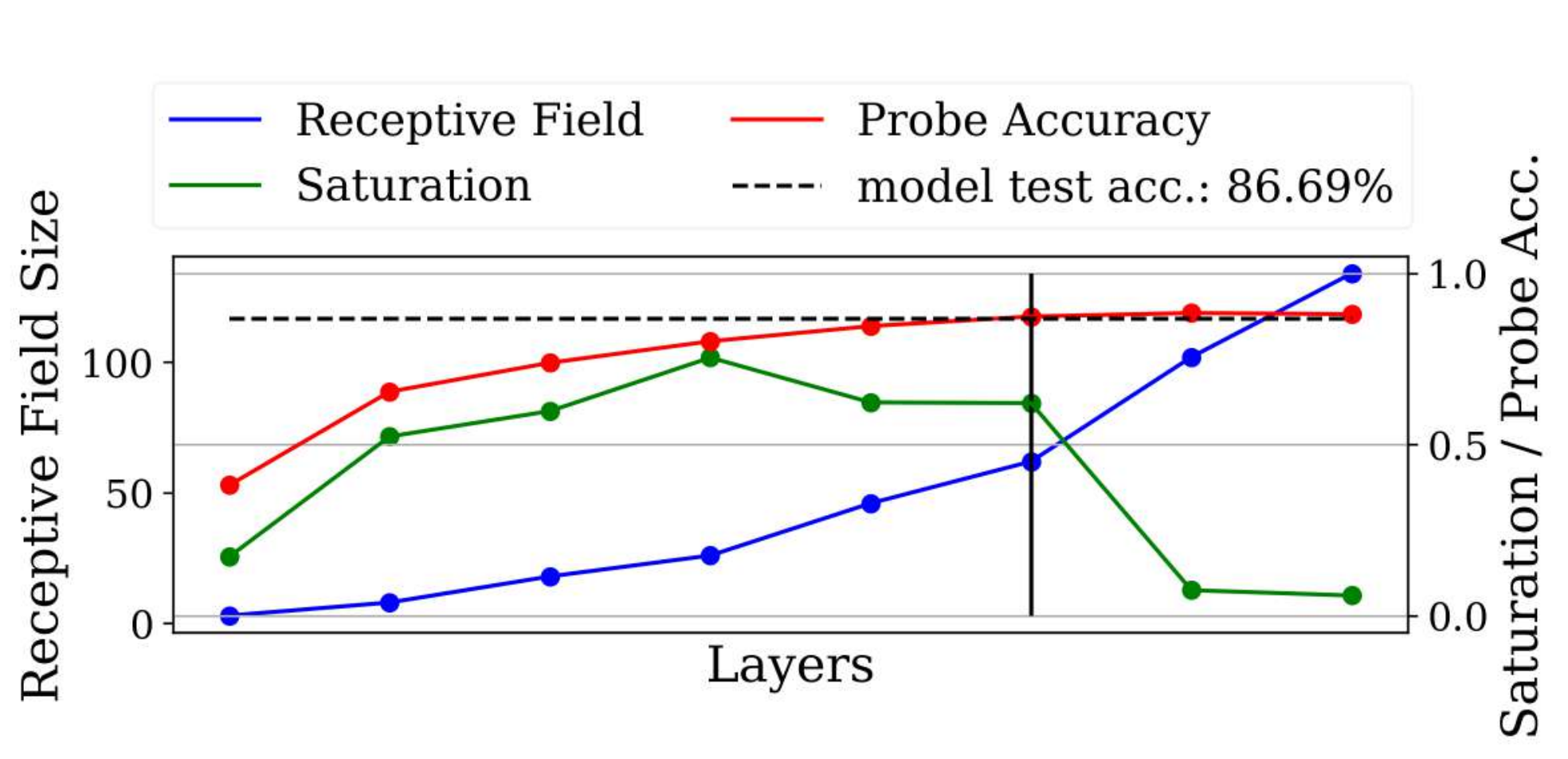}/
	\vspace{0.3cm}
	\caption{VGG11 - Mnist - $32 \times 32$ input resolution.}
\end{figure}

\begin{figure}[htb!]
	\centering
	\includegraphics[width=0.9\columnwidth]{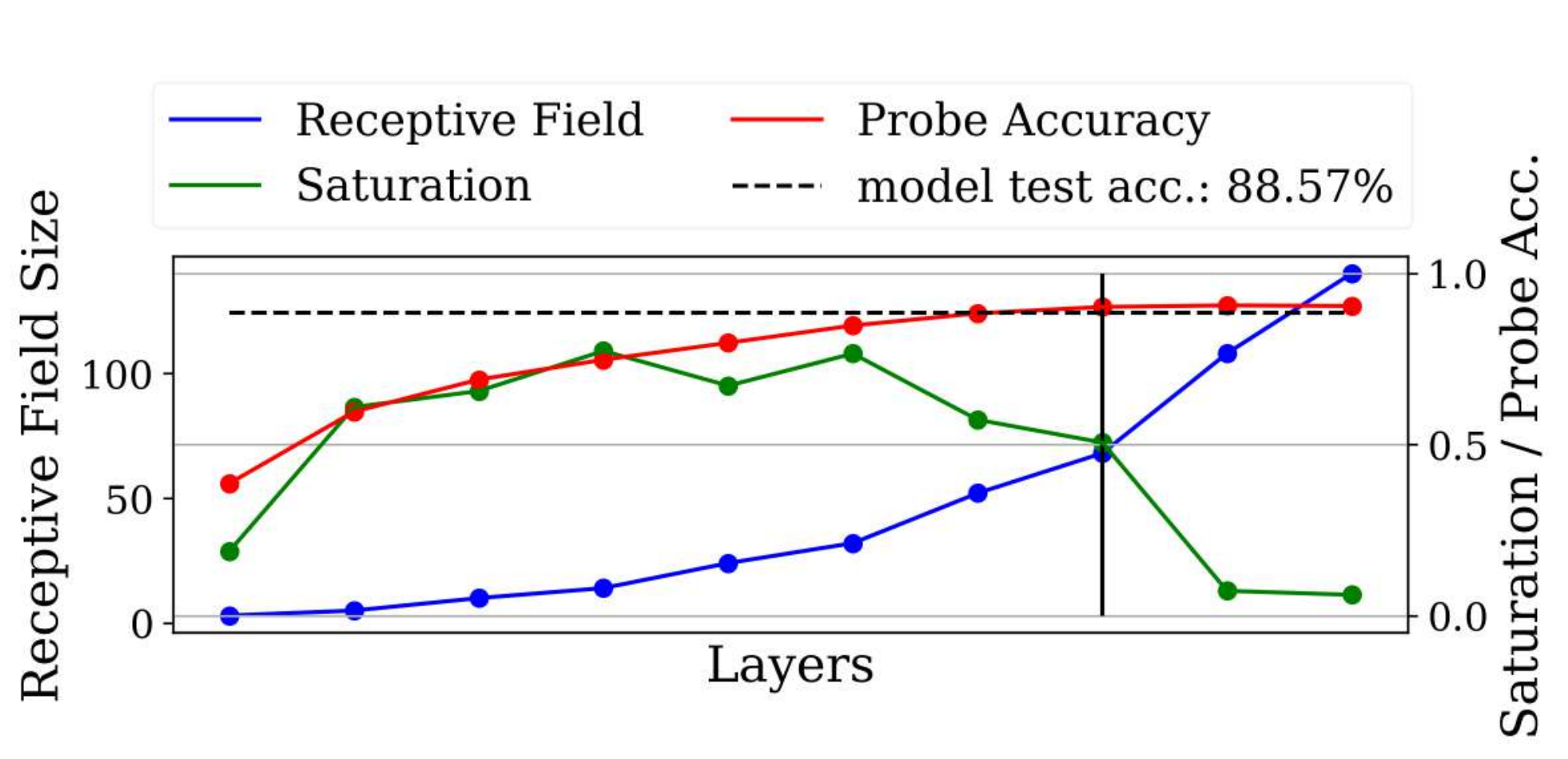}/
	\vspace{0.3cm}
	\caption{VGG13 - Mnist - $32 \times 32$ input resolution.}
\end{figure}

\begin{figure}[htb!]
	\centering
	\includegraphics[width=0.9\columnwidth]{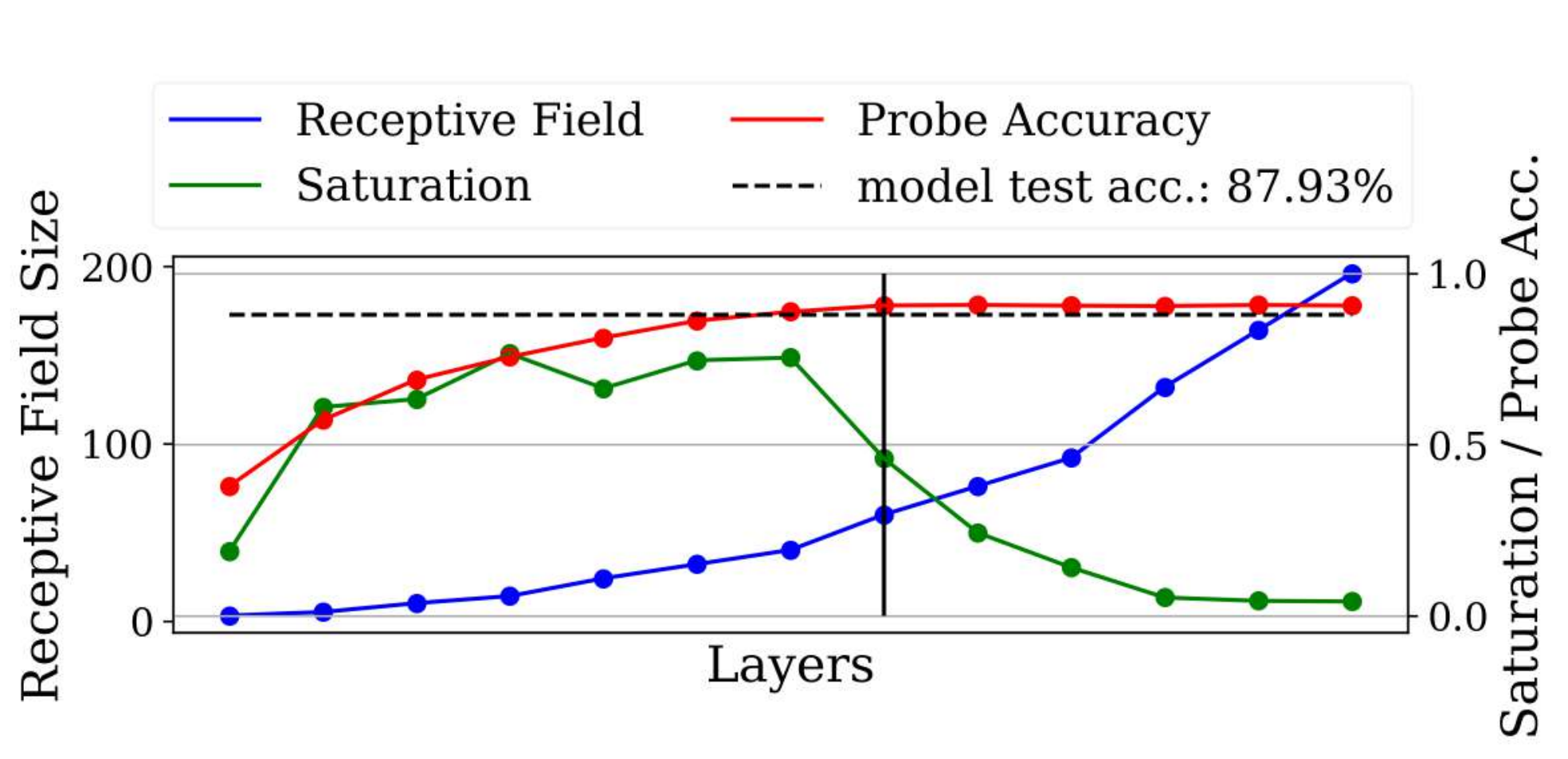}/
	\vspace{0.3cm}
	\caption{VGG16 - Mnist - $32 \times 32$ input resolution.}
\end{figure}

\begin{figure}[htb!]
	\centering
	\includegraphics[width=0.9\columnwidth]{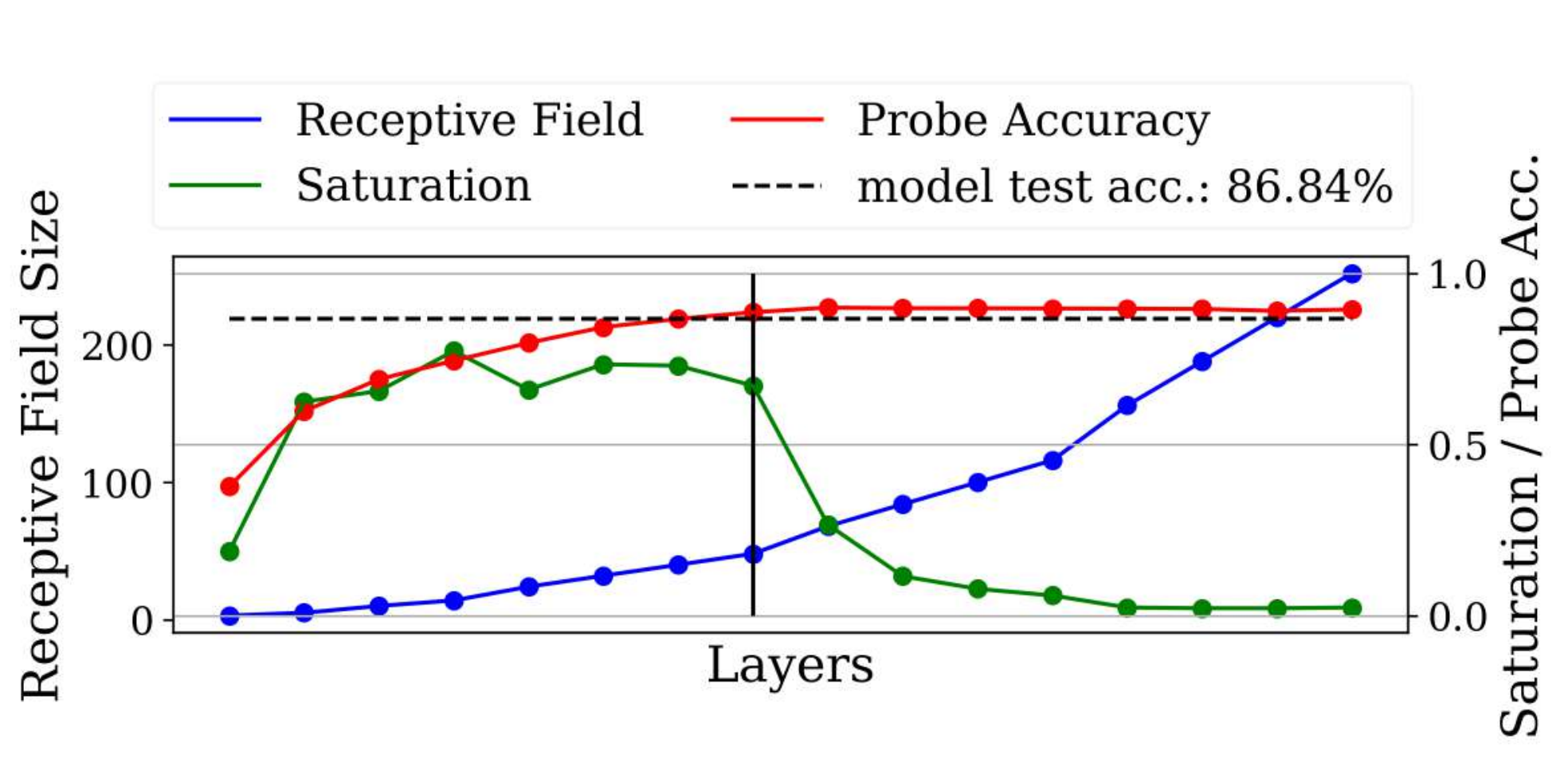}/
	\vspace{0.3cm}
	\caption{VGG19 - Mnist - $32 \times 32$ input resolution.}
\end{figure}

\clearpage
\subsection{VGG11, 13, 16, 19 - TinyImageNet}

\begin{figure}[htb!]
	\centering
	\includegraphics[width=0.9\columnwidth]{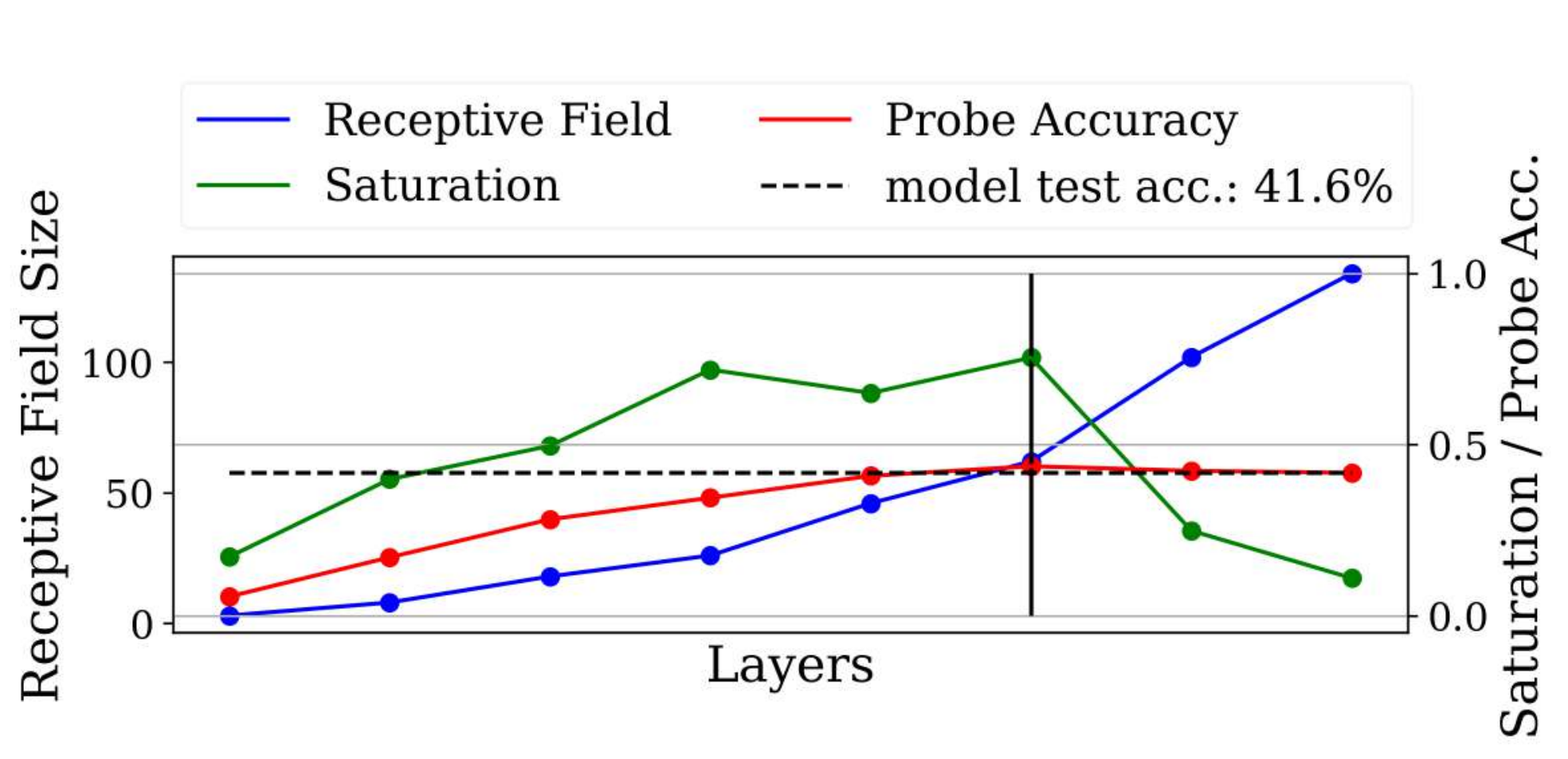}/
	\vspace{0.3cm}
	\caption{VGG11 - TinyImageNet - $32 \times 32$ input resolution.}
\end{figure}

\begin{figure}[htb!]
	\centering
	\includegraphics[width=0.9\columnwidth]{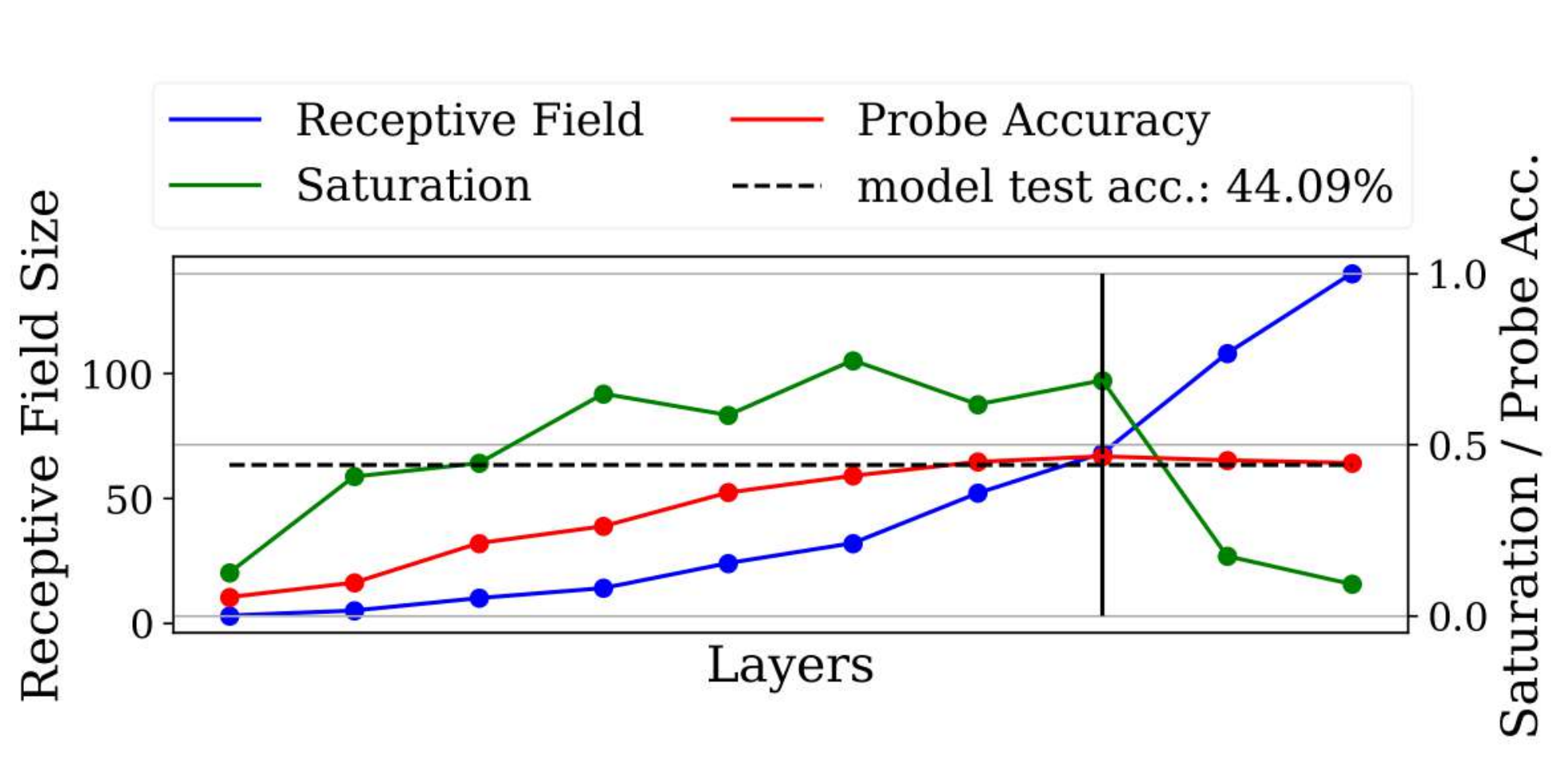}/
	\vspace{0.3cm}
	\caption{VGG13 - TinyImageNet - $32 \times 32$ input resolution.}
\end{figure}

\begin{figure}[htb!]
	\centering
	\includegraphics[width=0.9\columnwidth]{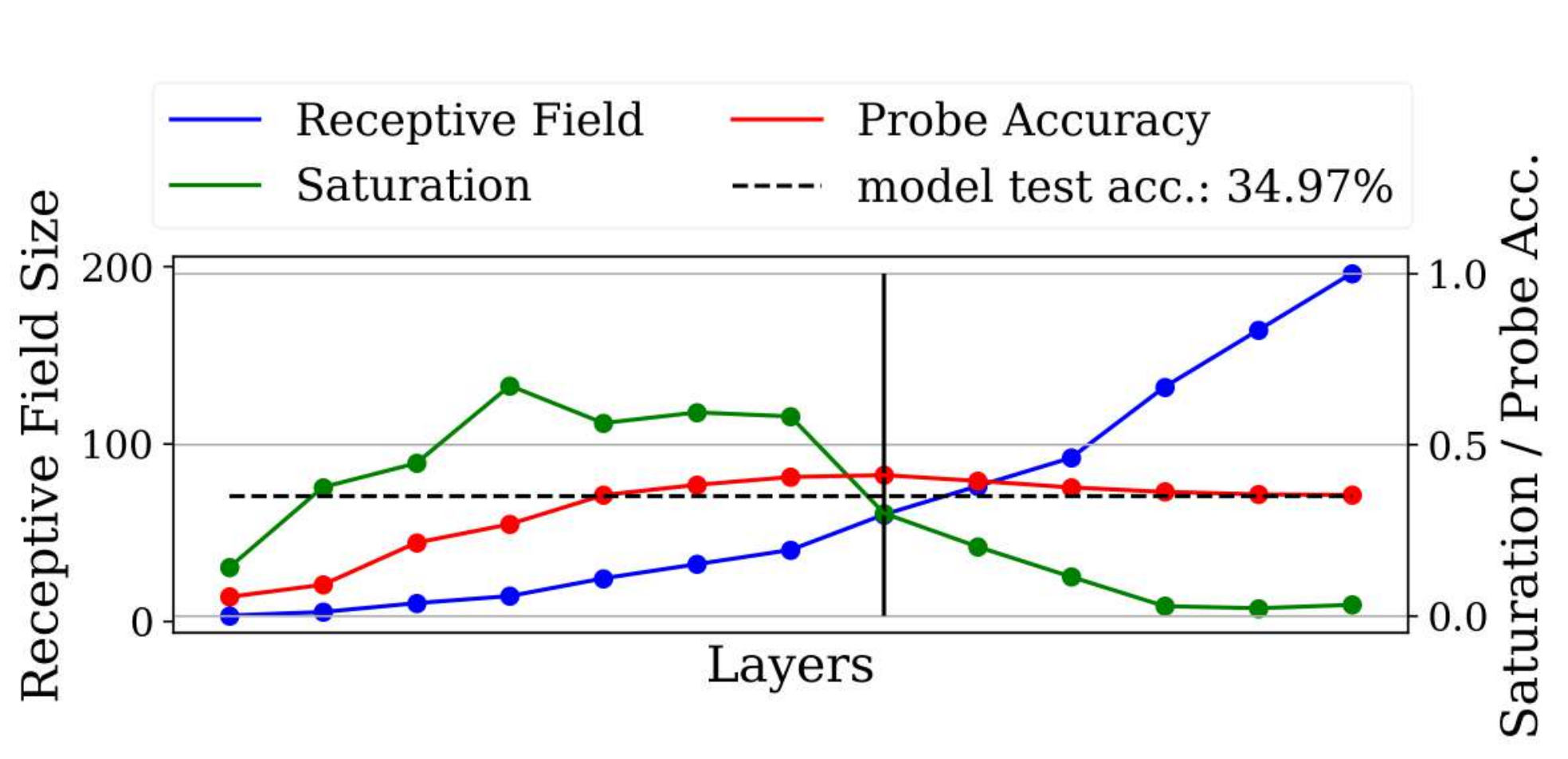}/
	\vspace{0.3cm}
	\caption{VGG16 - TinyImageNet - $32 \times 32$ input resolution.}
\end{figure}

\begin{figure}[htb!]
	\centering
	\includegraphics[width=0.9\columnwidth]{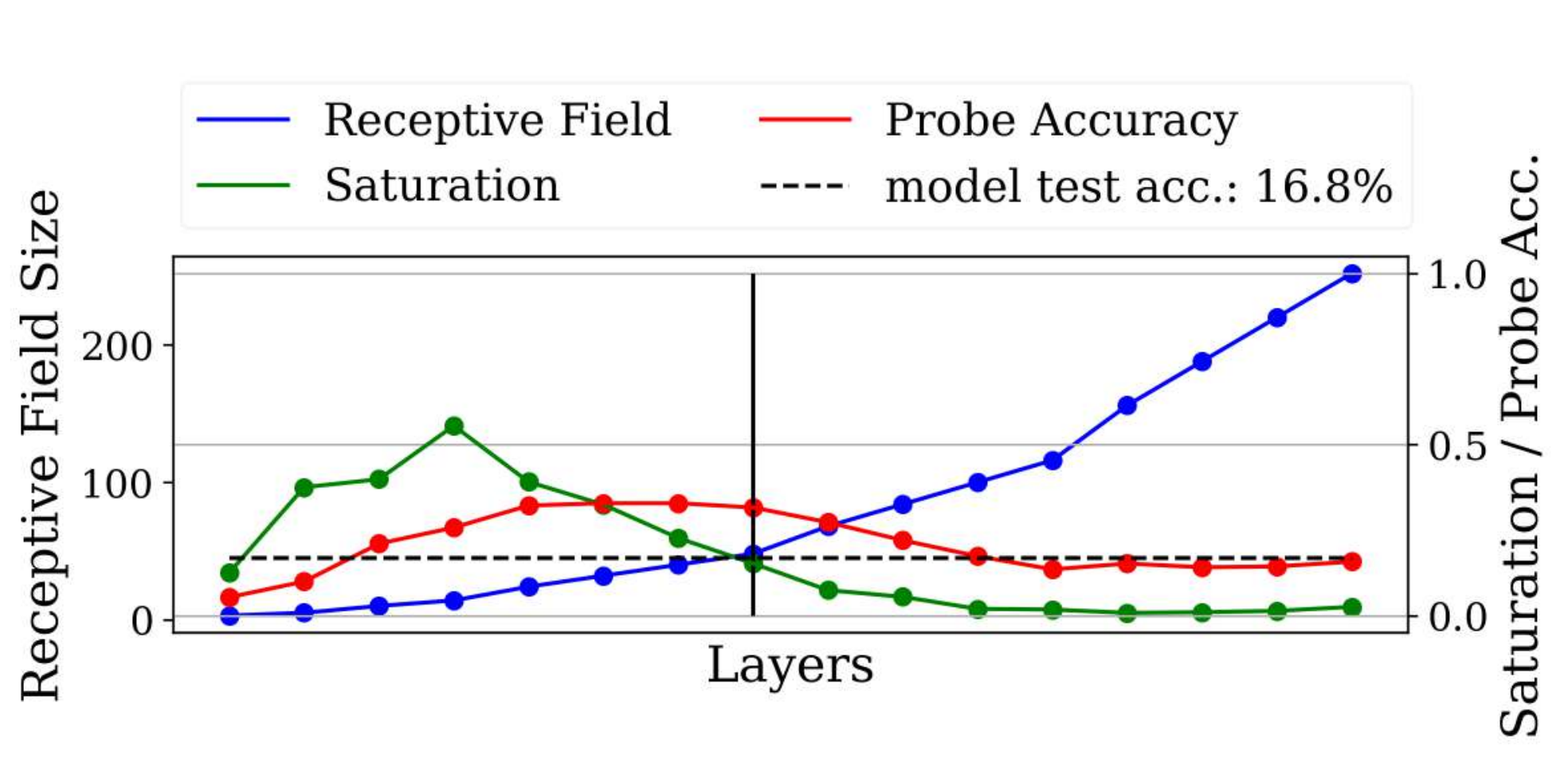}/
	\vspace{0.3cm}
	\caption{VGG19 - TinyImageNet - $32 \times 32$ input resolution.}
\end{figure}

\clearpage

\subsection{DenseNet18, 65 - Cifar10}
Interestingly, the skipping behavior observable in too deep ResNet-style architectures is not present in DenseNet-style networks. Instead, the probe accuracy degrades over entire regions of the network, indicating that these are likely skipped entirely.

\begin{figure}[htb!]
	\centering
	\includegraphics[width=0.9\columnwidth]{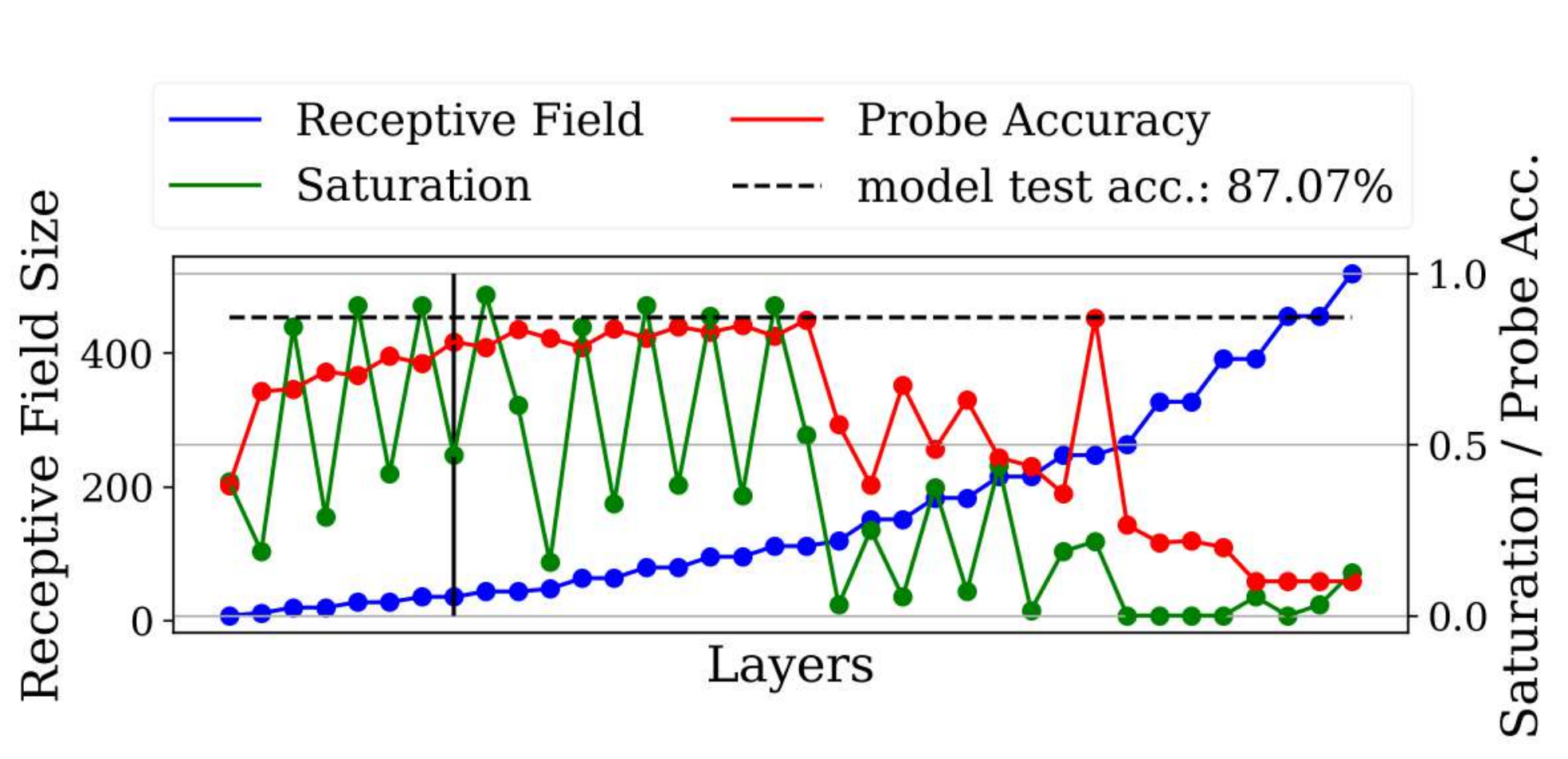}/
	\vspace{0.3cm}
	\caption{DenseNet18 - Cifar10 - $32 \times 32$ input resolution.}
\end{figure}

\begin{figure}[htb!]
	\centering
	\includegraphics[width=0.9\columnwidth]{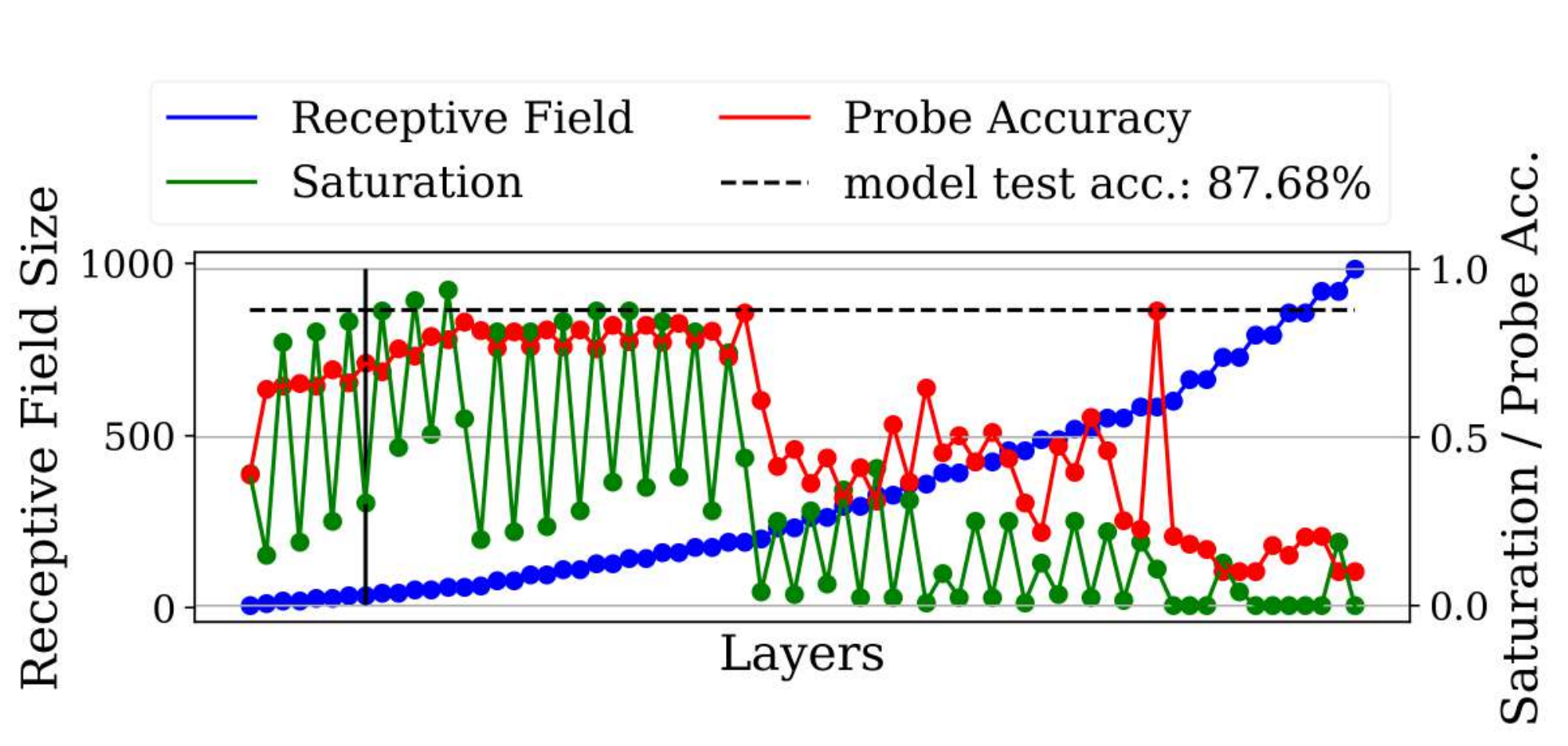}/
	\vspace{0.3cm}
	\caption{DenseNet65 - Cifar10 - $32 \times 32$ input resolution.}
\end{figure}

\clearpage
\subsection{ResNet50 - Cifar10}
Tail patterns are present in ResNet 50.

\begin{figure}[htb!]
	\centering
	\includegraphics[width=0.9\columnwidth]{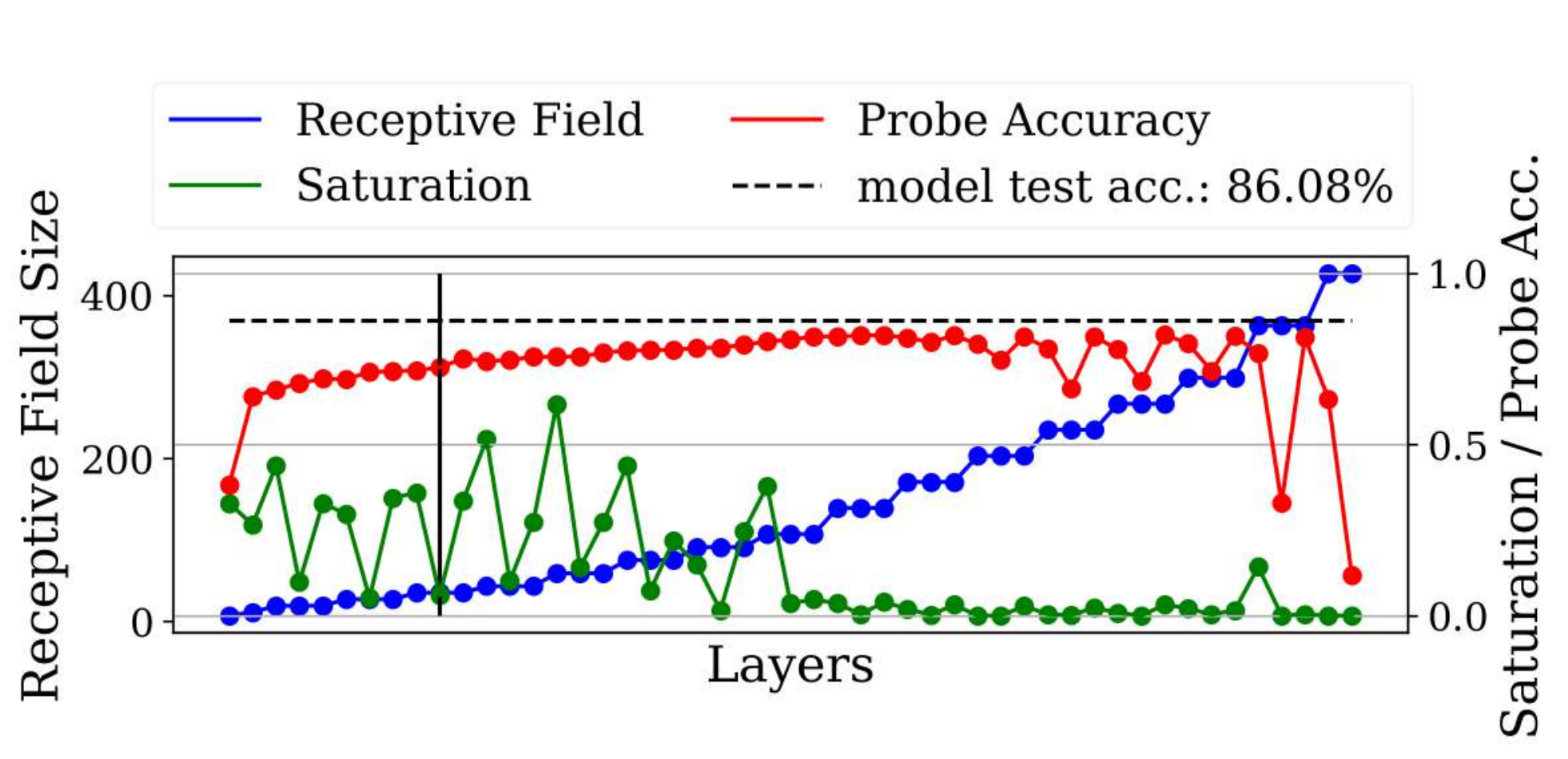}/
	\vspace{0.3cm}
	\caption{ResNet50 - Cifar10 - $32 \times 32$ input resolution.}
\end{figure}

\begin{figure}[htb!]
	\centering
	\includegraphics[width=0.9\columnwidth]{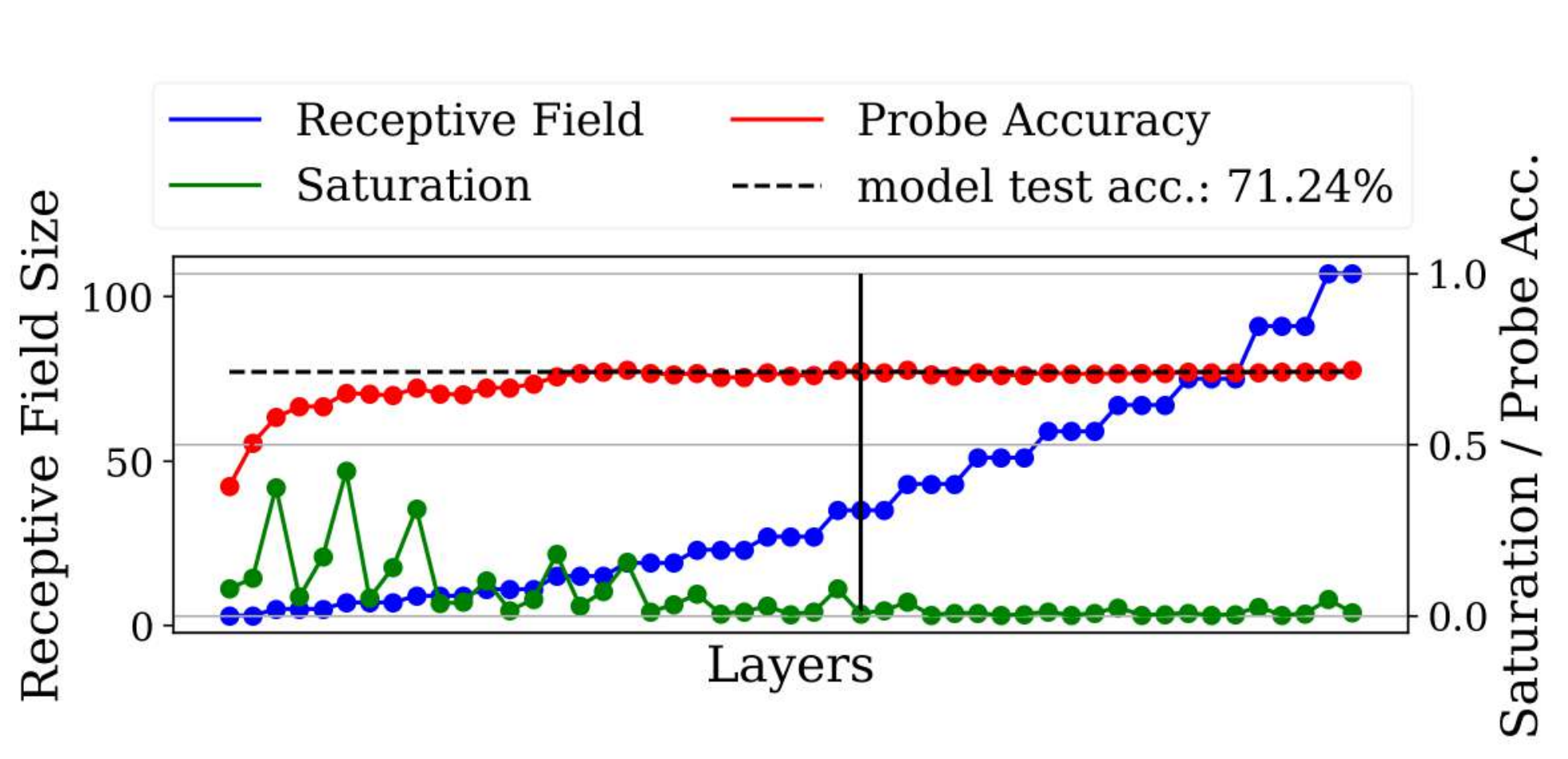}/
	\vspace{0.3cm}
	\caption{ResNet50 (removed stem) - Cifar10 - $32 \times 32$ input resolution.}
\end{figure}
\clearpage

\subsection{Experiments on ImageNet and iNaturalist}
This set of experiments is an attempt to recreate the tail pattern phenomenon on ImageNet and iNaturalist.
For these experiments, computing probe performances was not feasible due to resource limitations.
For this reason, only saturation is provided.
Each model is trained two times. Once on the design resolution of $224 \times 224$ pixels of the respective models (for reference purposes, we do not expect to see a tail pattern at this resolution) and once on $32 \times 32$ pixels, which reliably results in tail patterns for these models.

\begin{figure}[htb!]
	\centering
	\subfloat[ResNet18 - ImageNet - $32 \times 32$ - Test Accuracy: 25.88\%]{\includegraphics[width=0.9\columnwidth]{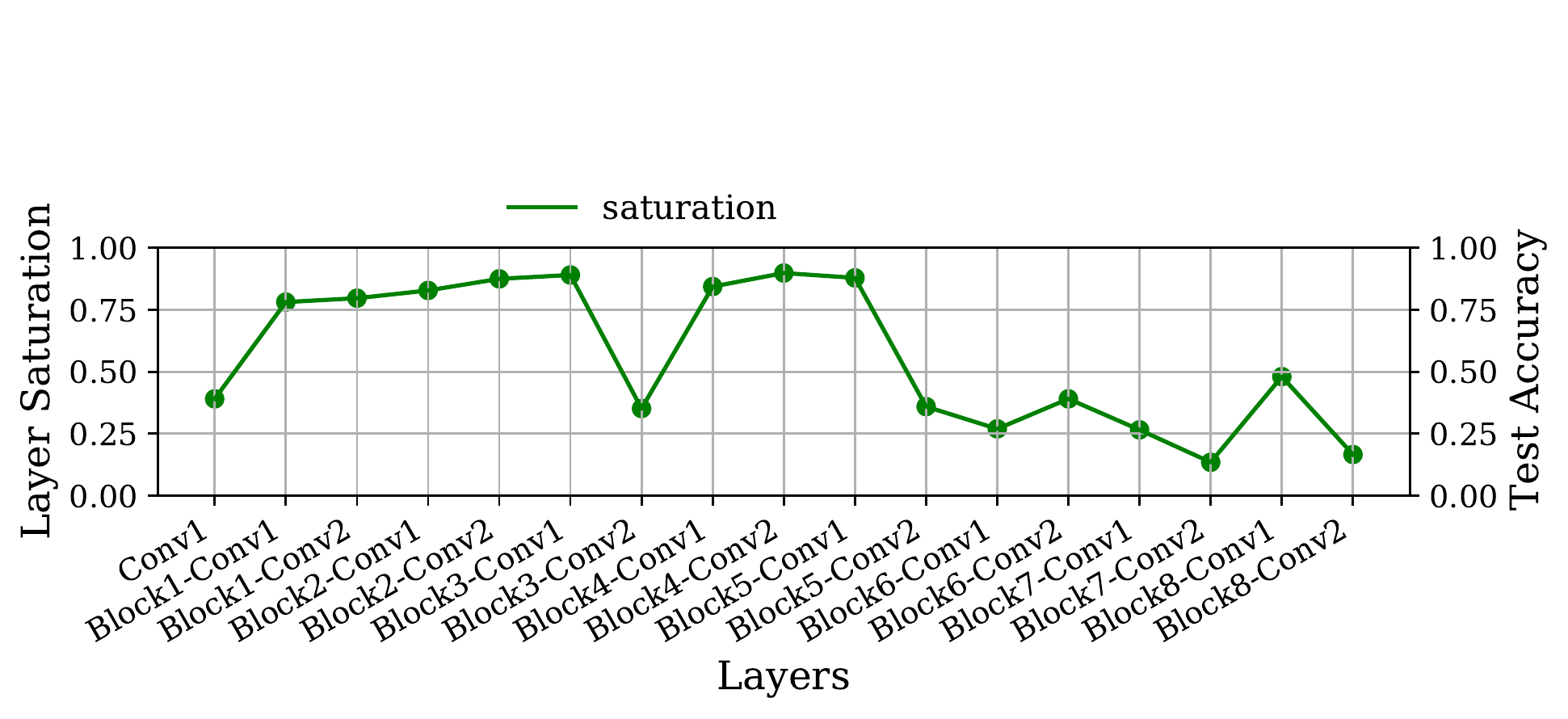}}\quad
	\subfloat[ResNet18 - ImageNet - $224 \times 224$ - Test Accuracy: 65.63\%]{\includegraphics[width=0.9\columnwidth]{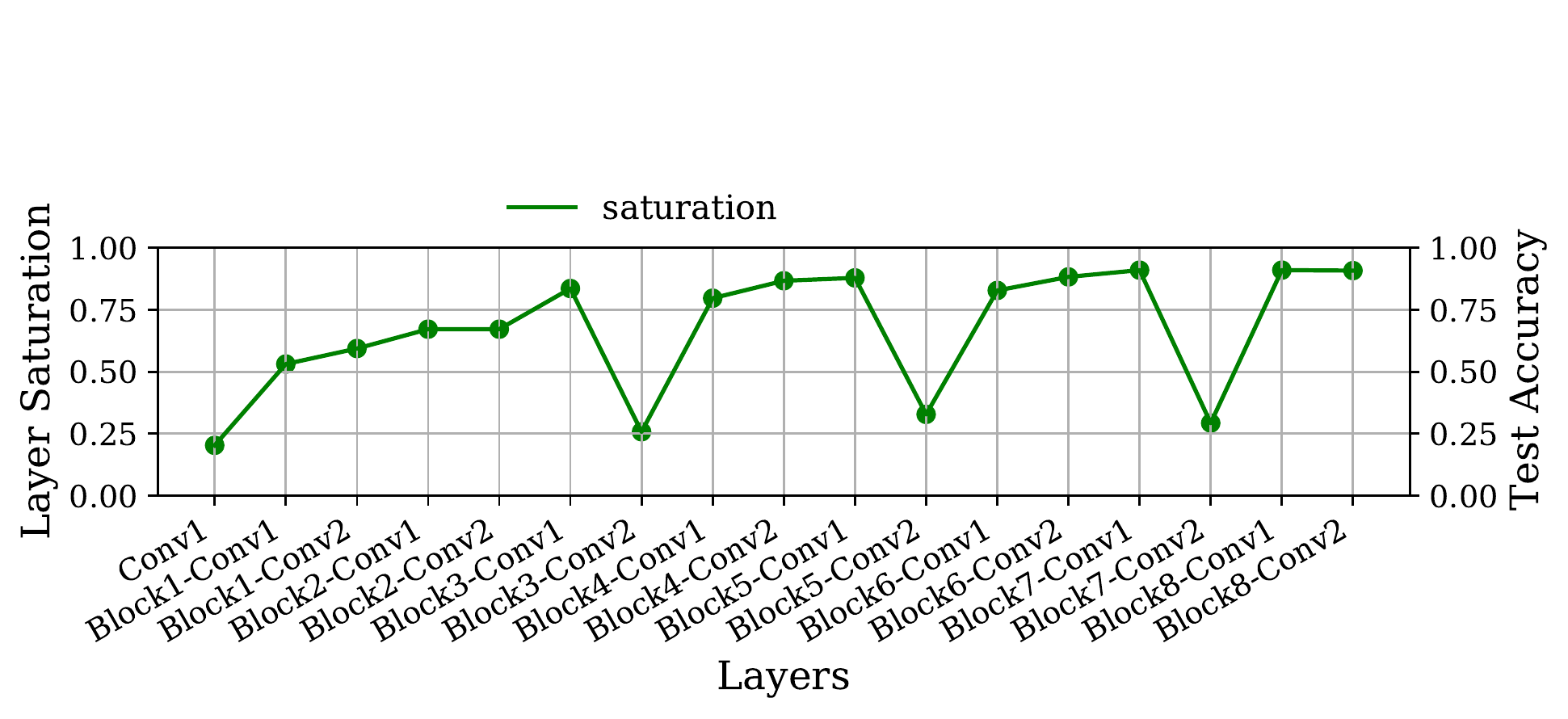}}
	\vspace{0.3cm}
	\captionsetup{justification=centering}
	\label{fig:imgnetresnet}
\end{figure}

\begin{figure}[htb!]
	\centering
	\subfloat[ResNet18 - iNaturalist - $32 \times 32$ - Test Accuracy: 12.15\%]{\includegraphics[width=0.9\columnwidth]{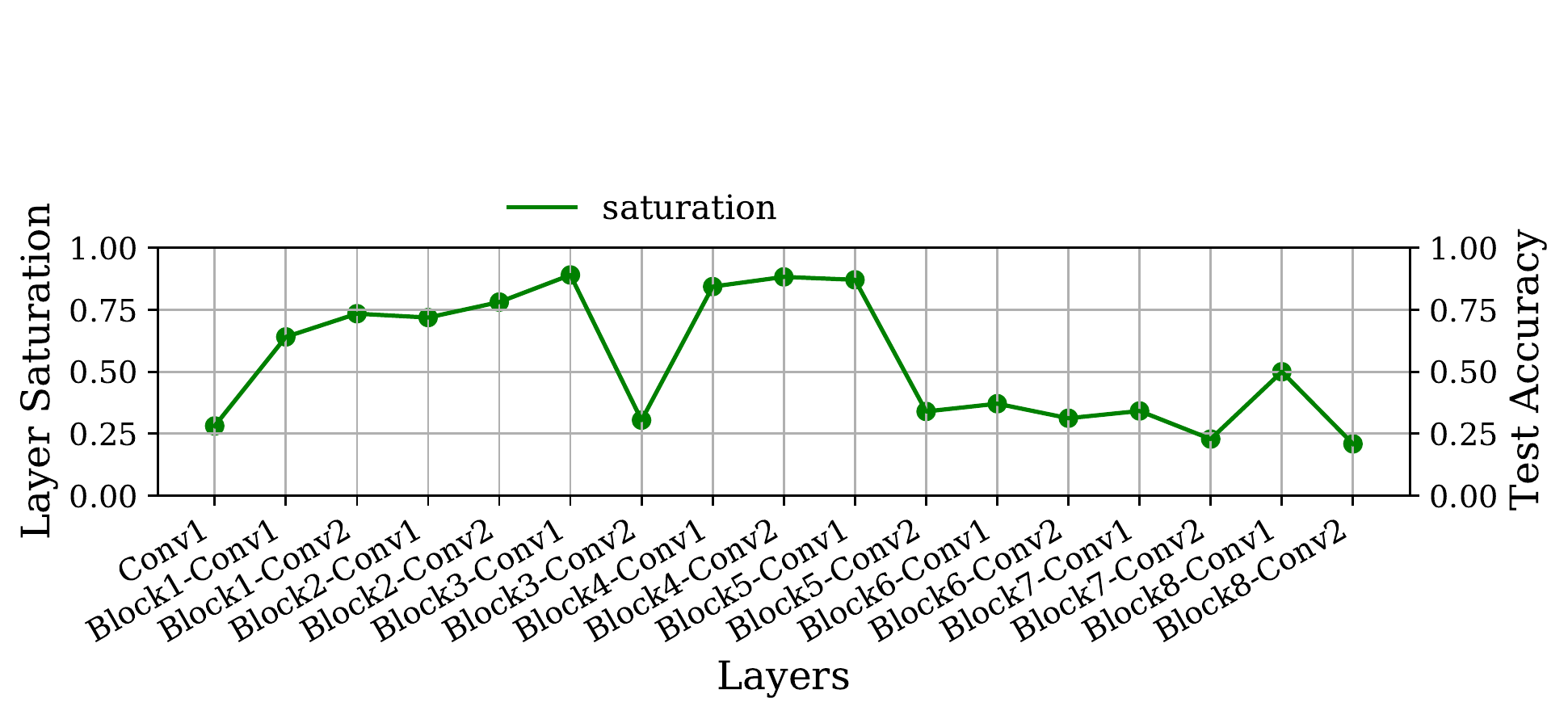}}\quad
	\subfloat[ResNet18 - iNaturalist - $224 \times 224$ - Test Accuracy: 39.91\%]{\includegraphics[width=0.9\columnwidth]{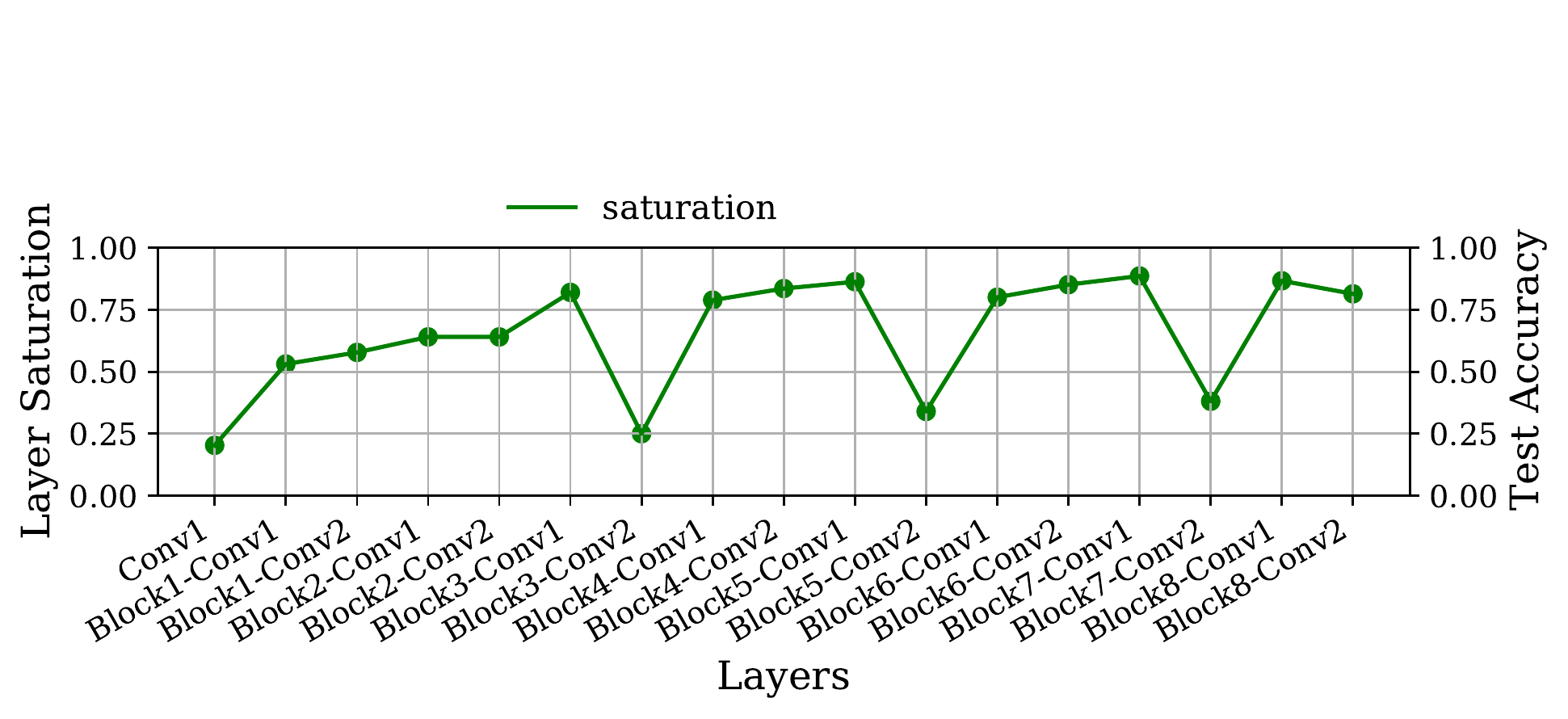}}
	\vspace{0.3cm}
	\captionsetup{justification=centering}
	\caption{ResNet18 trained on iNaturalist.}
	\label{fig:inaturalistresnet}
\end{figure}

\clearpage

\begin{figure}[htb!]
	\centering
	\subfloat[VGG16 - ImageNet - $32 \times 32$ - Test Accuracy: 10.13\%]{\includegraphics[width=0.9\columnwidth]{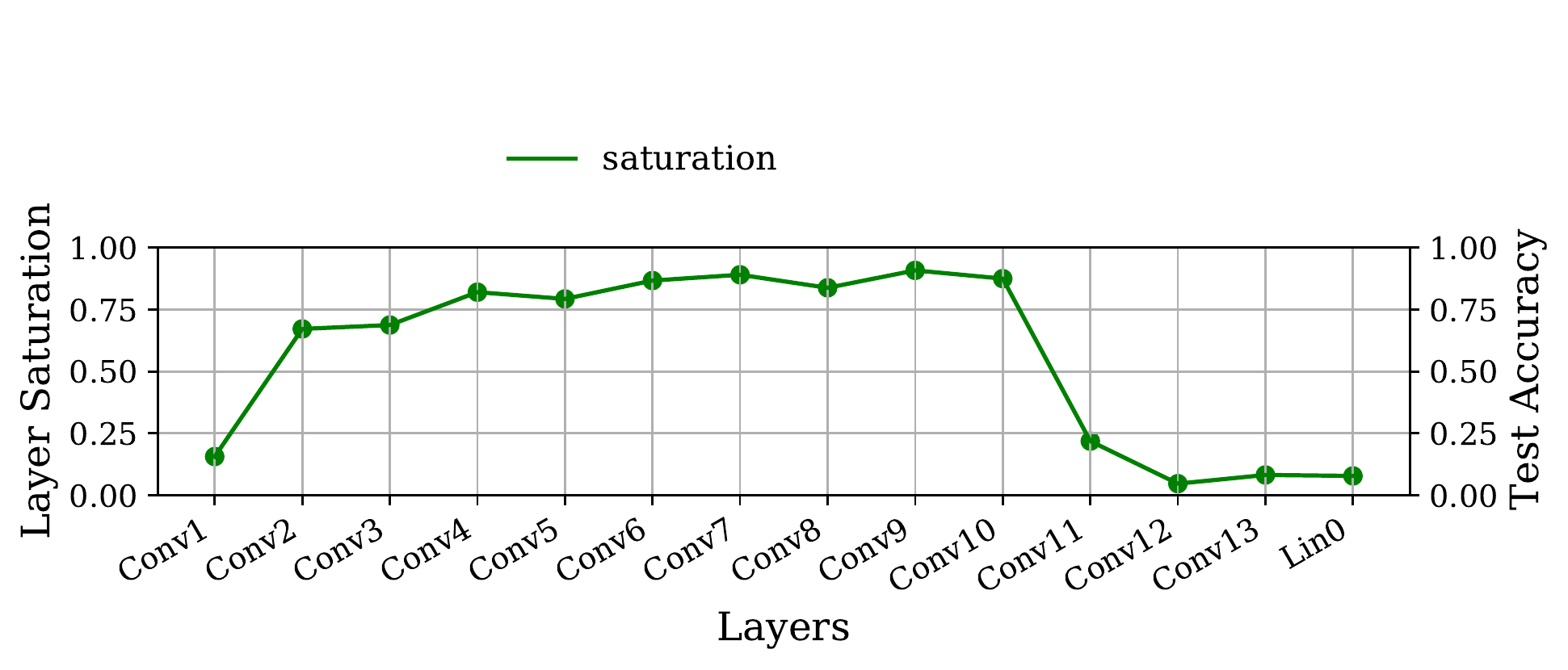}}\quad
	\subfloat[VGG16 - ImageNet - $224 \times 224$ - Test Accuracy: 63.96\%]{\includegraphics[width=0.9\columnwidth]{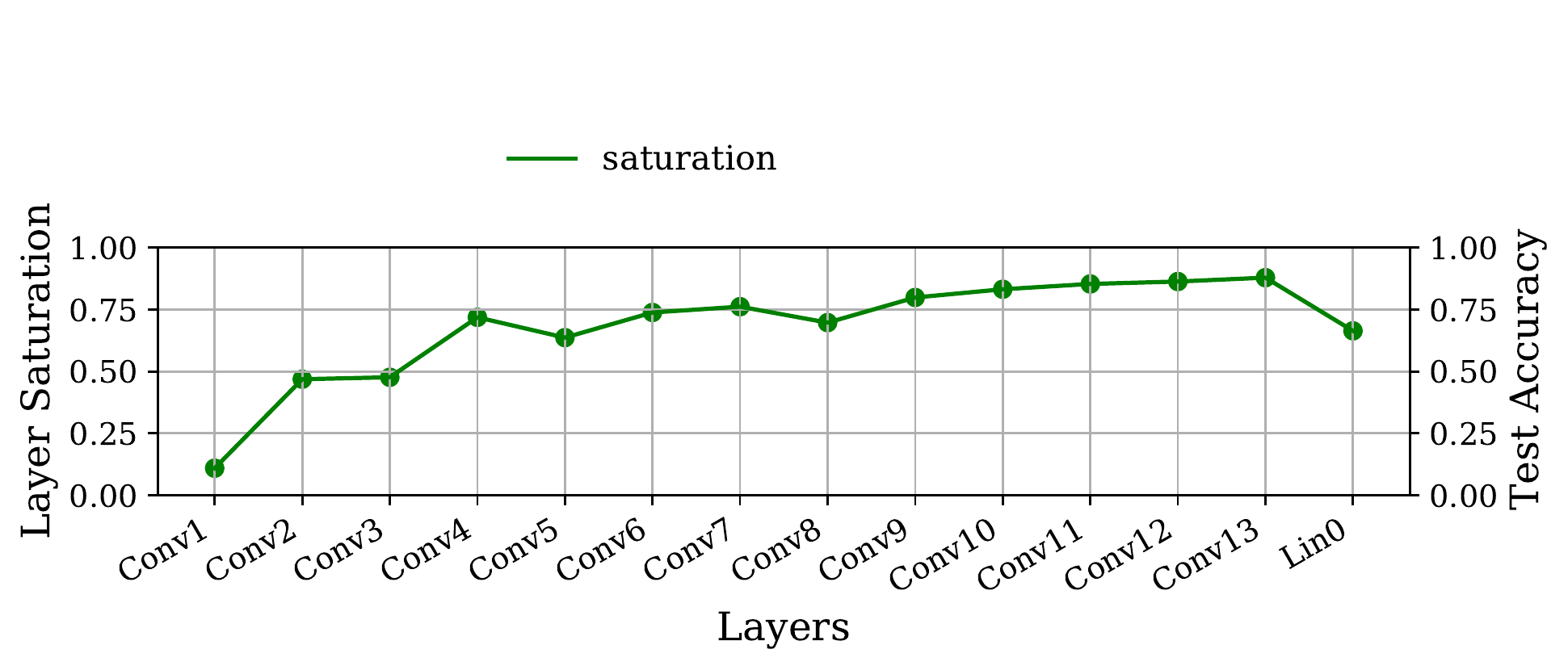}}
	\vspace{0.3cm}
	\captionsetup{justification=centering}
	\caption{VGG16 trained on ImageNet.}
	\label{fig:vggimnet}
\end{figure}

\begin{figure}[htb!]
	\centering
	\subfloat[VGG16 - iNaturalist - $32 \times 32$ - Test Accuracy: 17.85\%]{\includegraphics[width=0.9\columnwidth]{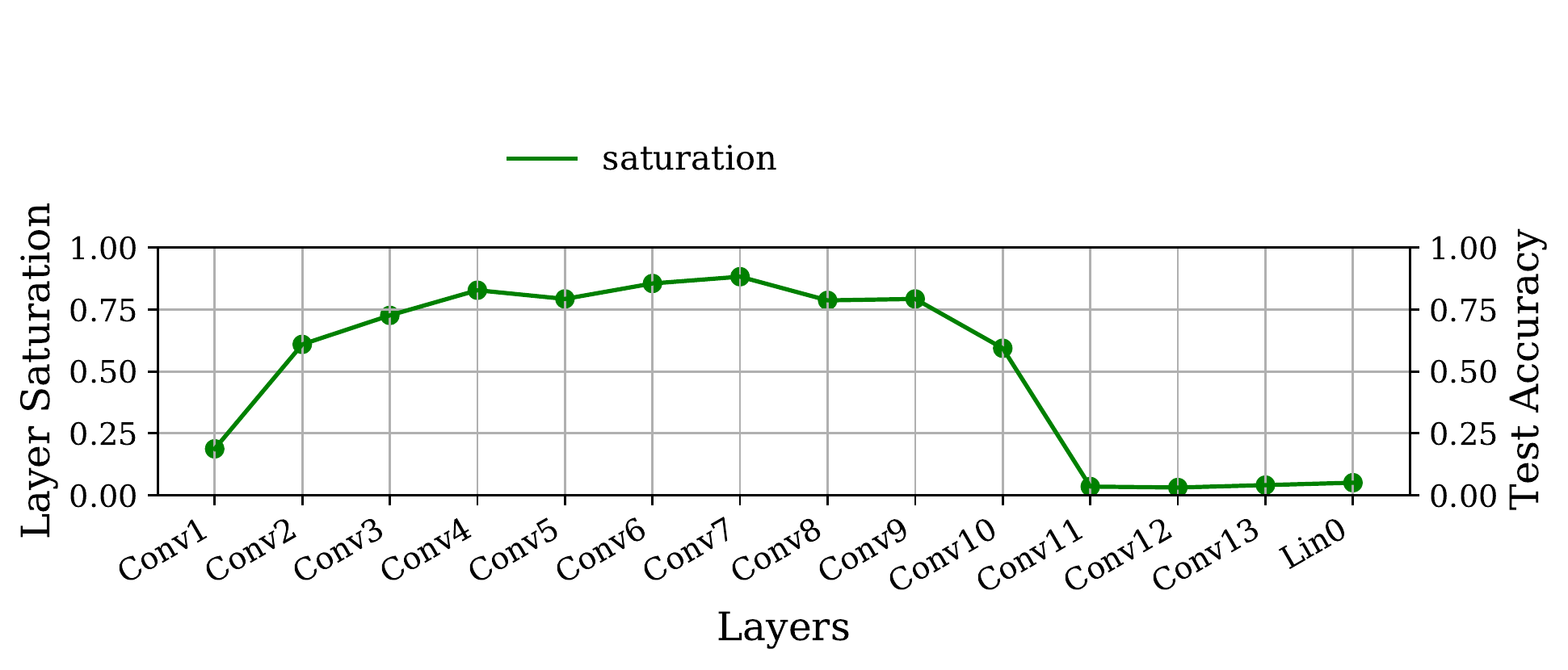}}\quad
	\subfloat[VGG16 - iNaturalist - $224 \times 224$ - Test Accuracy: 52.11\%]{\includegraphics[width=0.9\columnwidth]{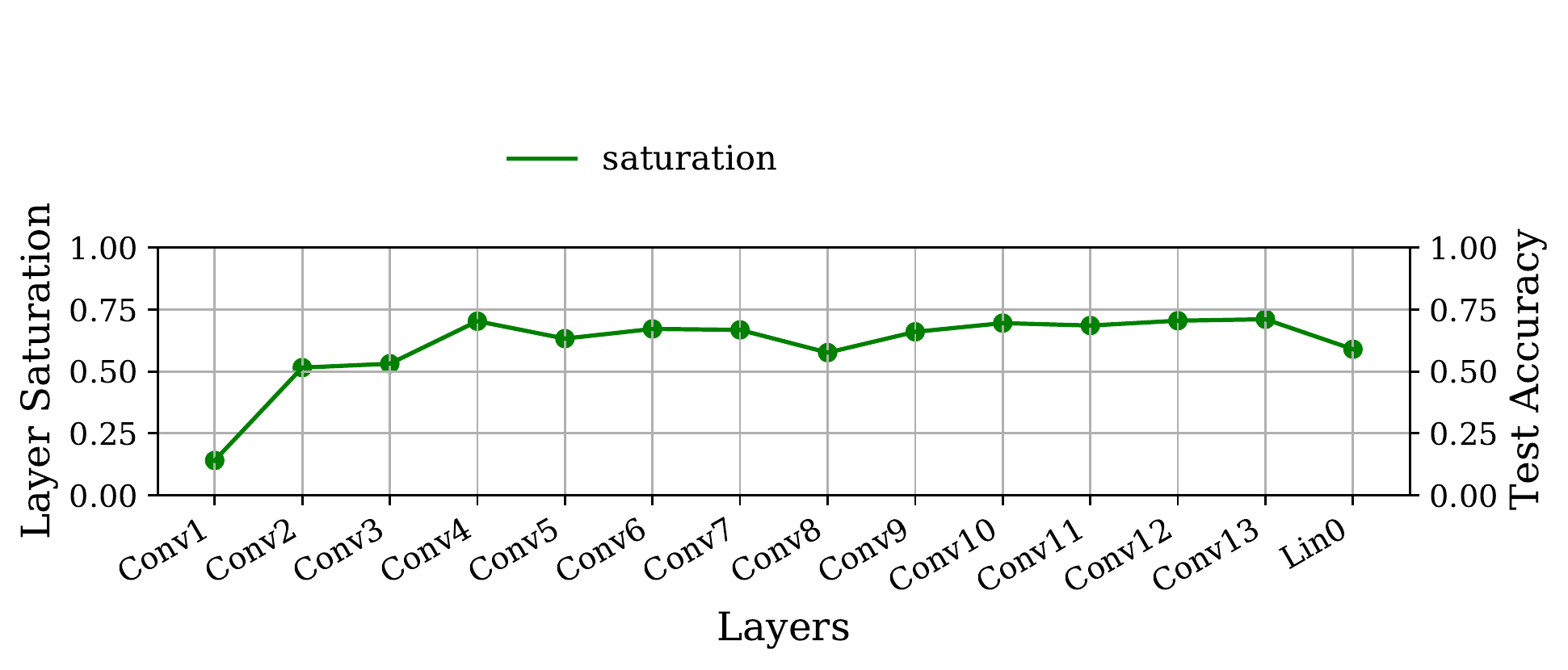}}
	\vspace{0.3cm}
	\captionsetup{justification=centering}
	\caption{VGG16 trained on iNaturalist.}
	\label{fig:vgginat}
\end{figure}

\clearpage

\section{Source Code}
The experiments conducted in this work are done in two distinct repositories.
The experiments themselves are conducted with the phd-lab-repository, which can be found here (including a manual): \url{https://github.com/MLRichter/phd-lab}.

The second repository is called \textit{delve} and contains the logic for PCA-Layers (see section \ref{sec:layer-eigenspaces}), on-line covariance approximation, as well experiment control.
All three features are used by phd-lab to conduct the experiments in question. This project is currently in the process of being open sourced, is installable over PyPi and can be found here: \url{https://github.com/delve-team/delve}.

\end{document}